\documentclass[1p]{elsarticle}

\usepackage{hyperref}

\journal{Neurocomputing}

\usepackage{times}
\usepackage[utf8]{inputenc} 
\usepackage[T1]{fontenc}    
\usepackage{hyperref}       
\usepackage{url}            
\usepackage{booktabs}       
\usepackage{amsfonts}       
\usepackage{nicefrac}       
\usepackage{microtype}      
\usepackage{bmpsize}
\usepackage{subfigure}
\usepackage{stmaryrd}

\usepackage{multirow}
\usepackage[thinlines]{easytable}
\usepackage{xcolor}
\usepackage{enumitem}
\usepackage{flushend}
\usepackage{dsfont}
\usepackage{textcomp}
\usepackage{units}
\newcommand\captionfont{}
\newcommand\captionlabelfont{}

\usepackage{algorithm}
\usepackage[noend]{algorithmic}
\usepackage{mdframed}
\usepackage[normalem]{ulem}
\usepackage{adjustbox}
\usepackage{xspace}
\usepackage{soul}
\usepackage{wrapfig}
\usepackage{comment}

\usepackage[multiple]{footmisc}

\newtheorem{theorem}{Theorem}
\newtheorem{lemma}[theorem]{Lemma}
\newdefinition{rmk}{Remark}
\newproof{proof}{Proof}
\newproof{pot}{Proof of Theorem \ref{thm2}}

\usepackage{amsmath,amsfonts,bm}

\def\1{\bm{1}}

\DeclareMathAlphabet{\mathsfit}{\encodingdefault}{\sfdefault}{m}{sl}
\SetMathAlphabet{\mathsfit}{bold}{\encodingdefault}{\sfdefault}{bx}{n}

\newcommand{\parag}[1]{\smallskip\noindent\textbf{#1}~~}

\newcommand{\unithyp}{\mathcal{S}_{d-1}}
\newcommand{\dirk}{\mathbf{c}_{k}^{\perp}}
\newcommand{\sqf}{d^{\nicefrac{1}{2}}}
\newcommand{\gradsph}{ \nabla
\mathcal{L}(\bu_k)}
\newcommand{\invf}{d^{\scalebox{0.5}[1.0]{\( - \)}1}}
\newcommand{\invsqf}{d^{\scalebox{0.5}[1.0]{\( - \)}\nicefrac{1}{2}}}
\newcommand{\lrn}[1]{\frac{\eta_k#1}{r_{k}^{2} \invsqf\|\mathbf{b}_k\|}}
\newcommand{\disto}{\frac{\mathbf{b}_k}{\invsqf\|\mathbf{b}_k\|}}

\newcommand{\operS}{{\mathsf{S}}}
\newcommand{\operT}{{\mathsf{T}}}
\newcommand{\operRT}{{\mathsf{RT}}}
\newcommand{\operR}{{\mathsf{R}}}
\newcommand{\mbf}[1]{\ensuremath{\mathbf{#1}}}

\newcommand{\mc}[1]{\ensuremath{\mathcal{#1}}} 

\renewcommand{\vec}[1]{\mathbf{#1}}

\newcommand{\elr}[1]{\eta^{e}_{#1}}
\newcommand{\ckortho}{\mathbf{c}_{k}^{\perp}}
\newcommand{\grad}{{\mathrm{grad}}}
\newcommand{\ckrad}{\langle \mathbf{c}_k, \mathbf{u}_k \rangle}
\newcommand\eqdef{\stackrel{\mathrm{def}}{=}}

\newcommand{\ba}{\mbf{a}}
\newcommand{\bb}{\mbf{b}}
\newcommand{\bc}{\mbf{c}}
\newcommand{\bg}{\mbf{g}}
\newcommand{\bmm}{\mbf{m}}
\newcommand{\bu}{\mbf{u}}
\newcommand{\bv}{\mbf{v}}
\newcommand{\bx}{\mbf{x}}

\makeatletter
\DeclareRobustCommand\onedot{\futurelet\@let@token\@onedot}
\def\@onedot{\ifx\@let@token.\else.\null\fi\xspace}

\def\ie{\emph{i.e}\onedot}

\makeatother

\definecolor{adamsrtcolor}{rgb}{0.74,0.84,0.93}
\newcommand{\adamsrttext}[1]{\colorbox{adamsrtcolor}{#1}}

\newif\ifcompact
\compactfalse

\makeatletter
\def\ps@pprintTitle{%
  \let\@oddhead\@empty
  \let\@evenhead\@empty
  \def\@oddfoot{\reset@font\hfil\thepage\hfil}
  \let\@evenfoot\@oddfoot
}
\makeatother

\begin{document}

\begin{frontmatter}

\title{Spherical Perspective on Learning \\ with Normalization Layers}

\author[1,3]{Simon Roburin\corref{cor1}\fnref{eqcontrib}}
\author[2]{Yann de Mont-Marin\fnref{eqcontrib}}
\author[3]{Andrei Bursuc}
\author[1,3]{Renaud Marlet}
\author[3]{Patrick P\'erez}
\author[1]{Mathieu Aubry}

\cortext[cor1]{Corresponding author.}
\fntext[fn1]{Equal contribution.}


\address[1]{LIGM, Ecole des Ponts, Univ Gustave Eiffel, CNRS, Marne-la-Vall\'ee, France}
\address[2]{D\'epartement d’informatique de l’ENS, PSL, Inria, Paris, France}
\address[3]{valeo.ai, Paris, France}


\begin{abstract}
Normalization Layers (NLs) are  widely used in modern deep-learning architectures. Despite their apparent simplicity, their effect on optimization is not yet fully understood.
This paper introduces a spherical framework to study the optimization of neural networks with NLs from a geometric perspective. Concretely, the radial invariance of groups of parameters, such as filters for convolutional neural networks, allows to translate the optimization steps on the $L_2$ unit hypersphere. This formulation and the associated geometric interpretation shed new light on the training dynamics. Firstly, the first effective learning rate expression of Adam is derived. Then the demonstration that, in the presence of NLs, performing Stochastic Gradient Descent (SGD) alone is actually equivalent to a variant of Adam constrained to the unit hypersphere, stems from the framework. Finally, this analysis outlines phenomena that previous variants of Adam act on and their importance in the optimization process are experimentally validated.
\end{abstract}


\end{frontmatter}


\section{Introduction}
\label{sec:intro}

\begin{figure}[t!]
\renewcommand{\captionfont}{\small}
\centering
\includegraphics[width=0.65\columnwidth]{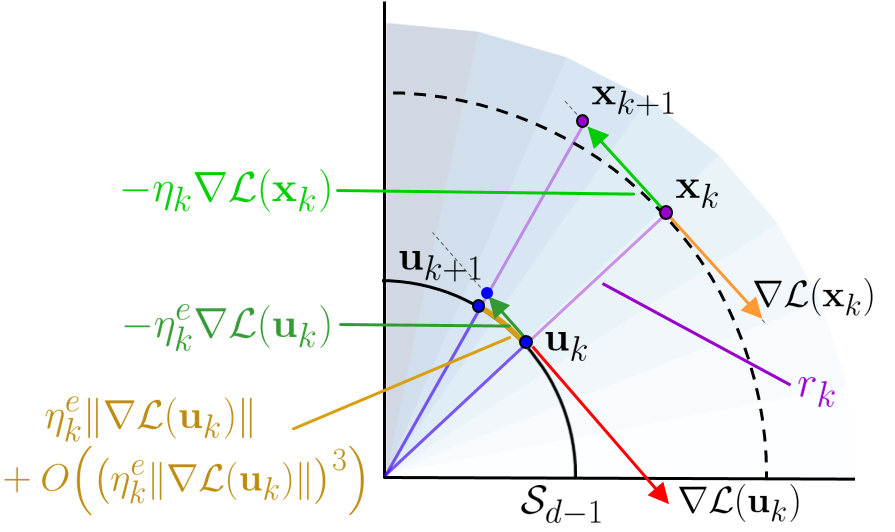}
\caption{\textbf{Illustration of the spherical perspective for SGD.} The loss function $\mathcal{L}$ of a NN w.r.t.\ the parameters $\bx_k \in \mathbb{R}^{d}$ of a neuron followed by a BN is radially invariant. 
The neuron update $\bx_k \to \bx_{k+1}$ in the original space, with velocity $\eta_{k}\nabla \mathcal{L} (\bx_k)$, corresponds to an update $\bu_k \to \bu_{k+1}$ of its projection through an exponential map on the unit hypersphere $\unithyp$ with velocity $\elr{k}\|\nabla \mathcal{L} (\bu_k)\|$  at order 2 (see details in Section~\ref{sec:framework}).}
\vspace{-3.5mm}
\label{fig:diagram_overview}
\end{figure}

The optimization process of deep neural networks is still poorly understood. Their training involves minimizing a high-dimensional non-convex function, which has been proved to be a NP-hard problem~\citep{blum1989training}. 
Yet, elementary gradient-based methods show good results in practice. 
To improve the quality of reached minima, numerous Normalization Layers (NLs) have stemmed in the last years and become common practices. One of the most prominent is Batch Normalization (BN) \citep{ioffe2015batch}, which improves significantly both the 
the training speed and the prediction performance; it has however a notable shortcoming: BN relies heavily on the batch size. To avoid the dependency on the batch size, normalization layers such as LayerNorm (LN)  \cite{ba2016layer}, WeightNorm (WN) \cite{salimans2016weight}, InstanceNorm (IN) \cite{ulyanov2016instance}, or GroupNorm (GN) \cite{wu2018group} were introduced.
Yet, the interaction of NLs with optimization remains an open research topic. \\
Previous studies highlighted some of the mechanisms of the interaction between BN and Stochastic Gradient Descent (SGD), both empirically~\citep{santurkar2018does} and theoretically~\citep{arora2018theoretical, bjorck2018understanding, hoffer2018fix}. But none of them provides a generic framework, nor studies the interaction between NLs and the Adam optimizer~\citep{kingma2014adam}. In this work, we provide an extensive analysis of the relation between NLs and any order-1 optimization scheme. Moreover, we theoretically relate SGD with NLs to a variant of Adam (AdaGradG), which is of interest as Adam probably is the most commonly-used adaptive scheme for Neural Networks (NNs).
A shared effect of all mentioned NLs is to make NNs invariant to positive scalings of groups of parameters. These groups of parameters may differ from one NL method to another.
The core idea of this paper is precisely to focus on these groups of radially-invariant parameters and analyze their optimization projected on the $L_2$ unit hypersphere (see Figure~\ref{fig:diagram_overview}), which is topologically equivalent to the quotient manifold of the parameter space by the scaling action.
In fact, one could directly optimize parameters on the hypersphere as
\cite{cho2017riemannian}. Yet, most optimization methods are still performed successfully in the original parameter space.
Here we propose to study an optimization scheme for a given group of radially-invariant parameters through its image scheme on the unit hypersphere. 
This geometric perspective sheds light on the interaction between normalization layers and Adam.\\
The paper is organized as follows.
In \textbf{Section \ref{sec:background}}, we introduce our spherical framework to study the optimization of any radially-invariant model. We also define a generic optimization scheme that encompasses methods such as SGD with momentum (SGD-M) and Adam. We then derive its image step on the unit hypersphere, leading to definitions and  expressions of \emph{effective learning rate} and \emph{effective learning direction}. These new definitions are explicit and have a clear interpretation, whereas the definition of \cite{van2017l2} is asymptotic and the definitions of \cite{arora2018theoretical} and of \cite{hoffer2018fix} are variational.
In \textbf{Section \ref{sec:SGD}},  we leverage the tools of our spherical framework to demonstrate that, in presence of NLs, SGD is equivalent to AdaGradG, a combination of AdaGrad~\cite{duchi2011adaptive} (a special case of Adam without momentum) and AdamG~\citep{cho2017riemannian} (a variant of Adam constrained to the unit hypersphere). In other words, AdaGradG is a variant of Adam without momentum and constrained to the unit hypersphere. 
In \textbf{Section \ref{sec:diagnosis}}, we analyze the effective learning direction for Adam. The spherical framework highlights phenomena that previous variants of Adam \citep{loshchilov2019decoupled, cho2017riemannian} act on. We perform an empirical study of these phenomena and show that they play a significant role in the training of convolutional neural networks (CNNs).
In \textbf{Section~\ref{sec:relwork}}, these results are put in perspective with related work.
\\
Our main contributions are the following:
\begin{itemize}
\item A framework to analyze and compare order-1 optimization schemes of radially-invariant models;
\item The first explicit expression of the effective learning rate for Adam;
\item The demonstration that, in the presence of NLs, standard SGD is equivalent to AdaGradG, a variant of Adam without momentum and constrained to the unit hypersphere;
\item The identification and the study of geometrical phenomena that occur with Adam and that impact significantly the training of CNNs with NLs.
\end{itemize}

\section{Spherical framework and effective quantities}
\label{sec:background}

In this section, we provide background on radial invariance and introduce a generic optimization scheme that encompasses SGD, SGD with momentum (SGD-M) and Adam, with and without $L_2$-regularization.

Projecting the scheme update on the unit hypersphere leads to the formal definitions of effective learning rate and learning direction for any order-1 optimization scheme. This geometric perspective leads to the first explicit expression of the effective learning rate for Adam.
The main notations are summarized in Figure~\ref{fig:diagram_overview}.

\subsection{Radial invariance}

We consider a family of parametric functions $\phi_{\mbf{x}}\colon \mathbb{R}^{\textit{in}} \rightarrow \mathbb{R}^{out}$
parameterized by a group of radially-invariant parameters $\mathbf{x} \in \mathbb{R}^{d} {\setminus}\, \{ \mathbf{0} \}$, i.e., $\forall \rho\,{>}\,0,\phi_{\rho\mbf{x}} \,{=}\, \phi_{\mbf{x}}$ (possible other parameters of $\phi_{\mbf{x}}$ are omitted for clarity),
a dataset $\mathcal{D} \subset \mathbb{R}^{\textit{in}} \times \mathbb{R}^{\textit{out}}$, a loss function $\ell:  \mathbb{R}^{\textit{out}} \times  \mathbb{R}^{\textit{out}} \to \mathbb{R}$ and a training loss function $\mathcal{L}\colon \mathbb{R}^{d} \to \mathbb{R}$ defined as:
\begin{equation}
    \mathcal{L}(\mbf{x}) \eqdef \frac{1}{|\mathcal{D}|} \sum_{(\mbf{s}, \mbf{t}) \in \mathcal{D}}\ell(\phi_\mbf{x}(\mathbf{s}),\mbf{t}).
\end{equation}
It verifies: $\forall \rho>0,~\mc L (\rho\bx) = \mc L (\bx)$. 
%
In the context of NNs, the group of radially-invariant parameters $\vec{x}$ can be the parameters of a single neuron in a linear layer or the parameters of a whole filter in a convolutional layer, followed by BN. 
See \ref{bn_rescale} for details, and \ref{other_norm_layer} for the application to other normalization schemes such as WeightNorm \cite{salimans2016weight}, InstanceNorm \citep{ulyanov2016instance}, LayerNorm \citep{ba2016layer} or GroupNorm \citep{wu2018group}. 
 
Please note that the loss is in fact evaluated on random batches of $\mathcal{D}$ at each optimization step. However, the study of the impact of splitting $\mathcal{D}$ into batches is out of the scope of this work. For notation purposes, we use $\mathcal{L}$ to actually denote the loss function evaluated on a random batch of the dataset.

The quotient of the parameter space by the equivalence relation associated to radial invariance is topologically equivalent to a sphere. We consider here the $L_2$ sphere
$\mathcal{S}_{d-1}=\{\bu\in\mathbb{R}^d / \|\bu\|_2 = 1\}$ whose canonical metric corresponds to angles: $d_{\mathcal{S}}(\bu_1, \bu_2) = \operatorname{arccos}(\langle \bu_1, \bu_2 \rangle)$.
This choice of metric is relevant to study NNs since filters in CNNs or neurons in MLPs are applied through scalar product to input data. Besides, normalization in NLs is also performed using the $L_2$ norm. Directly in WeightNorm, and indirectly in other NLs because the standard deviation can be interpreted as a centered $L_2$ norm.

Our framework relies on the decomposition of vectors into radial and tangential components. During optimization, we write the radially-invariant parameters at step $k \geq 0$ as $\bx_k = r_k \bu_k$ where $r_{k}=\left\|\bx_{k}\right\|$ and $\bu_{k}=\relax{\bx_{k}}/{\left\|\bx_{k}\right\|}$. For any quantity $\mbf{q}_k \in \mathbb{R}^d$ at step $k$, we write $\mbf{q}^{\perp}_k =\mbf{q}_k {-} \langle \mbf{q}_k, \bu_{k} \rangle \bu_{k}$ its tangential component relatively to the current direction $\bu_k$. 

The following lemma states that the gradient of a radially-invariant loss function is tangential and $-1$ homogeneous:
\begin{lemma}[Gradient of a function with radial invariance]\label{lem_invar}
If $ \mathcal{L}\colon \mathbb{R}^{d} \to \mathbb{R}$ is radially invariant and almost everywhere differentiable, then, for all $\rho>0$ and all $\bx \in \mathbb{R}^{d}$ where $\cal L$ is differentiable: 
\begin{align}
& \langle \nabla \mathcal{L}(\bx), \bx \rangle = 0 \quad \textnormal{and} \quad
\nabla \mathcal{L}(\bx) = \rho\, \nabla \mathcal{L}(\rho\bx) \label{rad_invar}.
\end{align}
\end{lemma}

\subsection{Generic optimization scheme}

There is a large body of literature 
on optimization schemes~\citep{sutskever13,duchi2011adaptive,tieleman2012lecture,kingma2014adam, loshchilov2019decoupled}. We focus here on two of the most popular ones, namely SGD and Adam \citep{kingma2014adam}. Yet, to establish general results that may apply to a variety of other schemes, we introduce here a \emph{generic optimization update}:
\begin{align}
     \bx_{k+1} & = \bx_{k} - \eta_{k} \ba_{k} \oslash\bb_{k}, \label{eq1a} \\
     \ba_k & = \beta\ba_{k-1} + \nabla \mathcal{L}(\bx_{k}) + \lambda \bx_{k}, \label{eq1b} 
\end{align}
where $\bx_k \in \mathbb{R}^d$ is the group of radially-invariant parameters at iteration $k$, $\mathcal{L}$ is the group's loss estimated on a batch of input data, $\ba_k \in \mathbb{R}^d$ is a momentum, $\bb_k \in \mathbb{R}^d$ is a division vector that can depend on the trajectory $\left(\bx_i, \nabla \mathcal{L}(\bx_i)\right)_{i\in  \llbracket  0,k  \rrbracket}$, $\eta_k \in \mathbb{R}$ is the scheduled trajectory-independent learning rate, $\oslash$ denotes the Hadamard element-wise division, $\beta$ is the momentum parameter, and $\lambda$
is the $L_2$-regularization parameter. We show how it encompasses several known optimization schemes.

{\it Stochastic gradient descent (SGD)}\label{SGD} has proven to be an effective optimization method in deep learning.
It can include $L_2$ regularization (also called weight decay) and momentum. Its updates are:
\begin{align}
     \bx_{k+1} & = \bx_{k} - \eta_{k}\bmm_{k}, \\
     \bmm_{k} & = \beta \bmm_{k-1} + \nabla \mathcal{L}(\bx_{k}) + \lambda \bx_{k},
\end{align}
where $\bmm_{k}$ is the momentum, $\beta$ is the momentum parameter, and $\lambda$ is the $L_2$-regularization parameter. It corresponds to our generic scheme (Eqs.~\ref{eq1a}-\ref{eq1b}) with $\ba_k =\bmm_k$ and $\mbf b_k= [1\cdots 1]^\top$.
    
{\it Adam} 
is likely the most common adaptive scheme for NNs. Its updates are:

\begin{align}
    \bx_{k+1} & 
    = \bx_{k} - \eta_{k}  \frac{\bmm_k}{1 - \beta_1^{k + 1}} \oslash \sqrt{\frac{\bv_k}{1 - \beta_2^{k + 1}} + \epsilon}, \label{preq1}
    \\ 
    \bmm_k &
    = \beta_1 \bmm_{k-1}\! + \! (1 - \beta_1)(\nabla\mathcal{L}(\bx_{k}) + \lambda \bx_k),
    \\
    \bv_k &
    = \beta_2 \bv_{k-1} + (1- \beta_2)(\nabla \mathcal{L}(\bx_{k}) + \lambda \bx_k)^2\!\!, \label{preq2}
\end{align}

where $\bmm_{k}$ is the momentum with parameter $\beta_1$, $\bv_{k}$ is the second-order moment with parameter $\beta_2$, and $\epsilon$ prevents division by zero. (Here and in the following, the square and the square root of a vector are to be understood as element-wise.) It corresponds to our generic scheme (Eqs.~\ref{eq1a}-\ref{eq1b}) with $\beta{=} \beta_1$ and:
%
\begin{align}
    \ba_{k} &
    = \frac{\bmm_k}{1 - \beta_1}, 
    \\
    \bb_k & 
    = \frac{1 - \beta_1^{k+1}}{1 - \beta_1}\sqrt{\frac{\bv_k}{1 - \beta_2^{k + 1}} + \epsilon}.\label{adam_algo}
\end{align}

\subsection{Image optimization on the hypersphere}\label{sec:framework}
The radial invariance implies that the radial part of the parameter update $\mbf{x}$ does not change the function $\phi_{\mbf{x}}$ encoded by the model, nor does it change the loss $\mathcal{L(\mbf{x})}$. Due to radial invariance, the parameter space projected on the unit hypersphere is topologically closer to the functional space of the network than the full parameter space. It hints that looking at optimization behaviour on the unit hypersphere might be interesting. 
To achieve this, we separate the quantities that can (tangential part) and cannot (radial part) change the model function. Theorem~\ref{thm:sphere_step} formulates the spherical decomposition of our generic optimization scheme (Eqs.~\ref{eq1a}-\ref{eq1b}) in simple terms. It relates the update of radially-invariant parameters in the parameter space $\mathbb{R}^d$ and their update on $\unithyp$ through an exponential map.

\begin{theorem}[Image step on $\mathcal{S}_{d{-}1}$]\label{thm:sphere_step}
Let's consider the update of a group of radially-invariant parameters $\bx_k$ 
at step $k$ following the generic optimization scheme (Eqs.~\ref{eq1a}-\ref{eq1b}) and the corresponding update of its projection $\bu_k$ on $\unithyp$. Under the hypothesis (H1) and (H2) formalized in \ref{calculus_32}, the update of $\bu_k$ is given by an exponential map at $\bu_k$ with velocity $\elr{k}\dirk$:
\begin{equation}
    \bu_{k+1} =	 \operatorname{Exp}_{\bu_{k}}\left(-\left[1 + O\left(\|\elr{k}\dirk\|^2\right)\right]\elr{k}\dirk\right),\label{eq:1_th}
\end{equation}

where $\operatorname{Exp}_{\bu_{k}}$ is the exponential map on $\unithyp$, and with 
\begin{align} 
\bc_k &
\eqdef r_k\ba_k \oslash \disto,
\\
\elr{k} &
\eqdef \lrn{} \left(1 - \lrn{\langle \bc_k, \bu_k \rangle} \right)^{-1}\label{lredef}.
\end{align}

More precisely:
\begin{align}
\bu_{k+1} &= \frac{\bu_k - \elr{k}\dirk}{\sqrt{1 + (\elr{k}\| \dirk \|)^{2}}}.\label{canonic}
\end{align}
\end{theorem}

The proof is given in \ref{calculus_32} and the theorem is illustrated in the case of SGD in Figure~\ref{fig:diagram_overview}. Note that for CNN training the hypothesis (H1) and (H2), empirically discussed in the appendix are typically verified. In particular, with typical values $1{-}\lrn{\langle \bc_k, \bu_k \rangle}> 0$ (H1) is true. The other hypothesis $\elr{k} \|\ckortho\|< \pi$ (H2) where steps are supposed shorter than $\pi$ is also true (see \ref{positivity}).

\subsection{Effective quantities}


In Theorem~\ref{thm:sphere_step}, the normalized parameters update in Eq.\,\ref{eq:1_th} can be read $\bu_{k+1} \approx	 \operatorname{Exp}_{\bu_{k}}\left(-\elr{k}\dirk\right)$, where $\elr{k}$ and $\dirk$ can then be respectively interpreted as the learning rate and the direction of an optimization step constrained to $\unithyp$. Since $\ba_k$ is the momentum and, with Lemma~\ref{lem_invar}, the quantity $r_k\ba_k$ in $\bc_k$ can be seen as \emph{a momentum on the hypersphere}.
Due to the radial invariance, only the change of parameter on the unit hypersphere corresponds to a change of model function.
Hence we can interpret $\elr{k}$ and $\dirk$ as \emph{effective learning rate} and \emph{effective learning direction}. In other words, these quantities correspond to the learning rate and direction on the hypersphere that reproduce the function update of the optimization step.

\begin{table}
\centering
\renewcommand{\arraystretch}{1.5}
\renewcommand{\captionlabelfont}{\bf}
\renewcommand{\captionfont}{\small}
\caption{Effective learning rate and direction for optimization schemes with $\nu = r\protect\invsqf\|\bb\|$ (we omit here the iteration index~$k$).}
\label{eff_lr_scheme}
\scalebox{1}
{
\begin{tabular}{lccr}
\toprule
Scheme     & $\elr{}$ & $\bc^{\perp}$ \\
\midrule
SGD        &       $\frac{\eta}{r^{2}}$       &  $\nabla \mathcal{L}(\bu)$    \\
SGD + $L_2$   &   $\frac{\eta}{r^{2}(1-\eta\lambda)}$           &     $\nabla \mathcal{L}(\bu)$ \\
SGD-M    &       $\frac{\eta}{r^2} \Large(1 - \frac{\eta\langle \bc, \bu \rangle}{r^2} \Large)^{-1} $       & $\bc^{\perp}$    \\
\midrule
Adam & $\frac{\eta}{r\nu} \Large(1 - \frac{\eta\langle \bc, \bu \rangle}{r\nu} \Large)^{-1}$  & $\bc^{\perp}$\\
\bottomrule
\end{tabular}
}
\end{table}
Using Theorem~\ref{thm:sphere_step}, we can derive actual effective learning rates for any optimization scheme that fits our generic framework. These expressions, summarized in Table~\ref{eff_lr_scheme} are explicit and have a clear interpretation, 
in contrast to learning rates in~\citep{van2017l2}, which are approximate and asymptotic, and in~\citep{hoffer2018norm, arora2018theoretical}, which are variational and restricted to SGD without momentum only.

In particular, we provide the first explicit expression of the effective learning rate for Adam:
\begin{equation}
\elr{k} = \frac{\eta_k}{r_k\nu_k} \left(1 - \frac{\eta_k\langle \bc_k, \bu_k \rangle}{r_k\nu_k} \right)^{-1}
\end{equation}
where $\nu_k = r_k\invsqf\|\bb_k\|$ is homogeneous to the norm of a gradient on the hypersphere and can be related to an \emph{second-order moment on the hypersphere} (see \ref{order2adam} for details). Using the variable $\nu$ also simplifies the in-depth analysis in Section~\ref{sec:diagnosis}, allowing a better interpretation of formulas.

The expression of the effective learning rate of Adam, i.e., the amplitude of the step taken on the hypersphere, reveals a dependence on the dimension $d$ (through $\nu$) of the update of the considered group of radially-invariant parameters. In the case of an MLP or CNN that stacks layers with neurons or filters of different dimensions, the learning rate is thus tuned differently from one layer to another. 

We can also see that for all schemes the learning rate is tuned by the dynamics of radiuses $r_k$, which follow:
\begin{equation}
    \frac{r_{k+1}}{r_k} \,{=}\, \left(1 \,{-}\, \lrn{\langle \bc_k, \bu_k \rangle} \right)\sqrt{1 + (\elr{k}\| \dirk \|)^{2}}. \label{rdynamic}    
\end{equation}
In contrast to previous studies
\citep{arora2018theoretical,van2017l2}, this result demonstrates that for momentum methods, $\langle \bc_k,\bu_k \rangle$, which involves accumulated gradients terms in the momentum as well as $L_2$ regularization, tunes the learning rate.

\section{SGD is equivalent to AdaGradG}
\label{sec:SGD}

In this section, we leverage the tools introduced in the spherical framework of Section~\ref{sec:background} to find a scheme constrained to the hypersphere that is equivalent to SGD. We show that, for radially-invariant models, SGD is actually an adaptive optimization method. Formally, SGD is equivalent to a special case of AdamG~\citep{cho2017riemannian} without momentum, where AdamG is a variant of Adam adapted and constrained to the unit hypersphere. Alternatively, we can also say that SGD is equivalent to a variant of AdaGrad adapted and constrained to the hypersphere, where AdaGrad is a special case of Adam without momentum. Besides, we illustrate this theoretical equivalence empirically.

\subsection{Equivalence between two optimization schemes}
Due to the radial invariance, the functional space of the model is encoded by $\unithyp$. In other words, two schemes with the same sequence of groups of radially-invariant parameters on the hypersphere $\left(\mbf{u}_k\right)_{k \geq 0}$ encode the same sequence of model functions. We say that two optimization schemes $S$ and $\Tilde{S}$ are equivalent if and only if $\forall k \geq 0, \bu_k = \Tilde{\bu}_k$. Hence, starting from the same parameters, they reach the same optimum.
By using Eq.\,\ref{canonic}, we obtain the following lemma, which is useful to prove the equivalence of two given optimization schemes:

\begin{lemma}[Sufficient condition for the equivalence of optimization schemes]\label{lm:eq_scheme}
\begin{equation}
     \left\{
    \begin{array}{l}
        \bu_0 = \Tilde{\bu}_0 \\
        \forall k \geq 0, \eta^e_k = \Tilde{\eta}^e_k, \bc^{\perp}_k = \Tilde{\bc}^{\perp}_k
    \end{array}
\right.\Rightarrow 
    \forall k \geq 0, \bu_k = \Tilde{\bu}_k. 
\end{equation}
\end{lemma}


\subsection{A hypersphere-constrained scheme equivalent to SGD}
\label{eff_lr_sgd}

We now study, within our spherical framework, SGD with $L_2$ regularization, i.e., the update $\bx_{k+1} = \bx_k - \eta_k (\nabla\mathcal{L}(\bx_k) - \lambda_k\bx_k)$.
From the effective learning rate expression, we know that SGD yields an adaptive behaviour because it is scheduled by the radius dynamic, which depends on gradients. In fact, the tools in our framework allow us to find that SGD is equivalent to a variant of Adam constrained to the unit hypersphere, similar to AdamG \citep{cho2017riemannian}, and without momentum, similar to AdaGrad. AdamG \citep{cho2017riemannian} uses the same updates as Adam (eq.~\ref{preq1}-\ref{preq2}) but project the weight on the hyper-sphere after each optimization step. More precisely, SGD is equivalent to AdamG with a null momentum factor $\beta_1 = 0$ (like AdaGrad), a non-null initial second-order moment $v_0$, an offset of the scalar second-order moment $k+1 \to k$ and without the bias correction term $1 - \beta_2^{k+1}$. Dubbed AdaGradG, this scheme reads:
\begin{equation*}
    \textnormal{(AdaGradG) : }
    \left\{
        \begin{array}{l}
            \hat{\bx}_{k+1} = \bx_k - \eta_k \frac{\nabla\mathcal{L}(\bx_k)}{\sqrt{v_k}},\\
            \bx_{k+1} = \frac{\hat{\bx}_{k+1}}{\|\hat{\bx}_{k+1}\|}, \\
            v_{k+1} = \beta v_k + \|\nabla\mathcal{L}(\bx_k)\|^2.
        \end{array}
    \right.
\end{equation*}

AdaGradG, like AdamG, is an adaptive method. Unlike Adam, which is adaptive with respect to the second-order moment for each parameter, AdaGradG and AdamG are adaptive for each group of radially-invariant parameters (e.g., filters for CNNs with BN or WN). In other words, each filter is adapted individually and independently by the optimization algorithm; it is not a global scheduling.

Now if we call « equivalent at order 2 in the step » a scheme equivalence that holds when we use for $r_k$ an expression that satisfies the radius dynamic with a Taylor expansion at order 2, then we have the following theorem:
\begin{theorem}[SGD equivalent scheme on the unit hypersphere]\label{th:eq_sgd}
For any $\lambda\geq0, \eta>0, r_0>0$, we have the following equivalence when using the radius dynamic at order 2 in $\relax{(\eta_k\|\nabla\mathcal{L}(\mathbf{u}_k)\|)^2}/{r_k^2}$:
\begin{equation*}
    \left\{
    \begin{array}{l}
        \textnormal{(SGD)}\\
        \bx_0 = r_0 \bu_0 \\
        \lambda_k=\lambda \\
        \eta_k=\eta
    \end{array}
    \right.
    \textnormal{is scheme-equivalent at order 2 in step with}
    \left\{
    \begin{array}{l}
        \textnormal{(AdaGradG)}\\
        \bx_0 = \bu_0 \\
        \beta = (1 - \eta\lambda)^4 \\
        \eta_k=(2\beta)^{-1/2} \\
        v_0 = r_0^4 (2\eta^{2}\beta^{1/2})^{-1}.
    \end{array}
    \right.
\end{equation*}
\end{theorem}

\paragraph{Sketch of proof.}
Starting from SGD, we first use  Lemma~\ref{lm:eq_scheme} to find a strict equivalence scheme with a simpler radius dynamic. We resolve this radius dynamic with a Taylor expansion at order 2 in $\relax{(\eta_k\|\nabla\mathcal{L}(\mathbf{u}_k)\|)^2}/{r_k^2}$. A second use of Lemma~\ref{lm:eq_scheme} finally leads to the scheme equivalence in Theorem~\ref{th:eq_sgd}. The formal complete proof can be found in \ref{sgd_proof}. The fact that $r_k$ is well approximated at order~2 in practice is illustrated in  \ref{tayler_exp_res} and Figure~\ref{fig:taylor}.
\\

This result is unexpected because SGD, which is not adaptive by itself, is equivalent to a second order moment adaptive method.
The scheduling performed by the radius dynamics actually replicates the effect of dividing the 
learning rate by the second-order moment of the gradient norm: $v_k$.

For standard values of hyperparameters $\lambda < 1$ (order of magnitude of $10^{-4}$) and $\eta < 1$ (order of magnitude at most $10^{-1}$), the higher-order terms of the radius in the Taylor expansion empirically become negligible in practice.

Second, with standard values of the hyper-parameters, namely learning rate $\eta < 1$ and $L_2$ regularization $\lambda <1$, we have $\beta \leq 1$ which corresponds to a standard value for a moment factor. Interestingly, the $L_2$ regularization parameter $\lambda$ controls the memory of the past gradients' norm. If $\beta=1$ (with $\lambda=0$), there is no attenuation, each gradient norm has the same contribution in the order-2 moment. If $\lambda \neq 0$, there is a decay factor ($\beta < 1$) on past gradients' norm in the order-2 moment. This gives a new interpretation of the role of the $L_2$ regularization parameter $\lambda$ in SGD with NLs.

\subsection{Empirical validation}

In order to illustrate the equivalence in Theorem~\ref{th:eq_sgd}, we experiment with learning an image classifier on CIFAR10 using different optimization schemes. We consider two architectures: ResNet20 with BN and ResNet20 with WN. 

The set of parameters $\boldsymbol{\theta}$ of the above architectures can be split into two disjoint subsets:
$\boldsymbol{\theta} = \mathcal{F} \cup \mathcal{R}$,
where $\mathcal{F}$ is the set of groups of radially-invariant parameters and $\mathcal{R}$ is the set of remaining parameters.
Note that the set of remaining parameters in $\mathcal{R}$ differs from one architecture to another: for ResNet20 BN, it includes the last linear layer as well as the scaling and bias of BN layers; for ResNet20 WN, it includes the magnitude parameters in each convolutional layer as well as the last linear layer.

For each architecture, we experiment with tracking the trajectory of parameters under different optimization schemes: SGD, AdaGradG and AdaGrad. As our analysis is restricted to radially-invariant parameters, we only track the trajectory of parameters belonging to $\mathcal{F}$, while the remaining parameters, belonging to $\mathcal{R}$, are always optimized in the same way, i.e., with SGD.
For stability purposes, we finetune a previously trained architecture with SGD. The order of batches as well as the random seed for data augmentation are fixed to obtain comparable trajectories. The hyperparameters are chosen for SGD and AdaGrad so that gradient steps have the same order of magnitude (see \ref{track_details} for details); the hyperparameters for AdaGradG are provided by the equivalence in Theorem~\ref{th:eq_sgd}.

\begin{figure}[t]
    \renewcommand{\captionfont}{\small}
    \renewcommand{\captionlabelfont}{\bf}
    \centering
    \hspace{-7mm}
    \subfigure[ResNet20 BN]{\mbox{}\hspace*{5mm}\includegraphics[scale=0.25]{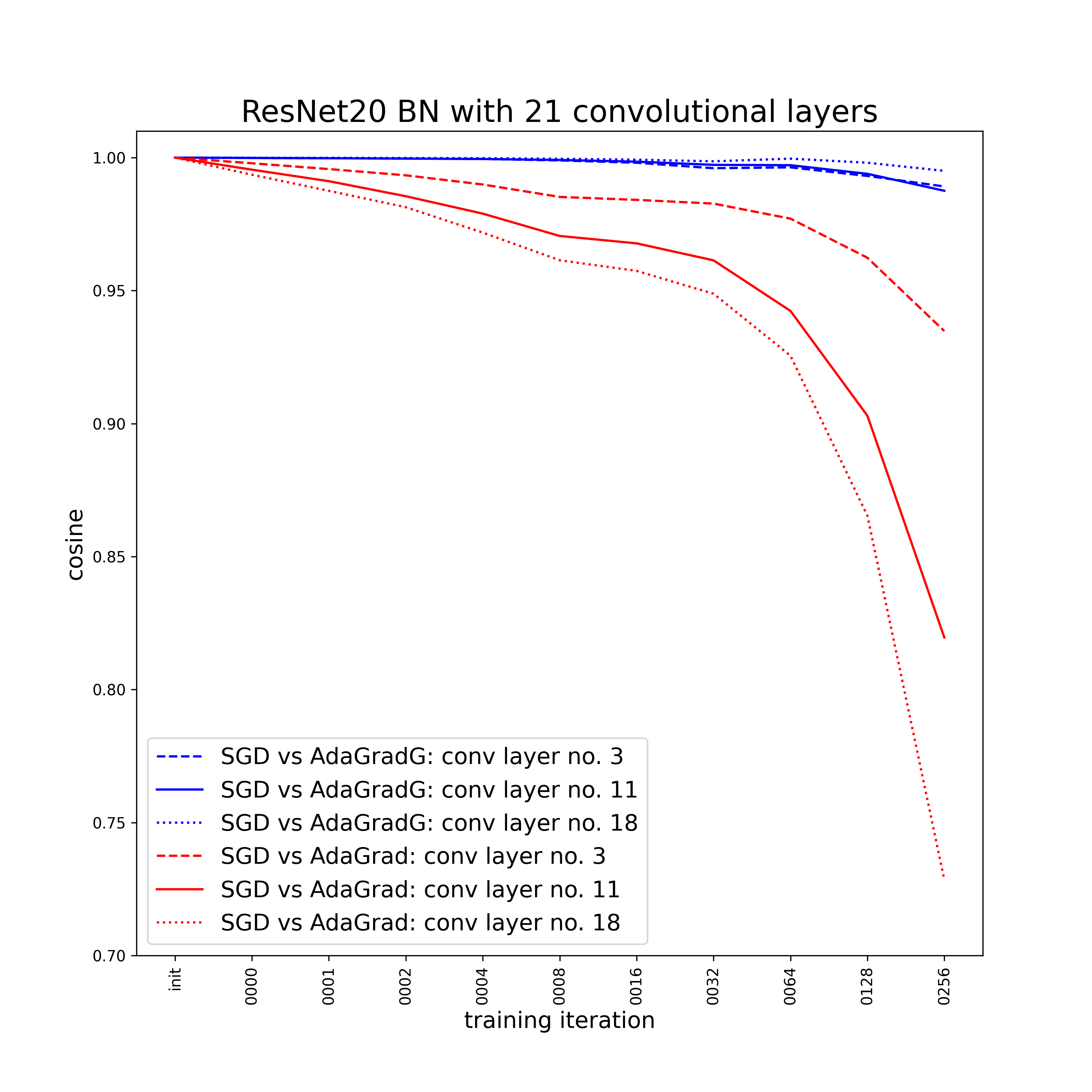}}
    \subfigure[ResNet20 WN]{\hspace{3mm}\includegraphics[scale=0.25]{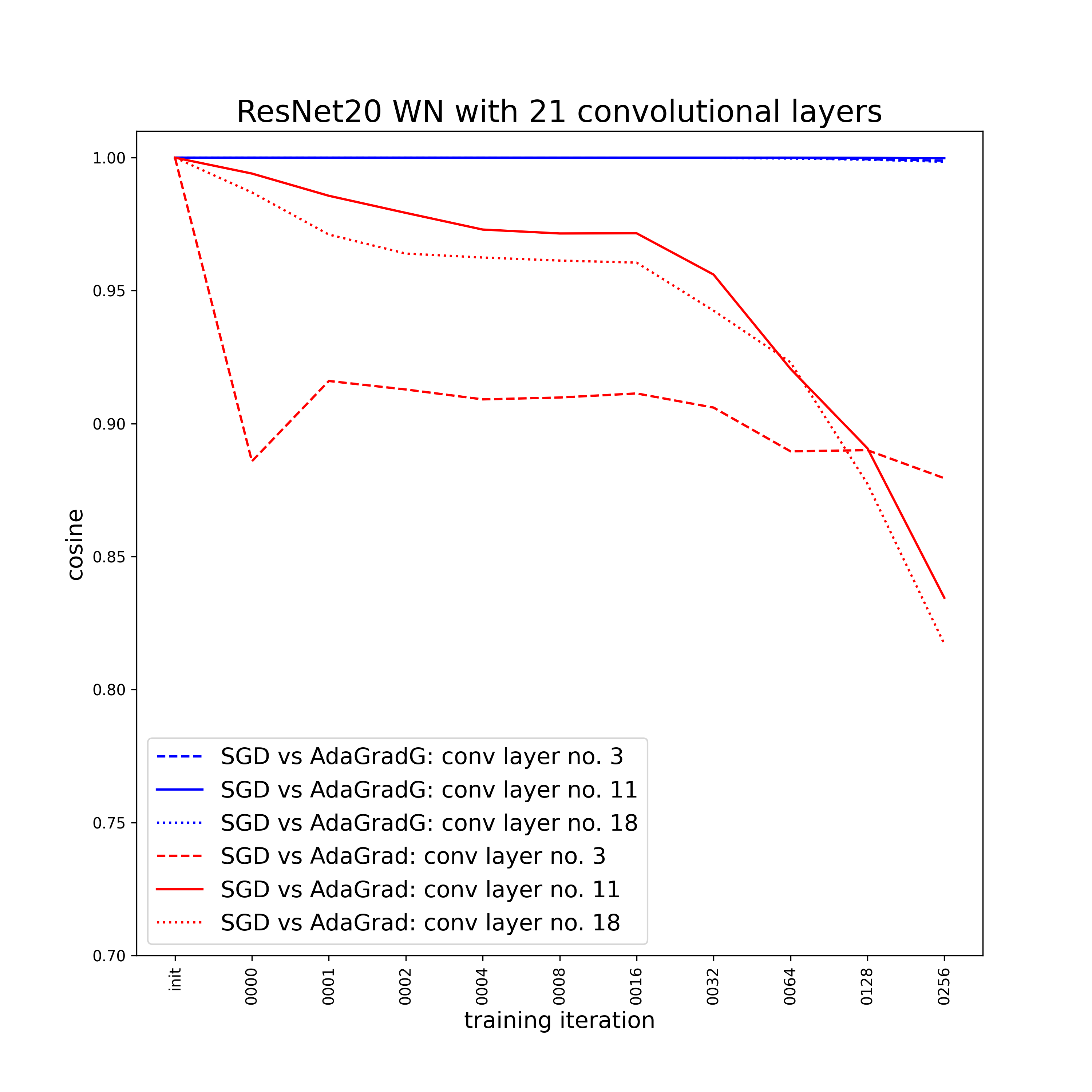}\hspace{-2mm}}
    \caption{\textbf{Comparison of the trajectories of radially-invariant parameters using different optimization schemes}. For three randomly selected filters in each block of a ResNet20 architecture, with BN (left) or WN (right), we compute the cosine similarity between the parameter values obtained with SGD and the parameters values obtained respectively by AdaGradG and AdaGrad, at different iteration stages of a classification training on CIFAR10.}
    \label{fig:track_adamgradg_sgd}
\end{figure}

In Figure~\ref{fig:track_adamgradg_sgd}, we show the angle between the training trajectories using SGD and AdaGradG (resp.\ SGD and AdaGrad), for three different filters in each block of the ResNet20 architectures. We observe that the trajectories on the $L_2$ unit hypersphere remain aligned for SGD and AdaGradG whereas, for SGD and AdaGrad, they quickly diverge. It empirically validates the equivalence mentioned in Theorem~\ref{th:eq_sgd}.

\section{Geometric phenomena in Adam}
\label{sec:diagnosis}


Our framework with its geometrical interpretation reveals intriguing 
behaviors occurring in Adam. 
Indeed, since the unit hypersphere is enough to represent the functional space encoded by the network, from the perspective of manifold optimization, the optimization direction should only depend on the trajectory on that manifold.
In the case of Adam, the effective direction not only depends on the trajectory on the hypersphere but also on the deformed gradients and additional radial terms. These terms are thus likely to play a role in Adam optimization.

In order to understand their role, we describe these geometrical phenomena in Section~\ref{subsec:phenomena}.
Interestingly, previous variants of Adam,  AdamW~\citep{loshchilov2019decoupled} and AdamG~\citep{cho2017riemannian} are related to these phenomena.
To study empirically their importance, 
we consider in Section~\ref{subsec:empirical} variants of Adam that first provide a direction intrinsic to the unit hypersphere, without deformation of the gradients, and then where radial terms are decoupled from the direction.
The empirical study of these variants over a variety of datasets and architectures suggests that these 
behaviors do play a significant role in CNNs training with BN.

\subsection{Identification of geometrical phenomena in Adam}\label{subsec:phenomena}
Here, we perform an in-depth analysis of the effective learning direction of Adam.

\begin{figure}[t]
    \renewcommand{\captionfont}{\small}
    \renewcommand{\captionlabelfont}{\bf}
    \centering
    \hspace{-7mm}
    \subfigure[Radial scheduling]{\mbox{}\hspace*{5mm}\includegraphics[scale=0.23]{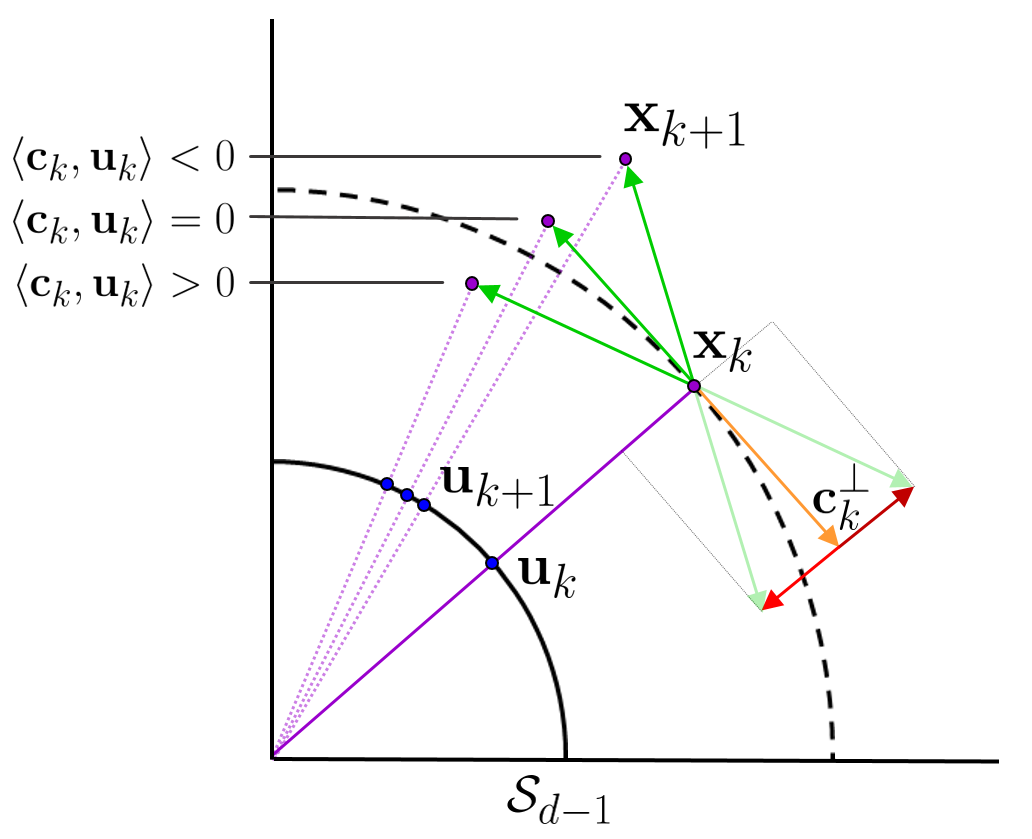}}
    \subfigure[Deformation anisotropy]{
    \includegraphics[scale=0.23]{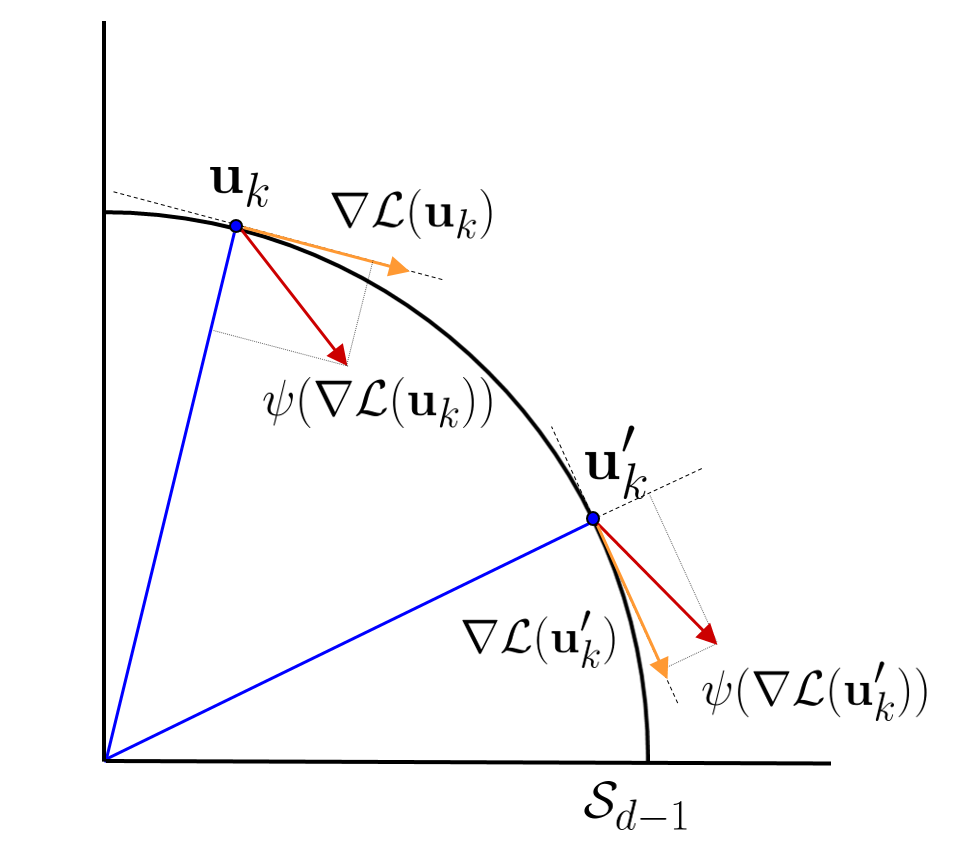}}
    \subfigure[Gradient-history contribution]{\hspace{3mm}\includegraphics[scale=0.29]{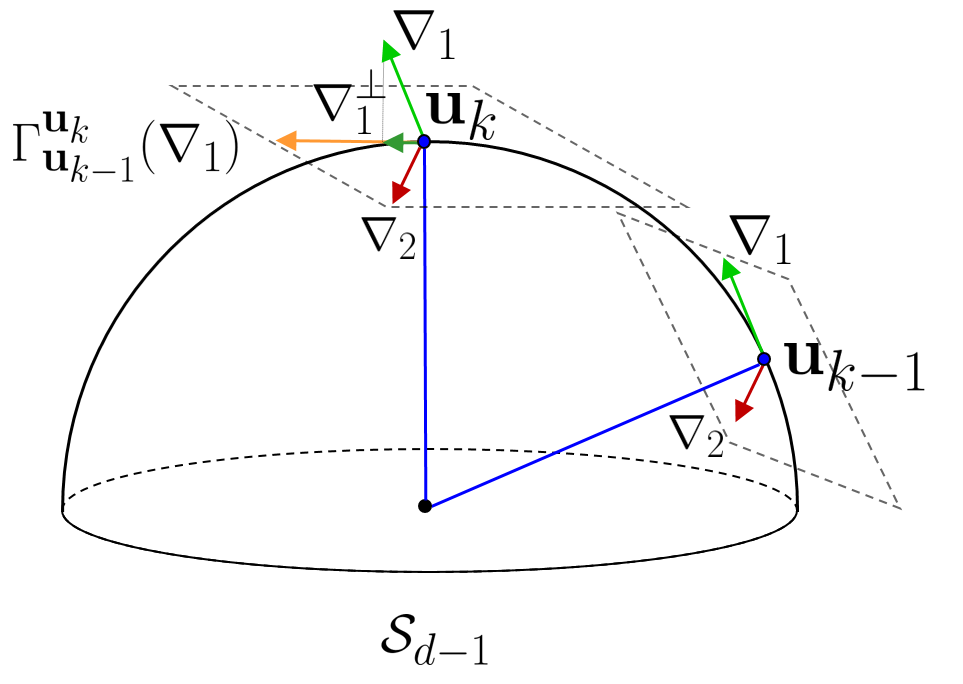}\hspace{-2mm}}
    \caption{
    (a)~Effect of the radial part of $\mathbf{c}_k$ on the displacement on $\unithyp$; 
    (b)~Example of anisotropy and sign instability for the deformation $\psi(\nabla\mathcal{L}(\bu_k)) = \nabla\mathcal{L}(\bu_k)\oslash\frac{|\nabla\mathcal{L}(\bu_k)|}{d^{-1/2}\|\nabla\mathcal{L}(\bu_k)\|}$ (where $|\cdot|$ is the element-wise absolute value) occurring in Adam's first optimization step; 
    (c)~Different contribution in $\ckortho$ of two past gradients $\nabla_1$ and $\nabla_2$ of equal norm, depending on their orientation. Illustration of the transport of $\nabla_1$ from $\bu_{k-1}$ to $\bu_k$ : $\Gamma_{\bu_{k-1}}^{\bu_{k}}(\nabla_1)$ (\textit{cf}. \ref{rt_precision} for details).}
    \label{fig:impact_radial_main}
\end{figure}


\hspace*{1em}
\textbf{(a) Deformed gradients.} Considering the quantities defined for a generic scheme in Eq.\,\ref{lredef}, $\bb_k$ has a deformation effect on $\ba_k$, due to the Hadamard division by $\disto$, and a scheduling effect $\invsqf  \| \bb_k \|$ on the effective learning rate.
In the case where the momentum factor is null $\beta_1=0$, the direction of the update at step $k$ is $\nabla \mathcal{L}(\bu_k)\oslash\disto$ (Eq.\,\ref{lredef}) and the deformation $\disto$  may push the direction of the update outside the tangent space of $\unithyp$ at $\bu_k$, whereas the gradient itself lies in the tangent space.
This deformation is in fact not isotropic: the displacement of the gradient from the tangent space depends on the position of $\bu_k$ on the sphere. We illustrate this anisotropy in Figure~\ref{fig:impact_radial_main}(b). 
\\
\hspace*{1em}
\textbf{(b) Additional radial terms.} In the momentum on the sphere $\bc_k$, quantities that are radial (resp.\ orthogonal) at a point on the sphere may not be radial (resp.\ orthogonal) at another point. To clarify the contribution of $\bc_k$ in the effective learning direction $\ckortho$, we perform the following decomposition (\textit{cf}.\ \ref{ck_decomp}):
\smallskip
\vspace*{-5mm}
\begin{align}
\bc_k &= (\bc_k^{\grad} + \lambda r_k^2\bc_k^{L_2})\oslash \disto \textnormal{\ \ \ with:} \label{eq:ckdecomp}\\
\bc_k^{\grad} &\eqdef  \nabla \mathcal{L}(\bu_k) + \sum_{i=0}^{k-1}\beta^{k-i}\frac{r_k}{r_i}\nabla\mathcal{L}(\bu_i)
\\
\bc_k^{L_2} &\eqdef \bu_k + \sum_{i=0}^{k-1}\beta^{k-i}\frac{r_i}{r_k}\bu_i.\label{eq:ckL2}
\end{align}
\\
\hspace*{2em}
\textbf{b1.\,\,\,Contribution of $\bc_{k}^{\grad}$}. At step $k$, the contribution of each past gradient corresponds to the orthogonal part $\nabla\mathcal{L}(\bu_i) - \langle \nabla\mathcal{L}(\bu_i), \bu_k \rangle \bu_k$. It impacts the effective learning direction depending on its orientation relatively to $\bu_k$.
Two past points, although equally distant from $\bu_k$ on the sphere and with equal gradient amplitude may thus contribute 
differently in $\ckortho$ due to their orientation (\textit{cf}.\ Figure~\ref{fig:impact_radial_main}(c)).\\
\hspace*{2em}
\textbf{b2.\,\,\,Contribution of $\bc_{k}^{L_2}$}. Naturally, the current point $\bu_k$ does not contribute to the effective learning direction $\ckortho$, unlike the history of points in $\sum_{i=0}^{k-1}\beta^{k-i}\frac{r_i}{r_k}\bu_i$, which does. This dependency can be avoided if we decouple the $L_2$ regularization,
in which case we do not accumulate $L_2$ terms in the momentum. This shows that the decoupling proposed in AdamW~\citep{loshchilov2019decoupled} actually removes the contribution of $L_2$ regularization in the effective learning direction.
\\
\hspace*{1em}
\textbf{(c) The radius ratio $\frac{r_k}{r_i}$} present in both $\bc_k^{\grad}$ and $\bc_k^{L_2}$ (in inverse proportion) impacts the effective learning direction $\dirk$: it can differ for identical sequences $(\bu_i)_{i\leq k}$ on the sphere but with distinct radius histories  $(r_i)_{i\leq k}$.  Since the radius is closely related to the effective learning rate, it means that the effective learning direction $\ckortho$ is adjusted according to the learning rates history.

Note that AdamG \citep{cho2017riemannian}, by constraining the optimization to the unit hypersphere and thus removing $L_2$ regularization, neutralizes all the above phenomena. However, this method has no scheduling effect allowed by the radius dynamics (\textit{cf}. Eq.\ref{rdynamic}) since it is kept constant during training.

\subsection{Empirical study}\label{subsec:empirical}
To study empirically the importance of the identified geometric phenomena, we perform an ablation study: we compare the performance (accuracy and training loss speed) of Adam and variants that neutralize each of them. 
We recall that AdamW neutralizes \textbf{(b2)} and that AdamG neutralizes all of above phenomena but loses the scheduling effect identified in Eq.\,\ref{rdynamic}. To complete our analysis, we use geometrical tools to design variations of Adam which neutralizes sequentially each phenomenon while preserving the natural scheduling effect in Theorem~\ref{thm:sphere_step}. We neutralize \textbf{(a)} by replacing the element-wise second-order moment, \textbf{(b1)} and \textbf{(b2)} by transporting the momentum from a current point to the new one, \textbf{(c)} by re-scaling the momentum at step $k$. The details are in \ref{proof:groupwise}. The final scheme reads:
%
\begin{align}
    \bx_{k+1} &= \bx_{k} - \eta_{k}  \frac{\bmm_k}{1 - \beta_1^{k + 1}} \bigg/ \sqrt{\frac{v_k}{1 - \beta_2^{k + 1}} + \epsilon}, \label{preq1SRT} 
    \\
    \bmm_k &= \beta_1 \frac{r_{k-1}}{r_k}\Gamma_{\mathbf{u}_{k-1}}^{\mathbf{u}_{k}}(\bmm_{k-1})\! + \! (1 - \beta_1)(\nabla\mathcal{L}(\bx_{k}) + \lambda \bx_k),
    \\
    v_k &= \beta_2 \frac{r_{k-1}^2}{r_k^2}v_{k-1} + (1- \beta_2)\invf\|\nabla \mathcal{L}(\bx_{k}) + \lambda \bx_k\|^2, \label{preq2SRT}
\end{align}
where $\Gamma_{\mathbf{u}_{k-1}}^{\bu_{k}}$ is the hypersphere canonical parallel transport from $\bu_{k-1}$ to $\bu_{k}$. Implementation details are in \ref{training_implem_details}.

\textbf{Protocol.} 
For evaluation, we conduct experiments on two architectures: VGG16~\citep{simonyan2014very} 
and 
ResNet~\citep{he2016deep} -- more precisely ResNet20, a simple variant designed for small images~\citep{he2016deep}, and ResNet18, a popular variant for image classification. We consider three datasets: SVHN~\citep{netzer2011reading}, CIFAR10 and CIFAR100~\citep{krizhevsky2009learning}.

Since our goal is to evaluate the significance of phenomena on radially-invariant parameters, i.e., the convolution filters followed by BN, we only apply variants of Adam including  AdamG and AdamW on convolution layers. 
For comparison consistency, we keep standard Adam on the remaining parameters, and we use a fixed grid hyperparameter search budget and frequency for each method and each architecture (see  \ref{training_implem_details} for details).

Besides, please also remember that the impact of the scaling and bias parameters (which belongs to the remaining non radially-invariant parameters) is out of the scope of this study. Nevertheless, we additionally evaluate the variants of Adam on CNNs with BN without scaling and bias parameters (BN w/o affine) in \ref{bn_wo_affine} and observe marginal performance difference in comparison to CNNs with standard BN. It hints that the normalization layer in BN plays the most important role regarding the model performance. 


\begin{table}[t!]
\centering
\renewcommand{\figurename}{Table}
\caption{\textbf{Accuracy of Adam and its variants when training with BN layers}.} The figures in this table are the mean top1 accuracy $\pm$ the standard deviation over 5 seeds on the test set for CIFAR10, CIFAR100 and on the validation set for SVHN. $^\dagger$~indicates that the original method is only used on convolutional filters while Adam is used for other parameters.

\vspace{2mm}
\scalebox{0.73}
{
\begin{tabular}{@{} l | ccc | c c | c c @{}}
\toprule
       & \multicolumn{3}{c|}{CIFAR10}   & \multicolumn{2}{c|}{CIFAR100} & \multicolumn{2}{c}{SVHN}\\    
Method & \texttt{ResNet20} & \texttt{ResNet18} & \texttt{VGG16} & \texttt{ResNet18} & \texttt{VGG16} & \texttt{ResNet18} & \texttt{VGG16}\\
\midrule
Adam\hphantom{W*}  & 90.98 $\pm$ 0.06 & 93.77 $\pm$ 0.20 & 92.83 $\pm$ 0.17 & 71.30 $\pm$ 0.36 & 68.43 $\pm$ 0.16 & 95.32 $\pm$ 0.23 & 95.57 $\pm$ 0.20\\
AdamW$^\dagger$  &90.19 $\pm$ 0.24 & 93.61 $\pm$ 0.12 & 92.53 $\pm$ 0.25 & 67.39 $\pm$ 0.27 & 71.37 $\pm$ 0.22 & 95.13 $\pm$ 0.15 & 94.97 $\pm$ 0.08 \\
AdamG$^\dagger$   & 91.64 $\pm$ 0.17 & 94.67 $\pm$ 0.12 &  93.41 $\pm$ 0.17 & 73.76 $\pm$ 0.34 & 70.17 $\pm$ 0.20  & 95.73 $\pm$ 0.05  & 95.70 $\pm$ 0.25\\
\midrule
Adam w/o (a)  & 91.15 $\pm$ 0.11 & 93.95 $\pm$ 0.23  & 92.92 $\pm$ 0.11 & 74.44 $\pm$ 0.22 & 68.73 $\pm$ 0.27 & 95.75 $\pm$ 0.09 & 95.66 $\pm$ 0.09\\
Adam w/o (ab)    &  \textbf{91.92} $\pm$ 0.18 &  \textbf{95.11} $\pm$ 0.10 & \textbf{93.89} $\pm$ 0.09 & \textbf{76.15} $\pm$ 0.25 & \textbf{71.53} $\pm$ 0.19 &  \textbf{96.05} $\pm$ 0.12 & \textbf{96.22} $\pm$ 0.09 \\
Adam w/o (abc) & 91.81 $\pm$ 0.20 & 94.92 $\pm$ 0.05  & 93.75 $\pm$ 0.06 & 75.28 $\pm$ 0.35 & 71.45 $\pm$ 0.13 & 95.84 $\pm$ 0.07 & 95.82 $\pm$ 0.05\\
\bottomrule
\end{tabular}
}
\label{tab:results}
\end{table}

\begin{figure}[t!]
    \vspace*{4mm}
    \renewcommand{\captionfont}{\small}
    \renewcommand{\captionlabelfont}{\bf}
    \centering
    \subfigure{
    \includegraphics[width=0.48\linewidth]{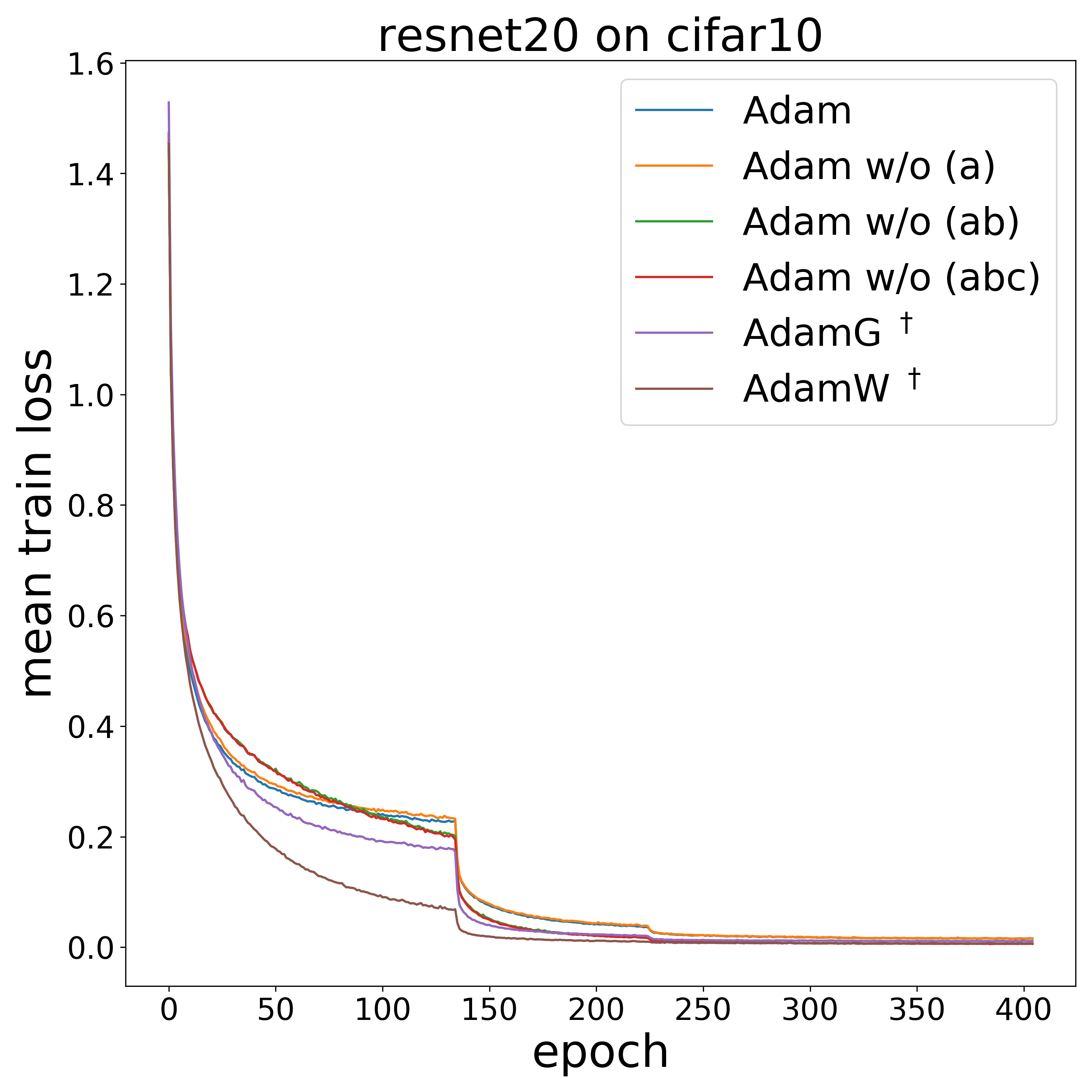}
    }
    \subfigure{
    \includegraphics[width=0.48\linewidth]{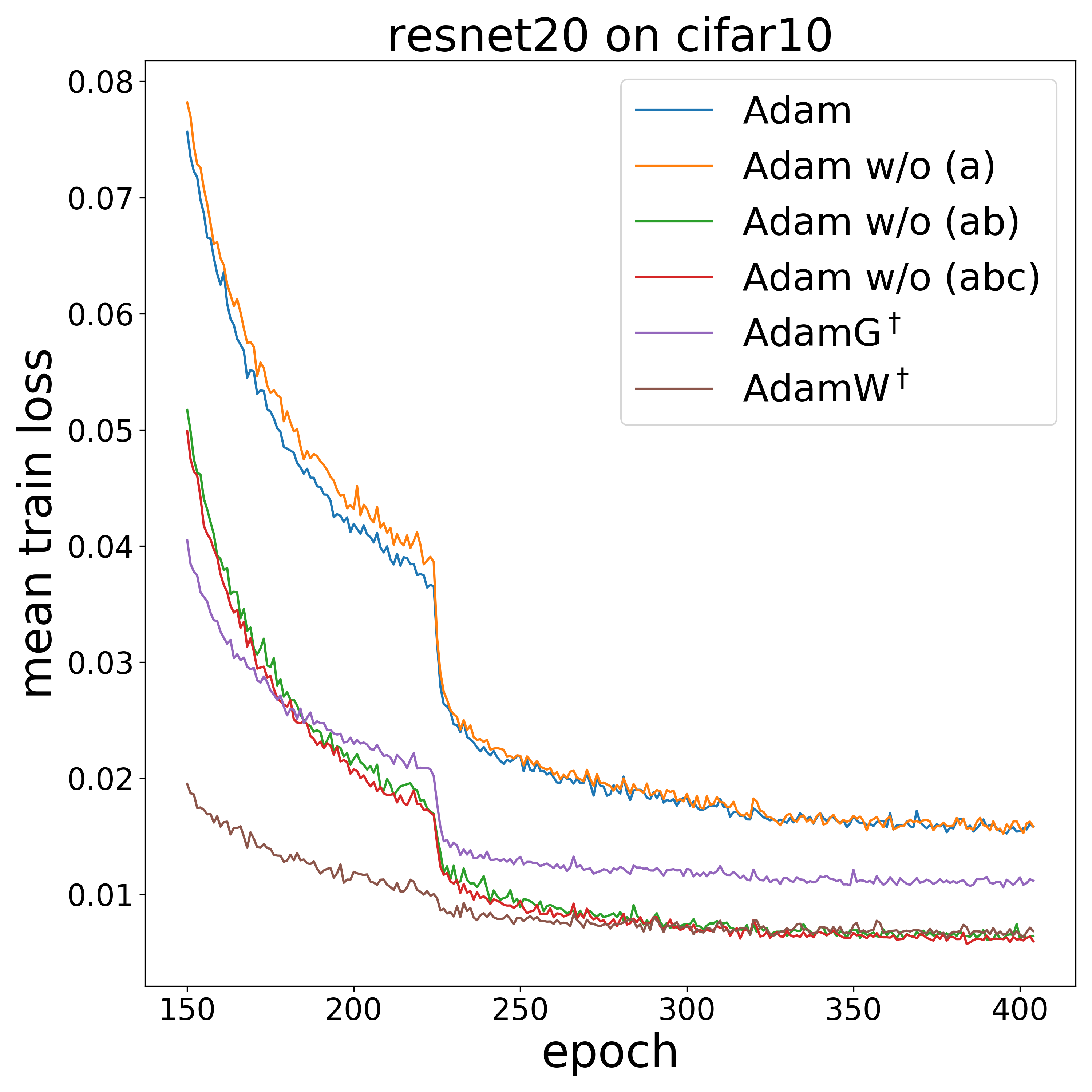}
    }
    \vspace*{-2mm}
    \caption{\textbf{Training speed comparison with ResNet20 BN on CIFAR10.} \emph{Left:} Mean training loss over all training epochs (averaged across 5 seeds) for different Adam variants. \emph{Right:} Zoom-in on the last epochs. Please refer to Table~\ref{tab:results} for the corresponding accuracies.}
    \label{fig:loss_comparison}
\end{figure}

\begin{figure}[t!]
    \vspace*{4mm}
    \renewcommand{\captionfont}{\small}
    \renewcommand{\captionlabelfont}{\bf}
    \centering
    \subfigure{
    \includegraphics[width=0.48\linewidth]{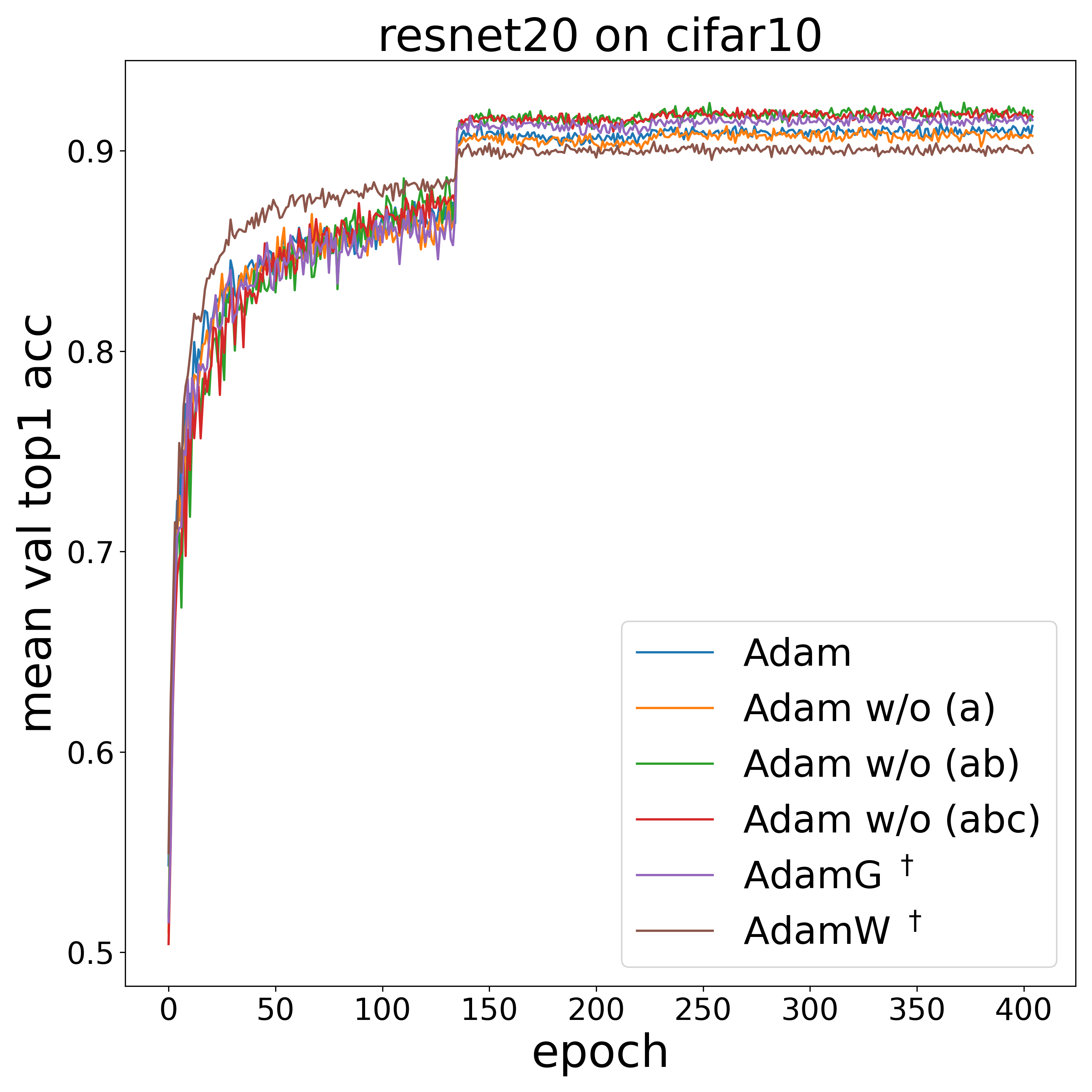}
    }
    \subfigure{
    \includegraphics[width=0.48\linewidth]{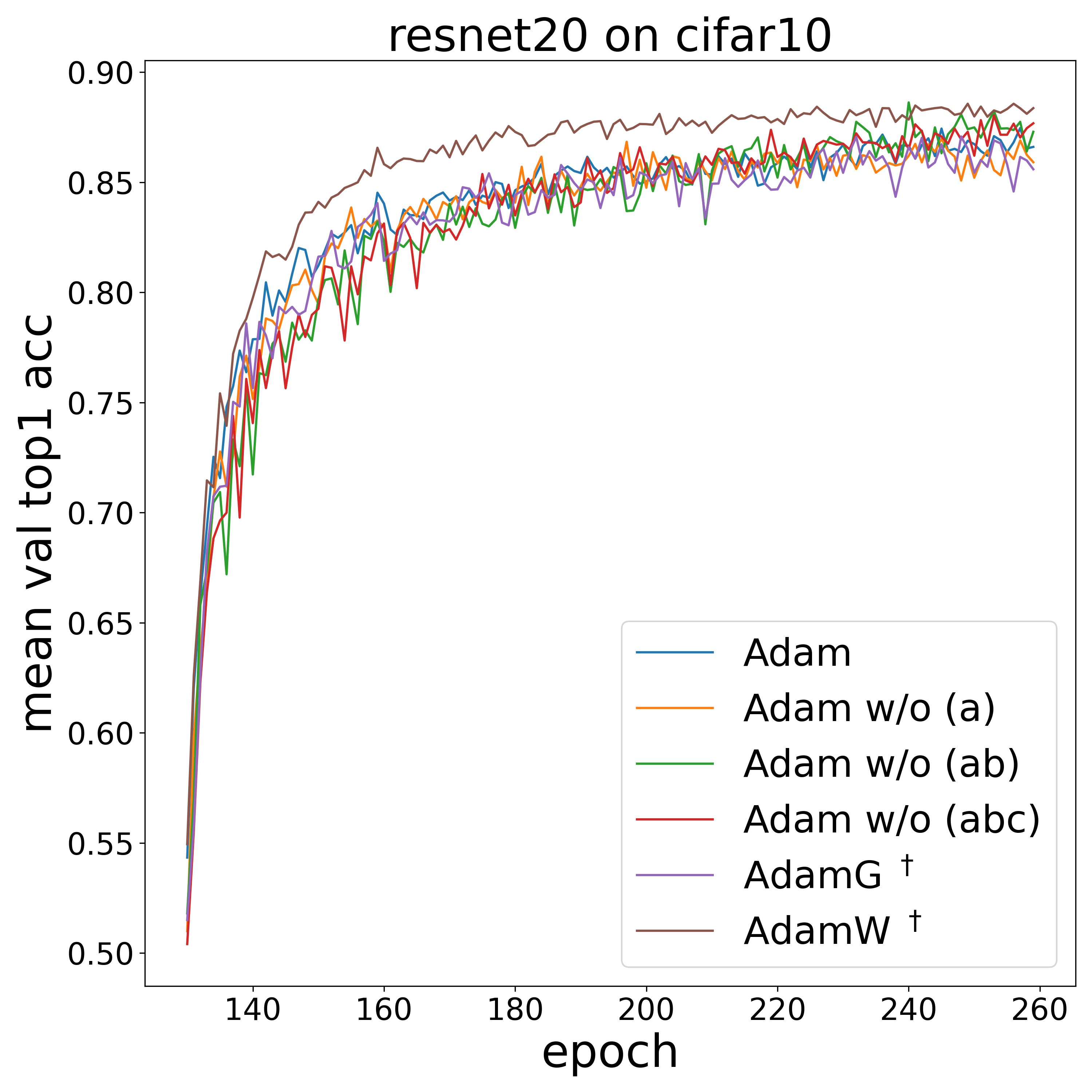}
    }
    \vspace*{-2mm}
    \caption{\textbf{Valid accuracy comparison with ResNet20 BN on CIFAR10.} \emph{Left:} Mean valid top1 acc over all training epochs (averaged across 5 seeds) for different Adam variants. \emph{Right:} Zoom-in on the first epochs. Please refer to Table~\ref{tab:results} for the corresponding accuracies.}
    \label{fig:acc_comparison}
\end{figure}

\textbf{Results.} In Table~\ref{tab:results}, we report quantitative results of Adam variants across architectures and datasets. To indicate that AdamW and AdamG are actually only used on convolutional filters, while Adam is used for the other parameters, we denote the experimented methods as AdamW$^\dagger$ and AdamG$^\dagger$.
In addition, we compare the evolution of the training loss in Figure~\ref{fig:loss_comparison} and the top1 accuracy on the validation set in Figure~\ref{fig:acc_comparison}. 

We observe that each phenomenon displays a specific trade-off between generalization (accuracy on the test set) and training speed, as following. Neutralizing \textbf{(a)} has little effect on the speed over Adam, yet achieves better accuracy on the train set.
Although it slows down training, neutralizing \textbf{(ab)} leads to minima with the overall best accuracy on test set in the case of BN equipped CNNS. 
Note that AdamW$^\dagger$ neutralizes \textbf{(b2)} with its decoupling and is the fastest method, but finds minima with overall worst generalization properties. By constraining the optimization to the hypersphere, AdamG$^\dagger$ speeds up training over the other variants. Finally, neutralizing \textbf{(c)} with Adam w/o \textbf{(abc)} brings a slight acceleration, though reaches lower accuracy than Adam w/o \textbf{(ab)}. In terms of generalization, before the first decrease of the learning rate, neutralizing \textbf{(a)} displays the same speed in terms of reached accuracy on the valid set compared to Adam. When neutralizing \textbf{(ab)} and \textbf{(abc)}, we observe a slight increase, comparable with AdamG$\dagger$. After the first decrease of the learning rate, we observe the same hierarchy as in Table~\ref{tab:results}.

These results show that the geometrical phenomena revealed by our analysis in the spherical framework have a significant impact on the training of BN-equipped CNNs. 

\section{Related work}
\label{sec:relwork}

\parag{Understanding Batch Normalization.} Albeit conceptually simple,  
BN has been shown to have complex implications over optimization. The argument of Internal Covariate Shift reduction~\citep{ioffe2015batch} has been challenged and shown to be secondary to smoothing of optimization landscape~\citep{santurkar2018does, ghorbani2019investigation} 
or its modification by 
creating a different objective function~\citep{lian2019revisit}, enabling of high learning rates through improved conditioning~\citep{bjorck2018understanding}, or alleviating the sharpness of the Fisher information matrix\cite{karakida2019normalization}. In an analysis of a variant of GD, Kohler et al.~\cite{kohler2019exponential} show that BN accelerates optimization by decoupling the optimization of the direction and length of parameters. Daneshmand et al.~\cite{daneshmand2020batch} argue that BN is an effective strategy to ensure that activations generated
by a randomly-initialized network have high rank, unlike vanilla networks where the rank in the final layers collapses with depth~\cite{saxe2013exact}. 
Arora et al.~\cite{arora2018theoretical} demonstrate that (S)GD with BN is robust to the choice of the learning rate, with guaranteed asymptotic convergence, while a similar finding for GD with BN is made 
in~\cite{cai2018quantitative}.

\parag{Invariances in neural networks.} Cho and Lee~\cite{cho2017riemannian} propose optimizing over the Grassmann manifold using Riemannian 
GD. 
Li and Arora~\cite{liu2017deep} project weights and activations on the unit hypersphere and compute a function of the angle between them
instead of inner products, and subsequently generalize these operators by scaling the angle~\citep{liu2018decoupled}. In~\citep{li2019exponential} the radial invariance is leveraged to prove that weight decay (WD) can be replaced by an exponential learning-rate scheduling for SGD with or without momentum.
Arora et al.~\cite{arora2018theoretical} investigate the radial invariance 
and show that radius dynamics depends on the past gradients, offering an adaptive behavior to the learning rate. Here we go further and show that SGD projected on the unit hypersphere corresponds to Adam constrained to the hypersphere, and we give an accurate definition of this adaptive behavior.

\parag{Effective learning rate.} Due to its scale invariance, BN can adaptively adjust the learning rate~\citep{van2017l2,cho2017riemannian,arora2018theoretical,li2019exponential}. Van Laarhoven \cite{van2017l2} shows that in BN-equipped networks, WD increases the effective learning rate by reducing the norm of the weights. Conversely, without WD, the norm grows unbounded~\citep{soudry2018implicit}, decreasing the effective learning rate. Zhang et al.~\cite{zhang2018three} bring additional evidence supporting 
hypothesis in \cite{van2017l2}, 
while Hoffer et al.~\cite{hoffer2018norm} find an exact formulation of the effective learning rate for SGD in normalized networks.
In contrast with prior work, we find generic definitions of the effective learning rate with exact expressions for SGD and Adam. 

\section{Conclusion}
The spherical framework introduced in this study provides a powerful tool to analyse Adam optimization scheme through its projection on the $L_2$ unit hypersphere. It allows us to give a precise definition and expression of the effective learning rate for Adam, to relate SGD to a variant of Adam, and to identify geometric phenomena which empirically impact training.
The framework also brings light to existing variations of Adam, such as $L_2$-regularization decoupling. The geometry of invariance properties appears as a promising research direction toward a better understanding of their impact on optimization.



\bibliography{mybibfile}

\begin{thebibliography}{10}
\bibitem{tieleman2012lecture}
T.~Tieleman, G.~Hinton, Lecture 6.5-rmsprop: Divide the gradient by a running
  average of its recent magnitude, COURSERA: Neural networks for machine
  learning.
\end{thebibliography}
\clearpage
\textbf{Simon Roburin} is a Ph.D. student in Deep Learning at Ecole National des Ponts ParisTech (ENPC) and valeo.ai research lab. He received a Bsc. in Mathematics from University of Paris as valecditorian of the class, a Bsc. in Engineerring and a M.Sc. degree in Applied Mathematics from Ecole Centrale Paris. He is also laureate of the Paris Graduate school of Mathematics from the Foundation Sciences Math\'ematiques de Paris. His research interests focus on Optimization as well as Robustness in Deep Learning.
\\
\\
\noindent \textbf{Yann de Mont-Marin} is a Ph D. student in Robotics and Machine Learning at Inria Paris. He received a a Bsc. in Engineerring and a M.Sc. degree in Applied Mathematics from Ecole Centrale Paris. He previously developed new language identification tools in the service industry. His research interests now focus on Robotic Motion as well as Machine Learning approach to Optimal Control.
\\
\\
\noindent \textbf{Andrei Bursuc} is Senior Research Scientist at valeo.ai in Paris, France. 
He completed his PhD at Mines ParisTech in 2012. He was a postdoc researcher at Inria Rennes and Inria Paris. In 2016, he moved to industry to pursue research on autonomous systems. His current research interests concern computer vision and deep learning, in particular learning  with limited supervision and predictive uncertainty quantification. 
\\
\\
\textbf{Renaud Marlet} is a Senior Researcher at \'Ecole des Ponts ParisTech (ENPC) and a Principal Scientist at valeo.ai, France. He previously held positions both in academia (researcher at Inria) and in the software industry (expert at Simulog, deputy CTO of Trusted Logic). He was the head of the IMAGINE group at LIGM/ENPC (2010-2019). He is currently interested in 3D scene understanding and reconstruction.
\\
\\
\noindent \textbf{Patrick Pérez} is Scientific Director of Valeo.ai, a Valeo research lab on artificial intelligence for automotive applications. Before joining Valeo, Patrick P\'erez has been Distinguished Scientist at Technicolor (2009-2918), researcher at Inria (1993-2000, 2004-2009) and at Microsoft Research Cambridge (2000-2004). 
His research revolves around machine learning for scene understanding, data mining and visual editing.
\\
\\
\noindent \textbf{Mathieu Aubry} received MSc degree from the Ecole Polytechnique, Ph D. from the Ecole Normale Superieure and Habilitation from Universite Paris-Est. In 2015, he spent a year as a visiting researcher at UC Berkeley and is a researcher at Ecole des Ponts. He works on unsupervised image and 3D shape analysis and developing Computer Vision approaches for Digital Humanities.
\clearpage
\appendix
\clearpage

\begin{center}
{\Large\bf Supplementary Material to \\ ``Spherical Perspective on Learning\\ with Normalization Layers''}
\end{center}

\appendix 
\section{Radial invariance of filters with BN}\label{bn_rescale}
In this section, we show the radial invariance of a set of filters equipped with BN.
Please note that the following notations are specific and restricted to this section. 

For the sake of simplicity, we only consider the case of a convolutional layer that preserves the spatial extension of the input. We also focus on a single filter. Since all filters act independently on input data, the following calculation holds for any filter.

Let $\mathbf{x} \in \mathbb{R}^{C \times K}$ be the parameters of a single filter, where $C$ is the number of input channels and $K$ is the kernel size. During training, this layer is followed by BN and applied to a batch $\mathbf{s} \in \mathbb{R}^{B \times C \times D}$ of $B$ inputs of spatial size $D$. 
The output of the convolution operator $\phi$ applied to a filter $\mathbf{x} \in \mathbb{R}^{C \times K}$ and to a given batch element $\mathbf{s}_{b} \in \mathbb{R}^{C \times D}$, with
$b \in \llbracket 1, B \rrbracket$, is thus:
\begin{align}
    \mathbf{t}_{b} &\eqdef 
    \phi(\mathbf{x}, \mathbf{s}_{b}) \in \mathbb{R}^{D}.
\end{align}

The application $(\mathbf{x}, \mathbf{s}_{b}) \mapsto \phi(\mathbf{x}, \mathbf{s}_b)$ is bilinear. BN then centers and normalizes the output $\mathbf{t}$ using the mean and variance over the batch and the spatial dimension: 
\begin{align}
    \mu &= \frac{1}{BD} \sum_{b,j} t_{b,j}, \\
    \sigma^2 &= \frac{1}{BD} \sum_{b,j} \left(t_{b,j} - \mu\right)^{2}, \\
    \hat{\mathbf{t}}_{b} &\eqdef
    (\sigma^2 +\epsilon)^{\nicefrac{{-}1}{2}} \left(\mathbf{t}_{b} -\mu\mathbf{1}_D\right), 
\end{align}
where $\mathbf{1}_D$ denotes the all-ones vector of dimension $D$ and $\epsilon$ is a small constant.

Now if the coefficients of the filter are rescaled by $\rho>0$, then, by bilinearity, the new output of the layer for this filter verifies:
\begin{align}
    \Tilde{\mathbf{t}}_{b} &=
    \phi(\rho \mathbf{x}, \mathbf{s}_{b}) = \rho \phi(\mathbf{x}, \mathbf{s}_{b}).
\end{align}
Since the variance of inputs is generally large in practice, for small $\epsilon$, the mean and variance are:
\begin{align}
    \Tilde{\mu} &= \rho \mu, \\
    \Tilde{\sigma}^2 & \approx \rho^{2} \sigma^2.
\end{align}
It can then be considered that the subsequent BN layer is invariant to this rescaling, \ie, 
$\hat{\tilde{\mathbf{t}}}_{b} \approx \hat{\mathbf{t}}_{b}$. 


\section{Extension to other normalization layers}\label{other_norm_layer}

The radial invariance for BN described above in \ref{bn_rescale} applies as well to InstanceNorm (IN) \citep{ulyanov2016instance} and WeightNorm (WN) \citep{salimans2016weight} as the normalization is also done with respect to channels but without the batch dimension. Regarding LayerNorm \citep{ba2016layer} (LN), the normalization is performed over all channels and the entire weight layer can thus be rescaled too, without impacting the output. 
As for GroupNorm \citep{wu2018group} (GN), it associates several channels for normalization; the radial invariance in this case concerns the corresponding group of filters.

Thanks to this general property of radial invariance, the results in this paper not only concern BN but also WN and IN. In fact, they apply as well to LN and GN when considering the suitable group of parameters.
The optimization in this case concerns the proper 
slice of the parameter tensor of the layer, i.e., the whole tensor for LN, and the selected group of filters for GN.


\section{Results in Sections 2 and 3}

In this section, we provide proofs and/or empirical results supporting the claims in Sections 2 and 3 of the paper.

In the following, the double parentheses around an equation number, e.g., (\eqref{eq:1_th}), indicate that we recall an equation that was previously stated in the main paper, rather than introduce a new one, e.g., noted \eqref{aklrn}. Also, framed formulas actually refer to results stated in the main paper, thus with double-bracket equation numbering. 

\subsection{Proof of theorems and validity of assumptions}

\subsubsection{Proof of Theorem~\ref{thm:sphere_step} (Image step on $\mathcal{S}_{d{-}1}$) in Section~\ref{sec:framework}}\label{calculus_32}

We recall the main theorem in Section~\ref{sec:framework}.
\smallskip
\begin{mdframed}
\textbf{Theorem~\ref{thm:sphere_step}} (Image step on $\mathcal{S}_{d{-}1}$)
\it
If the following hypothesis are verified:
\begin{itemize}
    \item (H1): $1{-}\lrn{\langle \bc_k, \bu_k \rangle}> 0$.
    \item (H2): $\elr{k} \|\ckortho\|< \pi$.
\end{itemize}
The update of a group of radially-invariant parameters $\bx_k$  at step $k$ following the generic optimization scheme (Eqs.~\ref{eq1a}-\ref{eq1b}) and the corresponding update of its projection $\bu_k$ on $\unithyp$ is given by an exponential map at $\bu_k$ with velocity $\elr{k}\dirk$:
\begin{equation}
    \bu_{k+1} =	 \operatorname{Exp}_{\bu_{k}}\left(-\left[1 + O\left(\left(\elr{k}\|\dirk\|\right)^2\right)\right]\elr{k}\dirk\right),\tag{\eqref{eq:1_th}}
\end{equation}
where $\operatorname{Exp}_{\bu_{k}}$ is the exponential map on $\unithyp$, and with \begin{align} 
\bc_k \eqdef r_k\ba_k \oslash \disto,\quad 
\elr{k} \eqdef \lrn{} \left(1 - \lrn{\langle \bc_k, \bu_k \rangle} \right)^{-1}\tag{\eqref{lredef}}.
\end{align}
More precisely:
\begin{align}
\bu_{k+1} &= \frac{\bu_k - \elr{k}\dirk}{\sqrt{1 + (\elr{k}\| \dirk \|)^{2}}}.\tag{\eqref{canonic}}
\end{align}
%
\end{mdframed}

\begin{proof}
To simplify the calculation in the demonstration, we introduce the following notation:
\begin{align}
    A_k \eqdef \lrn{}.\label{aklrn}
\end{align}
We first demonstrate the expression for the radius dynamics in Eq.\,\eqref{rdynamic} and the precise step for $\bu$ in Eq.\,\eqref{canonic}. Then we use geometric arguments and a Taylor expansion to derive the update on the sphere stated in Eq.\eqref{eq:1_th}.

\textbf{Radius dynamics.}
We first show Eq.\,\eqref{rdynamic}, which we recall here using the $A_k$ notation:
\begin{align}
    \boxed{\frac{r_{k+1}}{r_k} = \left(1 - A_k{\langle \bc_k, \bu_k \rangle} \right)\sqrt{1 + (\elr{k}\| \dirk \|)^{2}}.}
    \tag{\eqref{rdynamic}}
\end{align}

First, we rewrite the step of a generic scheme in Eqs.\,(\ref{eq1a}-\ref{eq1b}) along the radial and tangential directions and separate the division vector $\bb_k$ into its deformation $\disto$ and its scalar scheduling effect $\invsqf\|\bb_k\|$, as stated in the discussion:
\begin{align}
    r_{k+1}\bu_{k+1} &= r_k\bu_k - \frac{\eta_k}{\invsqf\|\bb_k\|}\ba_k\oslash \disto \nonumber\\
    &= r_k \left[ \bu_k - \lrn{}r_k\ba_k\oslash \disto\right]\nonumber\\
    &= r_k \left[ \bu_k - A_k r_k\ba_k\oslash \disto\right].\label{precanonic}
\end{align}
We can note the appearance of a new term $r_k\ba_k$. The vector $\ba_k$ is a gradient momentum and therefore homogeneous to a gradient. Using Lemma~\ref{lem_invar}, $r_k\ba_k$ is homogeneous to a gradient on the hypersphere and can be interpreted as the momentum on the hypersphere.

From Eq.\,\eqref{precanonic}, we introduce $\bc_k$ (the deformed momentum on hypersphere) as in Eq.\,\eqref{lredef} and decompose it into the radial and tangential components. We have:
\begin{align}
     \frac{r_{k+1}}{r_k} \bu_{k+1} &= \bu_k - A_k \bc_{k} \nonumber \\
    &= \left(1 - A_k\langle \bc_k, \bu_k \rangle\right)\bu_k - A_k\bc_{k}^{\perp}.  \label{proof_int}
\end{align}
By taking the squared norm of the equation, we obtain:
\begin{align}
    \frac{r_{k+1}^{2}}{r_{k}^{2}} &= \left(1 - A_k\ckrad \right)^{2} +\left(A_k \|\bc_{k}^{\perp} \|\right)^{2}.
\end{align}
Making the assumption that $1\,{-}\, A_k\ckrad > 0$, which is true in practice and discussed in the next subsection, we have:
\begin{align}
    \frac{r_{k+1}}{r_{k}} &= (1 - A_k\ckrad)\sqrt{1 + \left(\frac{A_k}{1 - A_k\ckrad}\|\bc_{k}^{\perp} \|\right)^2}.
\end{align}
After introducing $\elr{k} = \frac{A_k}{\left(1 - A_k \ckrad \right)}$ as in Eq.\,\eqref{lredef}, we obtain the result of \eqref{rdynamic}.

\textbf{Update of normalized parameters.}
We then show Eq.\,\eqref{canonic}:
\begin{align}
    \boxed{\bu_{k+1} = \frac{\bu_k - \elr{k}\dirk}{\sqrt{1 + (\elr{k}\| \dirk \|)^{2}}}.}\tag{\eqref{canonic}}
\end{align}
Combining the radius dynamics previously calculated with Eq.~\eqref{proof_int}, we have:
\begin{align}
    \bu_{k+1} = &\frac{(1 - A_k \langle \bc_k, \bu_k \rangle)\bu_k - A_k \bc_{k}^{\perp}}{(1 - A_k\langle \bc_{k}, \bu_{k} \rangle)\sqrt{1 + (\elr{k}\| \dirk \|)^{2}}}\\
    = & \frac{\bu_k - \frac{A_k}{1 - A_k\langle \bc_{k}, \bu_{k} \rangle} \bc_{k}^{\perp}}{\sqrt{1 + (\elr{k}\| \dirk \|)^{2}}}.
\end{align}
Hence the result \eqref{canonic} using the definition of $\elr{k}$. 

This result provides a unique decomposition of the generic step as a step in $\operatorname{span}(\bu_{k},\dirk)$ for the normalized filter (Eq.\,\eqref{canonic}) and as a radius update (Eq.\,\eqref{rdynamic}).

We split the rest of the proof of the theorem in three parts.

\textbf{Distance covered on the sphere.~} 
The distance covered on the hypersphere $\mathcal{S}_{d-1}$ by an optimization step is: 
\begin{equation}
\mathrm{dist}_{\mathcal{S}_{d-1}}(\mathbf{u}_{k+1}, \mathbf{u}_{k}) = \arccos(\langle \mathbf{u}_{k+1}, \mathbf{u}_{k} \rangle).
\end{equation}
From Eq.\,\eqref{canonic} and with Lemma~\ref{lem_invar}, we also have:
\begin{align}
    \langle \mathbf{u}_{k+1}, \mathbf{u}_{k} \rangle
    &= \frac{1}{\sqrt{1+(\eta^e_k \|\mathbf{c}_{k}^{\perp}\|)^2}}.
\end{align}
Therefore, $\mathrm{dist}_{\mathcal{S}_{d-1}}(\mathbf{u}_{k+1}, \mathbf{u}_{k})=\varphi(\eta^e_k \|\mathbf{c}_{k}^{\perp}\|)$ where $\varphi: z \mapsto \arccos\left(\frac{1}{\sqrt{1 + z ^{2}}}\right)$, which is equal to $\arctan$ on $\mathbb{R}_{+}$. Then a Taylor expansion at order~3 of $\arctan$ yields for $\eta^e_k \|\mathbf{c}_{k}^{\perp}\|$:
\begin{equation}
\mathrm{dist}_{\mathcal{S}_{d-1}}(\mathbf{u}_{k+1}, \mathbf{u}_{k})=\eta^e_k \|\mathbf{c}_{k}^{\perp}\| + O\left(\left(\eta^e_k \|\mathbf{c}_{k}^{\perp}\|\right)^{3}\right).\label{dist_calcul}
\end{equation}
The Taylor expansion validity is discussed in the next subsection.

\textbf{Exponential map on the sphere.~} Given a Riemannian manifold $\mathcal{M}$, for a point $\bu \in \mathcal{M}$ there exists an open set $\mathcal{O}$ of the tangent space $\mathcal{T}_{\bu}\mathcal{M}$ containing $\mathbf{0}$, such that for any tangent vector $\mathbf{w} \in \mathcal{O}$ there is a unique geodesic (a path minimizing the local distance on $\mathcal{M}$ when conserving the tangent velocity) $\gamma:[-1,1] \rightarrow \mathcal{M}$ that is differentiable and such that $\gamma(0) = \mathbf{u}$ and $\gamma'(0) = \mathbf{w}$. Then, the exponential map of $\mathbf{w}$ from $\bu$ is defined as $\operatorname{Exp}_{\bu}(\mathbf{w}) = \gamma(1)$.

In the case of the manifold $\unithyp$, the geodesics are complete (they are well defined for any point $\bu \in \unithyp$ and any velocity $\mathbf{w}\in \mathcal{T}_{\bu}\unithyp$) and are the great circles: for any $\bu \in \unithyp$ and any $\mathbf{w} \in \mathcal{T}_{\bu}\unithyp$, the map $\psi : t \in \mathbb{R} \mapsto \operatorname{Exp}_{\mathbf{u}}(t\mathbf{w}))$ verifies $\psi(\mathbb{R}) = \mathcal{S}_{d-1}  \cap  \operatorname{span}( \{ \bu, \mathbf{w} \})$ which is a great circle passing through $\bu$ with tangent $\mathbf{w}$. Furthermore, 
since the circumference of the great circle is $2\pi$, we have that for any $\mathbf{p} \in \unithyp \backslash \{-\bu\}$ there is a unique $\mathbf{w}$ verifying $\|\mathbf{w}\| < \pi$ such that $\mathbf{p} = \operatorname{Exp}_{\mathbf{u}}(\mathbf{w})$ and we have:
\begin{align}
\mathrm{dist}_{\mathcal{S}_{d-1}}(\bu, \mathbf{p}) &= \|\mathbf{w}\|\textnormal{ and } \langle \mathbf{p},\mathbf{w}\rangle \ge 0 .\label{sphere_prop}
\end{align}

\textbf{Optimization step as an exponential map.~} We will use the previously stated differential geometry properties to prove:
\begin{equation}
    \boxed{
    \bu_{k+1} =	 \operatorname{Exp}_{\bu_{k}}\left(-\left[1 + O\left(\left(\elr{k}\|\dirk\|\right)^2\right)\right]\elr{k}\dirk\right).}\tag{\eqref{eq:1_th}}
\end{equation}

For an optimization step we have:
\begin{itemize}[nosep]
    \item by construction, $\mathbf{c}_{k}^{\perp} \in \mathcal{T}_{\bu_k}\mathcal{S}_{d-1}$;
    \item from Eq.\,\eqref{canonic}, $\mathbf{u}_{k+1} \in \mathcal{S}_{d-1}  \cap  \operatorname{span}(\{ \bu_k, \mathbf{c}_{k}^{\perp}\})$;
     \item from Eq.\,\eqref{canonic}, $\langle \bu_{k+1},\ckortho \rangle \le 0$.
\end{itemize}
Then, there exists $\alpha$ that verifies $ \|\alpha\ckortho\|< \pi$ such that:
\begin{equation}
    \bu_{k+1} =	 \operatorname{Exp}_{\bu_{k}}\left(\alpha\ckortho\right).
\end{equation}
From Eq.\,\eqref{sphere_prop}, because of the inequality $\langle \bu_{k+1},\ckortho \rangle \le 0$, we have $\alpha < 0$. We also have that $ \|\alpha\ckortho\| = \mathrm{dist}_{\mathcal{S}_{d-1}}(\mathbf{u}_{k+1}, \mathbf{u}_{k})$. Then, using the distance previously calculated in Eq.\,\eqref{dist_calcul}, we have:
\begin{align}
    |\alpha|\|\ckortho\| &=\eta^e_k \|\mathbf{c}_{k}^{\perp}\| + O\left(\left(\eta^e_k \|\mathbf{c}_{k}^{\perp}\|\right)^{3}\right),\\
    |\alpha|&=\eta^e_k\left[1 + O\left(\left(\elr{k}\|\dirk\|\right)^2\right)\right].
\end{align}
Combining the sign and absolute value of $\alpha$, we get the final exponential map expression:
\begin{align}
    \bu_{k+1} &=	 \operatorname{Exp}_{\bu_{k}}\left(-\left[1 + O\left(\left(\elr{k}\|\dirk\|\right)^2\right)\right]\elr{k}\dirk\right),\tag{\eqref{eq:1_th}} \\
    &\approx \operatorname{Exp}_{\bu_{k}}\left(-\elr{k}\dirk\right).
\end{align}
Note that we implicitly assume here that $|\alpha|\|\ckortho\| \approx \elr{k} \|\ckortho\|< \pi$, which is discussed in the next subsection.

\end{proof}

\subsubsection{Assumptions in Theorem~\ref{thm:sphere_step} and validity}\label{positivity}

\begin{figure}[t]
\centering
\includegraphics[width=0.46\columnwidth]{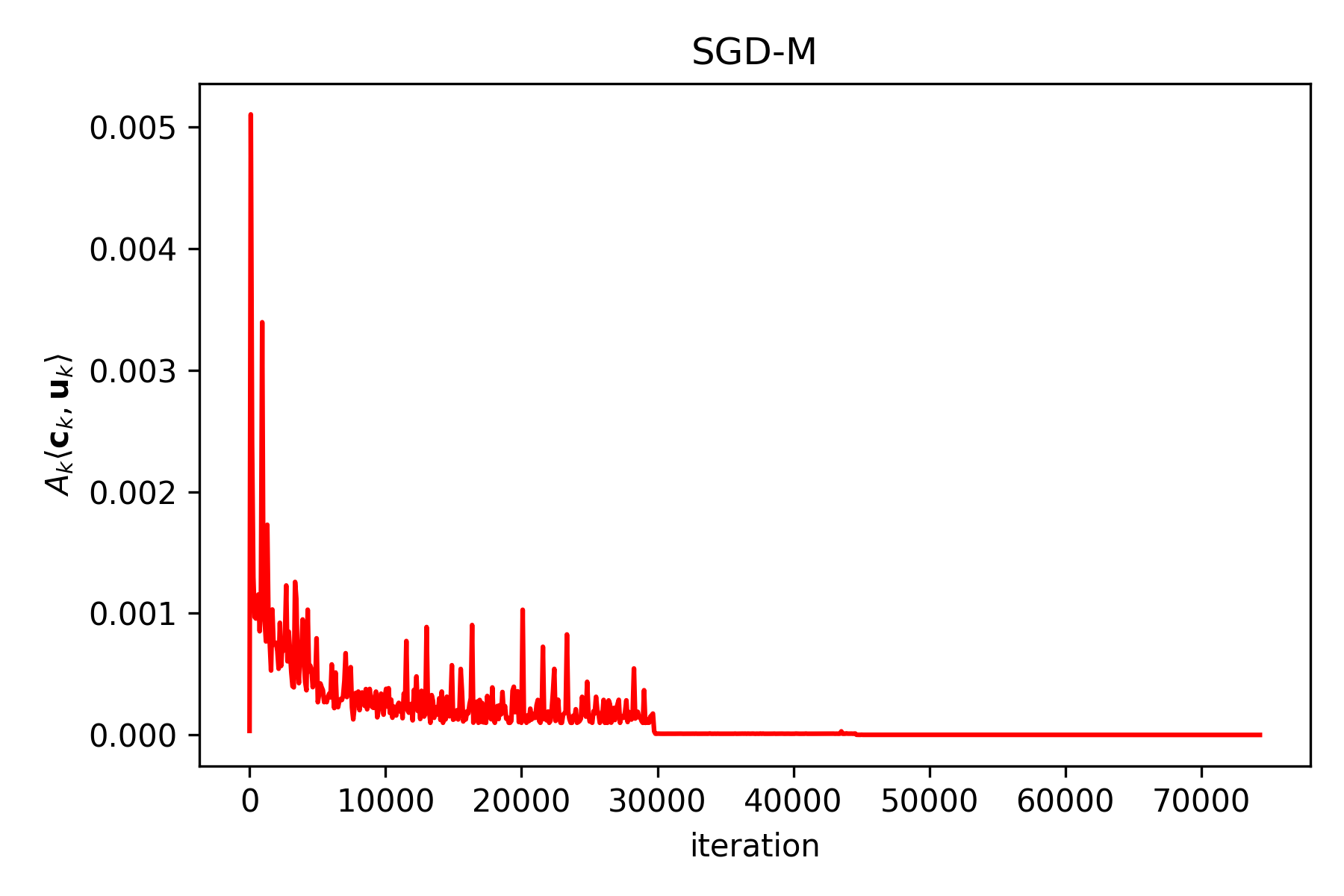}
\includegraphics[width=0.46\columnwidth]{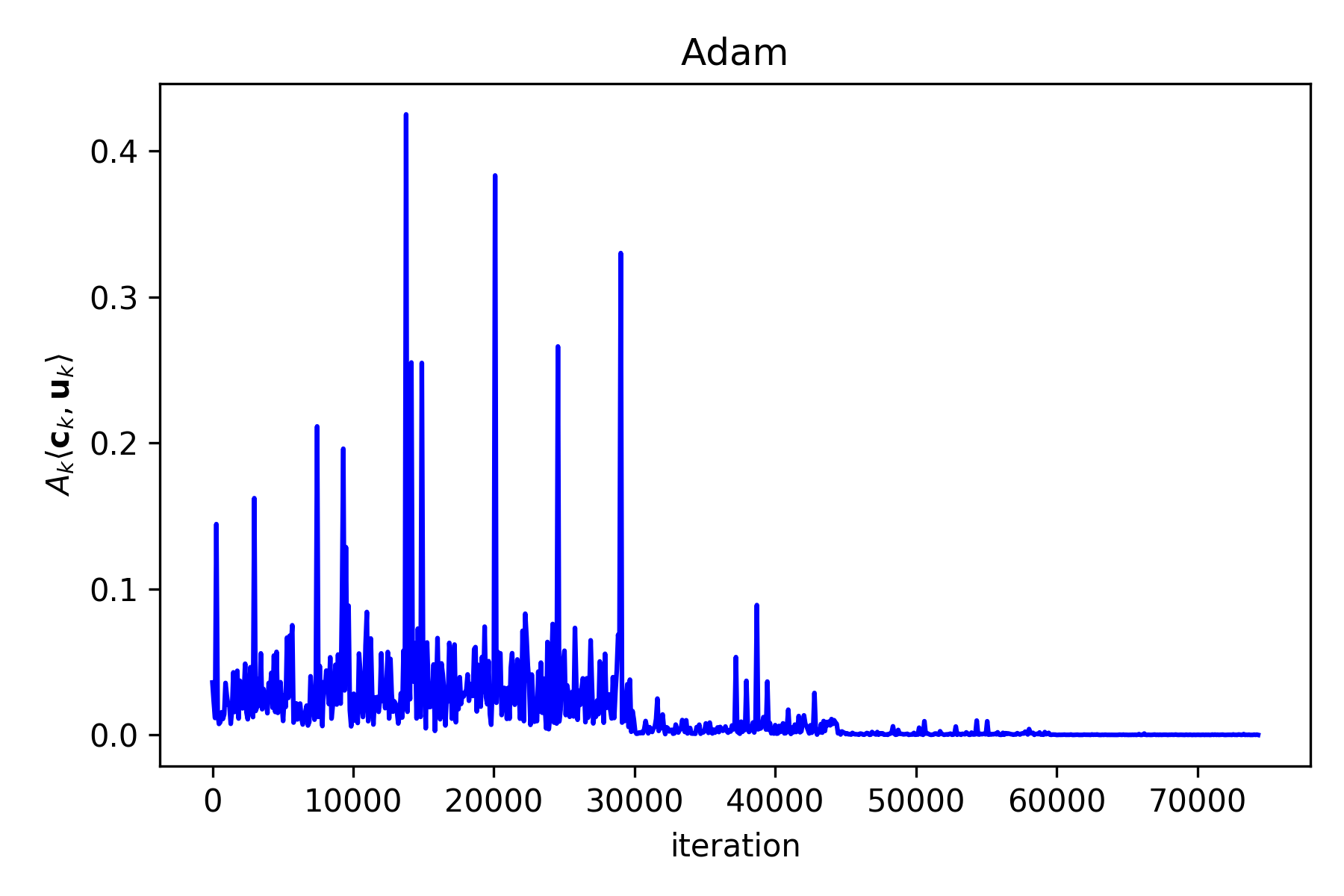}
\caption{\textbf{Tracking of $A_k\langle \bc_k, \bu_k \rangle$ for SGD-M and Adam.} The above graphs show the maximum of the absolute value of $A_k\langle \bc_k, \bu_k \rangle$ for all filters in all layers of a ResNet20 CIFAR trained on CIFAR10 and optimized with SGD-M (left) or Adam (right). The quantity is always small compared to~$1$. Therefore we may assume that $1 - A_k\langle \bc_k, \bu_k \rangle \ge 0$.
}
\label{sign_ratiodl}
\vspace{-2mm}
\end{figure}

\paragraph{Sign of $1 - A_k \langle \bc_k, \bu_k \rangle$.}
We tracked the maximum of the quantity $A_k\langle \bc_k, \bu_k \rangle$  
for all the  filters of a ResNet20 CIFAR trained on CIFAR10 and optimized with SGD-M or Adam (see Appendix \ref{training_implem_details} for implementation details). As can be seen on Figure~\ref{sign_ratiodl}, 
this quantity is always small compared to~$1$,  
making $1 - A_k\langle \bc_k, \bu_k \rangle$ always positive in practice. The order of magnitude of this quantity is roughly the same for different architectures and datasets. 

\begin{figure}[t]
\centering
\hspace*{6mm}\includegraphics[width=0.49\columnwidth]{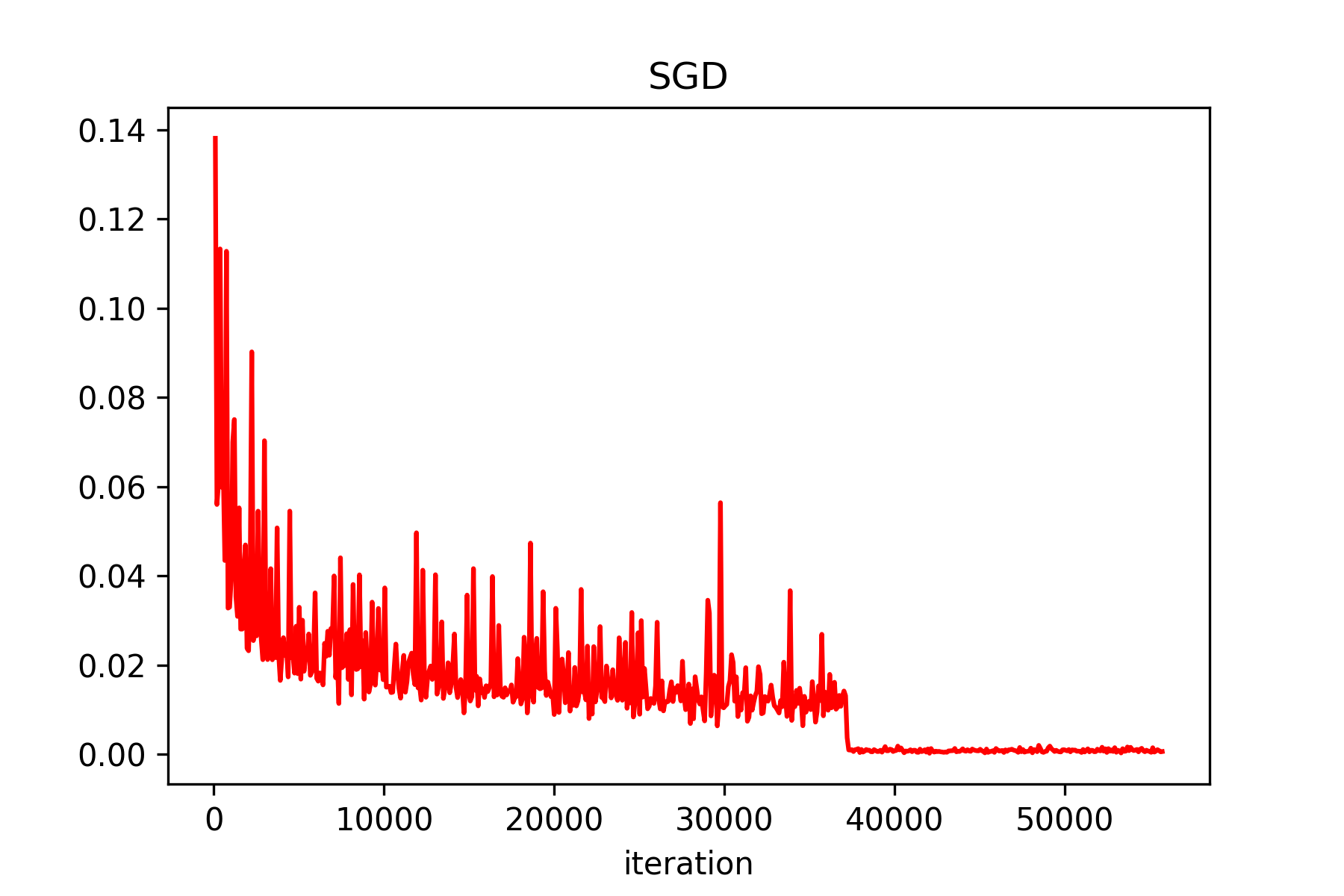}\hspace*{-5mm}
\includegraphics[width=0.49\columnwidth]{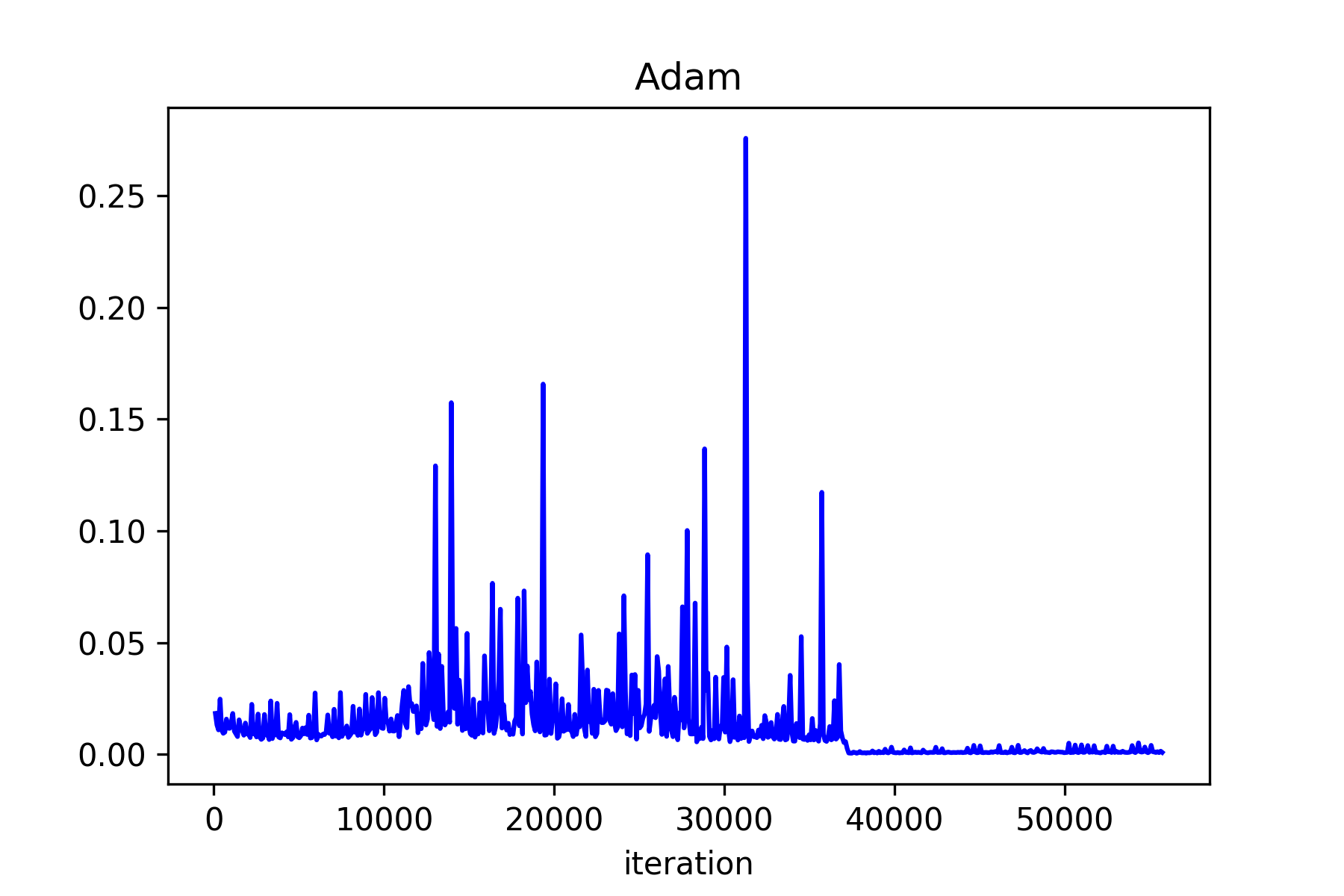}
\caption{\textbf{Tracking of $\elr{k}\|\ckortho\|$  
for SGD-M and Adam.} The above graphs show the maximum of the absolute value of $\elr{k}\|\ckortho\|$ for all filters in all layers of a ResNet20 CIFAR trained on CIFAR10 and optimized with SGD-M (left) or Adam (right). 
}
\label{step_value}
\vspace{-2mm}
\end{figure}

\paragraph{Taylor expansion.} We tracked the maximum of the quantity $\elr{k}\|\ckortho\|$ for all the  filters of a ResNet20 CIFAR trained on CIFAR10 and optimized with SGD-M or Adam. The observed values justify the Taylor expansion and validate the assumption $|\alpha|\|\ckortho\|\approx \elr{k} \|\ckortho\| < \pi$. (cf.\ Figure~\ref{step_value}). The order of magnitude of this quantity is roughly the same for other different architectures and datasets. 

\subsubsection{$\nu_k$, order 2 moment on the hypersphere for Adam}\label{order2adam}

\textbf{Scheduling effect of Adam division vector.}  With Eq.\,\eqref{calcul_vk} and using Lemma~\ref{lem_invar}, we can give the expression of the second-order moment on the sphere, defined as $\nu_k = r_k\invsqf\|\bb_k\|$:
\begin{equation}
\boxed{
\begin{aligned}
    \nu_k = &\invsqf\frac{1 - \beta_1^{k+1}}{1 - \beta_1}\Big(\frac{1 - \beta_2}{1 - \beta_2^{k + 1}}\Big)^{\nicefrac{1}{2}}\Big(\sum_{i=0}^{k}\beta_2^{k-i}\frac{r_k^2}{r_i^2}\|\nabla\mathcal{L}(\bu_i)+\lambda r_i^2\bu_i\|^2\Big)^{\nicefrac{1}{2}}.
\end{aligned}
}
\label{nuk}
\end{equation}

\subsubsection{Proof of Theorem 4 (SGD equivalent scheme on the unit hypersphere) in Section~\ref{eff_lr_sgd}}\label{sgd_proof}

We prove the following theorem: 
\begin{mdframed}
\textbf{Theorem 4 }(SGD equivalent scheme on the unit hypersphere.)\it 
\ For any $\lambda>0, \eta>0, r_0>0$, we have the following equivalence at order 2 in the radius dynamics:
\begin{equation*}
    \left\{
    \begin{array}{l}
        \textnormal{(SGD)}\\
        \bx_0 = r_0 \bu_0 \\
        \lambda_k=\lambda \\
        \eta_k=\eta
    \end{array}
    \right.
    \textnormal{ \it is scheme-equivalent at order 2 to } \left\{
    \begin{array}{l}
        \textnormal{(AdaGradG)}\\
        \bx_0 = \bu_0 \\
        \beta = (1 - \eta\lambda)^4 \\
        \eta_k=(2\beta)^{-1/2} \\
        v_0 = r_0^4 (2\eta^{2}\beta^{1/2})^{-1}
    \end{array}
    \right.
\end{equation*}
\end{mdframed}

\begin{proof}
As summarized in Table~\ref{eff_lr_scheme}, the expressions of the effective learning rates and directions for SGD are $\ckortho = r_k\nabla\mathcal{L}(\bx_k) = \nabla
\mathcal{L}(\bu_k)$ and $\elr{k} = \frac{\eta_k}{r_k^2(1-\eta_k \lambda_k)}$.

\textbf{Equivalence with SGD and $L_2$ regularization.} 
We look for conditions leading to an equivalence between SGD with $L_2$ regularization and SGD without $L_2$ regularization. Using Lemma~\ref{lm:eq_scheme}, the equality of effective directions is trivial and the equality of effective learning rates for any step $k$ yields the following equivalence: 
\begin{equation}
    \left\{
    \begin{array}{l}
        \textnormal{(SGD)}\\
        \tilde{\bx}_0 = r_0 \bu_0 \\
        \tilde{\lambda}_k=\lambda \\
        \tilde{\eta}_k=\eta
    \end{array}
    \right.
    \textnormal{ is scheme-equivalent to } \left\{
    \begin{array}{l}
        \textnormal{(SGD)}\\
        \bx_0 = r_0 \bu_0 \\
        \lambda_k=0 \\
        \eta_k=\eta (1 - \eta\lambda)^{-2k-1}
    \end{array}
    \right.
\label{sgd_lr_exp}
\end{equation}
$L_2$ regularization is equivalent to an exponential scheduling of the learning rate, as found in \cite{li2019exponential}. Here, we provide a proof in a constructive manner. We are going to use Lemma~\ref{lm:eq_scheme} and find a sufficient condition to have:
\begin{equation*}
     \left\{
    \begin{array}{l}
        \textnormal{(i) } \bu_0 = \Tilde{\bu}_0 \\
        \textnormal{(ii) } \forall k \geq 0, \eta^e_k = \Tilde{\eta}^e_k, \bc^{\perp}_k = \Tilde{\bc}^{\perp}_k.
    \end{array}
\right.
\end{equation*}

Equation (i) is trivially satisfied by simply taking the same starting point: $\tilde{\bx}_0 = \bx_0$.

Regarding (ii), because effective directions are the same and only depend on $\bu_k$, we only need a sufficient condition on $\elr{k}$.
For effective learning rates, using Eq.\,(\eqref{rdynamic}) and expressions in Table~\ref{eff_lr_scheme}, we have:
\begin{equation}
     \elr{k} = \tilde{\elr{k}} \Leftrightarrow \frac{\eta_k}{r_k^2} = \frac{\tilde{\eta}_k}{\tilde{r}_k^2(1-\tilde{\eta}_k\lambda)}.\label{proof_eq_sgd}
\end{equation}
Since $\tilde{\eta_k} = \eta$, we obtain:
\begin{equation*}
     \textnormal{(\ref{proof_eq_sgd})} \Leftrightarrow \eta_k = \left(\frac{r_k}{\tilde{r}_k}\right)^2\frac{\eta}{(1 -\eta\lambda)}. \\
 \end{equation*}
 Therefore:
  \begin{equation*}
     \frac{\eta_{k+1}}{\eta_k} = \left(\frac{r_{k+1}\tilde{r}_k}{\tilde{r}_{k+1}r_k}\right)^2 = \left(\frac{r_{k+1}/r_k}{\tilde{r}_{k+1}/\tilde{r}_k}\right)^2.
 \end{equation*}
 By using the radius dynamics in Eq.\,\eqref{rdynamic} for the two schemes, SGD and SGD with $L_2$ regularization, and by the equality of effective learning rates and directions, we have:
 \begin{align*}
     \frac{\eta_{k+1}}{\eta_k} &= \left(\frac{\sqrt{1 + (\elr{k}\| \dirk \|)^{2}}}{(1-\eta\lambda)\sqrt{1 + (\tilde{\elr{k}}\| \tilde{\dirk} \|)^{2}}}\right)^2\\
     &= (1 - \eta\lambda)^{-2}.
 \end{align*}
 
 By taking Eq.\,\eqref{proof_eq_sgd} for $k=0$, because $r_0 = \tilde{r}_0$ we have: $\eta_0 = \eta(1 - \eta\lambda)^{-1}$. Combining the previous relation and the initialization case, we derive by induction that  $\eta_k = \eta (1- \eta \lambda)^{-2k-1}$ is a sufficient condition. We can conclude, using Lemma~\ref{lm:eq_scheme}, the equivalence stated in Eq.\,\eqref{sgd_lr_exp}.

\textbf{Resolution of the radius dynamics.}
Without $L_2$ regularization, the absence of radial component in $\bc_k$ makes the radius dynamics simple:
\begin{equation}
    r_{k+1}^2 = r_{k}^2 + \frac{(\eta_k\|\nabla\mathcal{L}(\bu_k)\|)^2}{r_k^2}.
    \label{sgd_rdyn}
\end{equation}
With a Taylor expansion at order 2, we can show that for $k \ge 1$ the solution $$r_k^2 = \sqrt{2\sum_{i=0}^{k-1}(\eta_i\|\nabla\mathcal{L}(\bu_i)|)^2 + r_0^4}$$ satisfies the previous equation. Indeed using the expression at step $k+1$ gives:
\begin{align*}
r_{k+1}^2 &= \sqrt{2\sum_{i=0}^{k-1}(\eta_i\|\nabla\mathcal{L}(\bu_i)|)^2 + r_0^4 + 2(\eta_k\|\nabla\mathcal{L}(\bu_k)\|)^2}\\
&=r_k^2\sqrt{1 + 2\frac{(\eta_k\|\nabla\mathcal{L}(\bu_k)\|)^2}{r_k^4}}\\
&=r_k^2\left(1 + (1/2) 2\frac{(\eta_k\|\nabla\mathcal{L}(\bu_k)\|)^2}{r_k^4}  + o\left(\frac{(\eta_k\|\nabla\mathcal{L}(\bu_k)\|)^2}{r_k^4}\right)\right)\\
&=r_k^2 + \frac{(\eta_k\|\nabla\mathcal{L}(\bu_k)\|)^2}{r_k^2} + o\left(\frac{(\eta_k\|\nabla\mathcal{L}(\bu_k)\|)^2}{r_k^2}\right) .
\end{align*}

Using $\eta_k=\eta (1 - \eta\lambda)^{-2k-1}$, introducing $\beta = (1 - \eta \lambda)^4$, omitting the $o\left(\frac{(\eta_k\|\nabla\mathcal{L}(\bu_k)\|)^2}{r_k^2}\right)$ and injecting the previous solution in the effective learning rate, we obtain the closed form:
\begin{align}
\elr{k} &= \frac{\eta (1 - \eta\lambda)^{-2k-1}}{\sqrt{2\sum_{i=0}^{k-1}\eta^2(1 - \eta\lambda)^{-4i-2}\|\nabla\mathcal{L}(\bu_i)\|^{2} + r_0^4}}\nonumber\\
&=\frac{(2\beta)^{-\frac{1}{2}}}{\sqrt{\sum_{i=0}^{k-1}\beta^{(k-1)-i}\|\nabla\mathcal{L}(\bu_i)\|^{2} + \beta^k\frac{r_0^4}{2\eta^2\beta^{\frac{1}{2}}}}}.
\end{align}

\textbf{AdaGradG.} The AdaGradG scheme is constrained on the hypersphere thanks to the normalization; the radius is therefore constant and equal to 1. The absence of radial component in the update gives: $\ckortho = \nabla\mathcal{L}(\bu_k)$ and $\elr{k} = \frac{\eta_k}{\sqrt{v_k}}$. Thus, the resolution of the induction on $v_k$ leads to the the closed form:
\begin{equation}
\elr{k} = \frac{\eta_k}{\sqrt{\sum_{i=0}^{k-1}\beta^{(k-1)-i}\|\nabla\mathcal{L}(\bu_i)\|^{2} + \beta^k v_0}}.
\end{equation}
Hence the final theorem, when identifying the closed-form expressions of effective learning rates and using Lemma~\ref{lm:eq_scheme}. 
\end{proof}

\subsubsection{Validity of the assumptions in Theorem~\ref{th:eq_sgd}}\label{tayler_exp_res}
\textbf{Validity of the Taylor expansion.}
\begin{figure}[t]
\centering
\includegraphics[width=0.8\columnwidth]{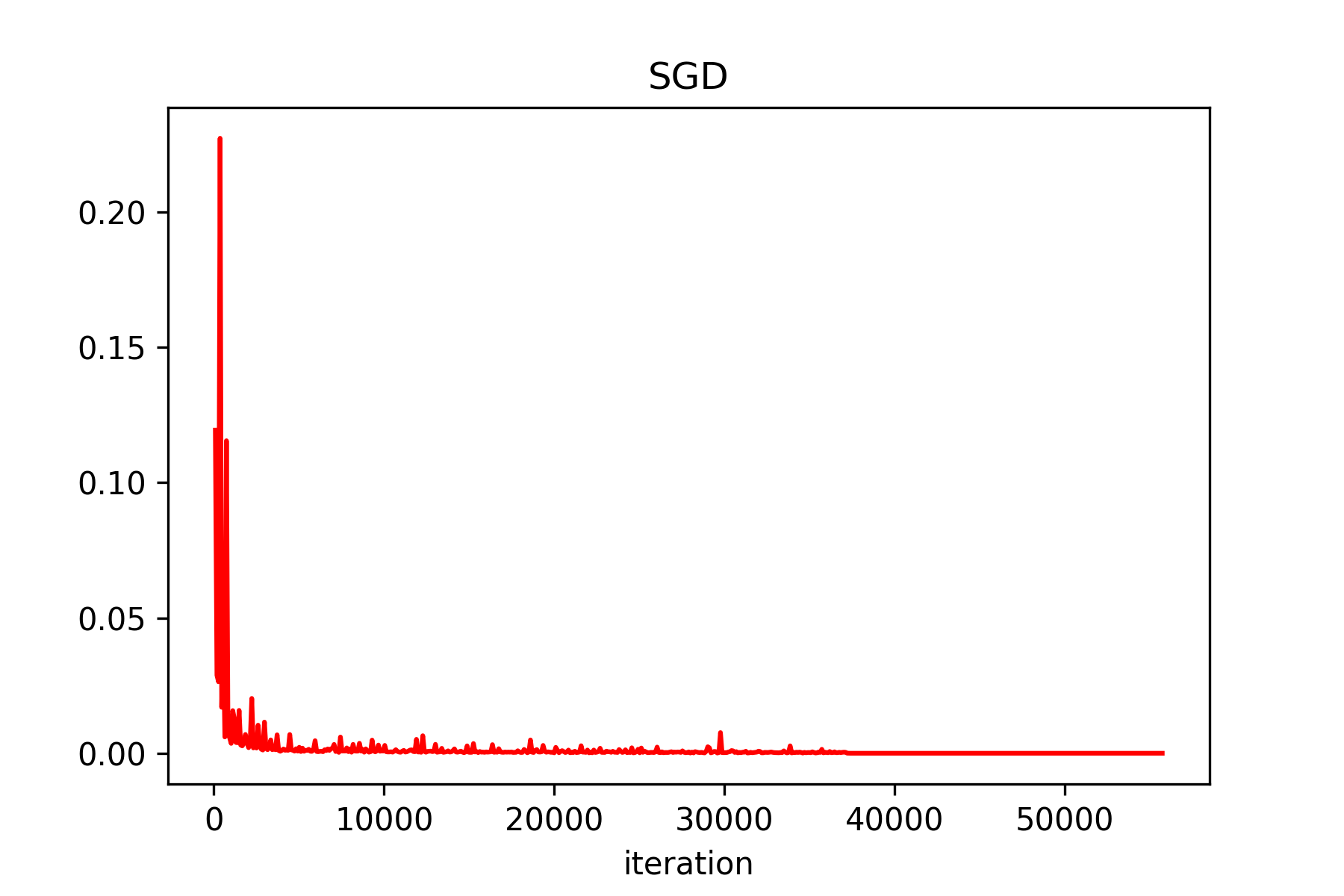}
\caption{\textbf{Validity of Taylor expansion}. We tracked the maximum value of $(\eta_k\|\gradsph\|)^2 / r_k^2$ for all filters in all layers of a ResNet20 CIFAR trained on CIFAR10 with SGD. The order of magnitude of the gradient is roughly the same for other architectures or datasets. It empirically validates the approximation by the Taylor expansion.}
\label{fig:taylor}
\end{figure}
For a CNN trained with SGD optimization, we tracked the quantity $(\eta_k\|\gradsph\|)^2 / r_k^2$, which is the variable of the Taylor expansion. As can be seen in Figure~\ref{fig:taylor}, the typical order of magnitude is $10^{-2}$, justifying the Taylor expansion.

A quick formal analysis also suggests the validity of this hypothesis. Thanks to the expression of $\eta_k = (1-\eta\lambda)^{-2i-k}\eta$ shown in the previous section, if we replace $\|\nabla\mathcal{L}(\bu_k)\|$ by a constant for asymptotic analysis, the comparison becomes:
\begin{align}
    (1-\eta\lambda)^{-4k-2} &\ll (1-\eta\lambda)^{-2}\frac{1-(1-\eta\lambda)^{-4k}}{1- (1-\eta\lambda)^{-4}}\\
    1 &\ll  \frac{1 - (1-\eta\lambda)^{4k} }{(1-\eta\lambda)^{-4} - 1}.
\end{align}
It is asymptotically true.

\subsubsection{Implementation details of weight trajectory tracking}\label{track_details}

Due to the high non-convexity of the optimization landscape, we choose to start from a relatively stable point in the parameter space. The finetuning of each architecture (ResNet20 BN, ResNet20 BN w/o affine and ResNet WN) starts from previously trained architectures on CIFAR10 via a simple SGD with an initial learning rate of $10^{-1}$, a $L_2$-regularization parameter of $10^{-4}$ and a momentum parameter of~$0.9$. The training is performed during 200 epochs, and the learning rate is multiplied by $0.1$ at epochs $80$, $120$ and $160$.

Then we track the trajectory obtained with SGD, AdaGrad and AdaGradG. The effective learning rate for SGD is fixed to $10^{-2}$ and the $L_2$-regularization parameter is set to $10^{-3}$ during finetuning. It gives us the following equivalent parameters for AdaGradG: order-2 moment parameter $\beta \approx 0.99996$ and learning rate $\eta \approx 0.71$. Since the effective direction is the same for both SGD and AdaGrad (Adam without momentum), in order to have the same order of  magnitude for the gradient steps we need to have effective learning rates of same order of magnitude. From Table~\ref{eff_lr_scheme}, in the case of SGD we have ${\elr{k}}_{\text{SGD}} = \frac{\eta_k}{r_k^2}$, and in the case of AdaGrad we have ${\elr{k}}_{\text{AdaGrad}} = \frac{\eta_k}{r_k \nu_k} = \frac{\eta_k}{r_k^2 \invsqf\|\bb_k\|} = {\elr{k}}_{\text{SGD}} \frac{1}{\invsqf\|\bb_k\|}$.We track the quantity $\frac{1}{\invsqf\|\bb_k\|}$ during training, which is roughly in the order of magnitude of $10^{-1}$. Therefore, to have gradient steps of equivalent order of magnitude between SGD and AdaGrad, we have to choose a learning rate of $10^{-3}$ for AdaGrad.

\section{Results in Section 4 (Geometric phenomena in Adam optimization)}

\subsection{Results in Section~\ref{subsec:empirical} (Identification of geometrical phenomena in Adam)}\label{ck_decomp}

\textbf{Decomposition of the effective direction.}
We decompose the effective direction as a gradient term and an $L_2$ regularization term:
\begin{align}
   \boxed{\bc_k^{\text{grad}} = \nabla \mathcal{L}(\bu_k) + \sum_{i=0}^{k-1}\beta^{k-i}\frac{r_k}{r_i}\nabla\mathcal{L}(\bu_i),} \tag{\eqref{eq:ckL2}}
\end{align}
\begin{align}
    \boxed{\bc_k^{L_2} = \bu_k + \sum_{i=0}^{k-1}\beta^{k-i}\frac{r_i}{r_k}\bu_i.} \tag{\eqref{eq:ckL2}}
\end{align}
Note that these expressions highlight the main terms at step $k$ and the dependency on $r_i$.

Developing the recurrence in Eq~\,\eqref{eq1b}, we obtain:
\begin{align}
    \ba_k &= \sum_{i=0}^{k}\beta^{k-i}\left(\nabla\mathcal{L}(\bx_i) + \lambda\bx_i  \right).
\end{align}
Using Lemma \ref{lem_invar} and decomposing on $\nabla\mathcal{L}(\bu_i)$ and $\bu_i$, we have:
\begin{align}
    \ba_k &= \sum_{i=0}^{k}\beta^{k-i}\left(\frac{1}{r_i}\nabla\mathcal{L}(\bu_i) + \lambda r_i\bu_i  \right)\\
    &= \frac{1}{r_k}\left(\sum_{i=0}^{k}\beta^{k-i}\left(\frac{r_k}{r_i}\nabla\mathcal{L}(\bu_i) + \lambda r_k r_i\bu_i  \right)\right).
\end{align}
Thus:
\begin{align}
    r_k\ba_k &= \sum_{i=0}^{k}\beta^{k-i}\frac{r_k}{r_i}\nabla\mathcal{L}(\bu_i) + \lambda r_k^2 \sum_{i=0}^{k}\beta^{k-i}\frac{r_i}{r_k}\bu_i,
\end{align}
which leads to the expression of $\bc_k^{\grad}$ and $\bc_k^{L_2}$ when we define $\bc_k \eqdef r_k \ba_k \oslash \disto$ (Eq.\,\eqref{lredef}).
\\
\hspace{1em}

\subsection{Results in Section~\ref{subsec:empirical} (Empirical study)}\label{proof:groupwise}\label{rt_precision}
\textbf{Clarification on Adam without deformation of gradients (a).} 
Following Theorem~\ref{thm:sphere_step}, the division vector $\bb_k$ has two contributions in the decomposition:
\begin{itemize}
    \item a deformation in $\bc_k$ applied to $\ba_k$: $\bc_k = r_k\ba_k \oslash \disto$;
    \item a scheduling effect in the effective learning rate $\invsqf \| \bb_k \|$ (Eq.\,\eqref{lredef}).
\end{itemize}
The goal is to find a new division vector $\operS(\bb_k)$ that does not create a deformation while preserving the scheduling effect of $\bb_k$ in the effective learning rate. This means:
\begin{align}
    \frac{\operS(\bb_k)}{\invsqf \|\operS(\bb_k)\|} = [1\cdots 1]^\top ,\\
    \invsqf  \|\operS(\bb_k)\| = \invsqf  \|\bb_k\|.
\end{align}
This leads to $\operS(\bb_k) = \invsqf \|\bb_k\|[1\cdots 1]^\top$. 

In the case of $\beta_1 = 0$, $\ba_k = \nabla \mathcal{L}(\bx_k)$, for any $\bb_k$. When we apply the standardization, we obtain:
\begin{equation}
    \bc_k = r_k \nabla\mathcal{L}(\bx_k) \oslash \frac{\operS(\bb_k)}{\invsqf \|\operS(\bb_k)\|} = \nabla\mathcal{L}(\bu_k) \oslash [1\cdots 1]^\top = \nabla\mathcal{L}(\bu_k). 
\end{equation}

The direction lies in the tangent space because, by Lemma \ref{lem_invar}, the gradient belongs to it.

In the generic scheme, using the standardization gives:
\begin{align}
    \bx_{k+1} & = \bx_{k} - \eta_{k} \ba_{k} \oslash\operS(\bb_{k})\\
     & = \bx_{k} - \eta_{k} \ba_{k} \oslash(\invsqf \|\bb_k\|[1\cdots 1]^\top)\\
     & = \bx_{k} - \eta_{k} \ba_{k} / (\invsqf \|\bb_k\|).
\end{align}
This means that the standardization consists in replacing the Hadamard division by $\bb_k$ with a scalar division by $\invsqf \|\bb_k\|$.

In the case of Adam, we recall that: 
\begin{equation}
    \bb_k = \frac{1 - \beta_1^{k+1}}{1 - \beta_1}\sqrt{\frac{\bv_k}{1 - \beta_2^{k + 1}} + \epsilon}~.
    \tag{\eqref{adam_algo}}
\end{equation}
Omitting $\epsilon$ for simplicity we have:
\begin{equation}
    \invsqf\|\bb_k\| = \frac{1 - \beta_1^{k+1}}{1- \beta_1} \left(\frac{1}{1 - \beta_2^{k+1}}\right)^{\frac{1}{2}}\invsqf \|\sqrt{\bv_k}\|. 
    \label{scal_div}
\end{equation}
Let us calculate $\|\sqrt{\bv_k}\|$. Developing the recursion of $\bv_k$, as defined in Eq.\,\eqref{preq2}, leads to:
\begin{align}
\bv_k &= (1-\beta_2) \sum_{i=0}^{k}\beta_2^{k-i}\left(\nabla\mathcal{L}(\bx_i) + \lambda\bx_i \right)^2, \label{eq:vk}\\
\sqrt{\bv_k}&= \sqrt{1 - \beta_2} \sqrt{\sum_{i=0}^{k}\beta_2^{k-i}\left(\nabla\mathcal{L}(\bx_i) + \lambda\bx_i \right)^2},
\end{align}
where the square and the square-root are element-wise operations. Hence, if we take the square norm:
\begin{align}
\|\sqrt{\bv_k}\|^2 & = (1 - \beta_2) \sum_{j=1}^{d} \left(\sqrt{\sum_{i=0}^{k}\beta_2^{k-i}\left(\nabla\mathcal{L}(\bx_i) + \lambda\bx_i \right)^2}\right)_j^2  \nonumber\\
 &=(1 - \beta_2) \sum_{j=1}^{d}\sum_{i=0}^{k}\beta_2^{k-i}\left(\nabla\mathcal{L}(\bx_i) + \lambda\bx_i \right)_{j}^2 \nonumber\\
& = (1 - \beta_2)\sum_{i=0}^{k}\beta_2^{k-i}\sum_{j=1}^{d}\left(\nabla\mathcal{L}(\bx_i) + \lambda\bx_i \right)_{j}^2 \nonumber\\
& = (1 - \beta_2)\sum_{i=0}^{k}\beta_2^{k-i}\|\nabla\mathcal{L}(\bx_i) + \lambda\bx_i \|^2 \label{calcul_vk},
\end{align}
where the $j$ subscript denotes the $j$-th element of the vector. It is exactly the order-2 moment of the gradient norm.

Therefore, we define the scalar $v_k$:
\begin{equation}
v_k = \beta_2 v_{k-1} + (1- \beta_2)\invf\|\nabla \mathcal{L}(\bx_{k}) + \lambda \bx_k\|^2,
\end{equation}
which is the order-2 moment of the gradient norm with a factor $\invf$. It verifies $\sqrt{v_k} = \invsqf \|\sqrt{\bv_k}\|$, needed for the scalar division stated in Eq.\,\eqref{scal_div}. By applying the bias correction, it gives the formula given in the paper of Adam w/o (a):
\begin{align}
    \bx_{k+1} &= \bx_{k} - \eta_{k}  \frac{\bmm_k}{1 - \beta_1^{k + 1}} / \sqrt{\frac{v_k}{1 - \beta_2^{k + 1}} + \epsilon}, \label{preq1SRTappendix} 
    \\
    \bmm_k &= \beta_1 \bmm_{k-1}\! + \! (1 - \beta_1)(\nabla\mathcal{L}(\bx_{k}) + \lambda \bx_k),
    \\
    v_k &= \beta_2 v_{k-1} + (1- \beta_2)\invf\|\nabla \mathcal{L}(\bx_{k}) + \lambda \bx_k\|^2. \label{preq2S}
\end{align}
Note that the previous demonstration makes the factor $\invf$ appear in $v_k$ to have exactly the scheduling effect of Adam without the deformation.

\textbf{Clarification on Adam without deformed gradients and no additional radial terms (ab).}\label{rt_transform}
We introduce the rescaling and transport transformation of the momentum to neutralize the identified effects on the effective direction (cf. Section~\ref{subsec:empirical}). The resulting, new $\bc_k$ is orthogonal to $\bu_k$ and does not contribute in the effective learning rate tuning with its radial part.

To avoid gradient history leaving the tangent space and thus neutralize \textbf{(b)}, we perform a parallel transport of the momentum $\ba_{k-1}$ from the corresponding point on the sphere $\mathbf{u}_{k-1}$ to the new point $\mathbf{u}_{k}$ denoted as $\Gamma_{\mathbf{u}_{k-1}}^{\mathbf{u}_{k}}(\mathbf{a}_{k-1})$ at each iteration $k \geq 1$. Figure\,\ref{fig:impact_radial_main}(c) illustrates the transport of a gradient. The parallel transport between two points associates each vector of the tangent space of the first point to a vector of the second tangent space by preserving the scalar product with the derivatives along the geodesic.
Consequently, the gradients accumulated in the resulting momentum now lie in the tangent space of $\bu_k$ at each step. This neutralizes the additional radial terms phenomena from $\bc_k^{\text{grad}}$.
Since $\bu_{k-1}$, $\bu_k$ and $\ba_k$ are coplanar, the transport of the momentum on the hypersphere can be expressed as a rotation:
\begin{align}
    \operT(\ba_{k-1}) &\eqdef \Gamma_{\mathbf{u}_{k-1}}^{\mathbf{u}_{k}}(\ba_{k-1}) \!=\! \langle \mathbf{u}_{k-1},\mathbf{u}_{k}\rangle\ba_{k-1} -\langle \ba_{k-1},\mathbf{u}_{k}\rangle\mathbf{u}_{k-1},\label{rotation}\\
    \ba_k&=\beta\operT(\ba_{k-1})  + \nabla\mathcal{L}(\bx_k) + \lambda\bx_k.
\end{align}
Although the transport operation is strictly defined on the tangent space only, the scalar product formulation enables its extension to the whole space. The transformation is linear and $\operT(\mathbf{u}_{k-1})=0$. We thus have:
\begin{equation}
\operT(\ba_{k-1} - \lambda\mathbf{u}_{k-1}) = \operT(\ba_{k-1}).
\end{equation}
In the previous formulation, we see that the $L_2$ component is not transported and does not contribute in the new momentum. Finally, the momentum only contains the contribution of the current $L_2$ regularization. This means that the $\operRT$ transformation decouples the $L_2$ regularization and thus neutralizes the additional radial terms from $\mathbf{c}_k^{L_2}$.

\textbf{Clarification on Adam without deformed gradients, no additional radial terms and no radius ratio (abc).}
To avoid the ratio $\frac{r_k}{r_i}$ in the effective learning direction and thus to cancel (\textbf{c}), we rescale the momentum in the update by the factor $\frac{r_{k-1}}{r_{k}}$ at each iteration $k \geq 1$.
From Lemma.~\ref{lem_invar}, we obtain:
 \begin{align}
    \operR(\ba_{k-1}) &\eqdef \frac{r_{k-1}}{r_k}\ba_{k-1}\\
     \ba_k &= \beta \operR(\ba_{k-1}) + \nabla\mathcal{L}(\bx_k) + \lambda\bx_k \\
     &=\frac{1}{r_k}\Big(\sum_{i=0}^{k}\beta^{k-i}(\nabla\mathcal{L}(\bu_k) + \lambda r_k r_i \bu_i)\Big).
 \end{align}
Note that now, the factor $\frac{r_k}{r_i}$ is not contained anymore in the gradient contribution of $\bc_k = r_k\ba_k$, which neutralizes the radius ratio phenomenon. 

We can note that $\operR$ and $\operT$ are commutative and that we can combine them in a simple concise scalar expression:
\begin{align}\label{eq:rt_update}
    \operRT(\ba_{k-1}) &\eqdef \frac{\langle\bx_k,\bx_{k-1}\rangle\ba_{k-1} - \langle\bx_k,\ba_{k-1}\rangle \bx_{k-1}}{\langle\bx_k,\bx_k\rangle},\\
    \ba_k &= \beta\operRT(\ba_{k-1}) + \nabla\mathcal{L}(\bx_k) + \lambda\bx_k.
\end{align}
This new momentum leads to $\bc_k = \bc_k^{\operRT} + r_k^2\lambda\bu_k$ with $\ckrad = \lambda r_k^2$ and $\bc_k^{\perp} = \bc_k^{\operRT}$. The latter relies only on the trajectory on the hypersphere and always lies in the tangent space:
\begin{equation}
    \mathbf{c}^{\operRT}_{k} = \beta\Gamma_{\mathbf{u}_{k-1}}^{\mathbf{u}_{k}}(\mathbf{c}^{\operRT}_{k-1}) + \nabla\mathcal{L}(\mathbf{u}_{k}).
\end{equation}
The final Adam w/o (abc) scheme reads:
\begin{align}
    \bx_{k+1} &= \bx_{k} - \eta_{k}  \frac{\bmm_k}{1 - \beta_1^{k + 1}} / \sqrt{\frac{v_k}{1 - \beta_2^{k + 1}} + \epsilon}, 
    \\
    \bmm_k &= \beta_1 \operRT(\bmm_{k-1})\! + \! (1 - \beta_1)(\nabla\mathcal{L}(\bx_{k}) + \lambda \bx_k),
    \\
    v_k &= \beta_2 \frac{r_{k-1}^2}{r_k^2}v_{k-1} + (1- \beta_2)\invf\|\nabla \mathcal{L}(\bx_{k}) + \lambda \bx_k\|^2.
\end{align}
We also rescale the introduced scalar $v_k$ at each step with the factor $ \frac{r_{k-1}^2}{r_k^2}$. This removes the radius from the gradient contribution of the scheduling $\nu^{\operR}=r_k v_k$, in contrast with $\nu_k$ from Eq.\,\eqref{nuk}. The new scheduling effect reads:
\begin{equation}
\boxed{
    \nu_k^\operR = \invsqf \frac{1 - \beta_1^{k+1}}{1 - \beta_1}\sqrt{\frac{1 - \beta_2}{1 - \beta_2^{k + 1}}} \nonumber  \Big(\sum_{i=0}^{k}\beta_2^{k-i}\|\nabla\mathcal{L}(\bu_i)+\lambda r_i r_k\bu_i\|^2\Big)^{1/2}
}.    \label{dkrt}    
\end{equation}

\subsection{Training and implementation details}\label{training_implem_details}

To assess empirically the significance of the above phenomena in the context of CNNs with BN and BN w/o affine, we evaluate the different variants of AdamW, AdamG, Adam w/o (a), w/o (ab), w/o (abc) over a variety of datasets and architectures. 

Note that the set of parameters $\boldsymbol{\theta}$ of a CNN with NLs can be split in two disjoint subsets:
$\boldsymbol{\theta} = \mathcal{F} \cup \mathcal{R}$,
where $\mathcal{F}$ is the set of groups of radially-invariant parameters and $\mathcal{R}$ the remaining parameters. As demonstrated in   \ref{bn_rescale}, the subset $\mathcal{F}$
includes parameters of all filters followed by BN. Since we are only interested in comparing optimization on $\mathcal{F}$, Adam variants w/o (a), w/o (ab), w/o (abc), AdamW AdamG are applied only to the optimization of the parameters in $\mathcal{F}$ whereas the ones in $\mathcal{R}$ are optimized with the original Adam scheme. The algorithm of Adam w/o (abc) is illustrated in Algorithm\,\ref{algo_srt}.

For each optimization scheme, each dataset and each architecture, the same grid search range and budget was performed while mini-batch size was fixed. We used a mini-batch size of 128 for SVHN, CIFAR10 and CIFAR100. The learning rates $\eta$ varied in $\{10^{-4}, 10^{-3}, 10^{-2}, 10^{-1}\}$, the weight decay in $10^{-3} \cdot \{0, \frac{1}{128}, \frac{1}{64}, \frac{1}{32}, \frac{1}{32}, \frac{1}{16}, \frac{1}{8}, \frac{1}{4}\}$ (similar to \cite{loshchilov2019decoupled}), the momentum was fixed to $0.9$ ($\beta_1$ for variants of Adam) and the order-two moment $\beta_2$ in $\{0.99, 0.999, 0.9999\}$ (as in \cite{kingma2014adam}).

We used the same step-wise learning rate scheduler for each method. For SVHN, CIFAR10 and CIFAR100, models were trained for 405 epochs, and the learning rate multiplied by 0.1 at epochs 135, 225 and 315.

The  optimization schemes introduced in this paper do not change the complexity in time of the algorithm. During the update of parameters in a layer, we only do a temporary copy of the parameter tensor just before the update to perform the RT transformation. This temporary copy is flushed after the RT transformation. 
Nothing permanent is stored in the optimizer.

Note that, for each architecture and each dataset, the same learning rate was systematically found for each method while the momentum factor was fixed at $0.9$ (cf. Table~\ref{lr_gsearch}). Best other hyperparameters, i.e., $L_2$ regularization and order-2 moment, are shown in Table~\ref{tab:results-transposed}.

\begin{algorithm}[t!]
  \caption{\textbf{Adam w/o (abc)} and its algorithm illustrated for a filter $\bx \in \mathbb{R}^d$ followed by BN. Steps that are different from Adam are shown \adamsrttext{in highlight}.
  For non-convolutional layers we use standard Adam. 
  }\label{algo_srt}
\begin{algorithmic}[1]
   \REQUIRE $\beta_1, \beta_2 \in [0,1)$; $\lambda, \eta \in \mathbb{R}$; $\mathcal{L}(\bx)$
\STATE \textbf{initialize} step $k\leftarrow-1$; $\bmm_k \leftarrow 0$;  $v_k \leftarrow 0$; $\bx \in \mathbb{R}^{d}$ 
   \WHILE{\textit{stopping criterion not met}}
       \STATE $k \leftarrow k + 1$
       \STATE $\bg \leftarrow \nabla \mathcal{L}(\bx_{k}) + \lambda \bx_{k}$
       \STATE $\bmm_k\leftarrow \beta_1\bmm_{k-1} + (1-\beta_1)\bg$
       \STATE \adamsrttext{$v_k \leftarrow \beta_2 v_{k-1} + (1-\beta_2)\invf\bg^{\top}\bg$}
       \STATE $\hat{\bmm} \leftarrow \bmm_k/(1-\beta_1^{k+1})$ 
       \STATE $\hat{v} \leftarrow v_k/(1-\beta_2^{k+1})$
       \STATE $\bx_{k+1} \leftarrow \bx_k - \eta\hat{\bmm}/(\sqrt{\hat{v}} + \epsilon)$
       \STATE  \adamsrttext{$\bmm_k\leftarrow \bmm_k(\bx_{k+1}^{\top}\bx_k \bmm_k - \bmm_k^{\top}\bx_{k+1}\bx_k) / (\bx_{k+1}^{\top}\bx_{k+1})$}
       \STATE  \adamsrttext{$v_k\leftarrow v_k(\bx_{k}^{\top}\bx_{k} / \bx_{k+1}^{\top}\bx_{k+1}) $}
   \ENDWHILE
   \STATE \textbf{return} resulting parameters $\bx_k$
\end{algorithmic}
\vspace{-2pt}
\end{algorithm}

\subsection{Additional empirical results}\label{addresulst}

\subsubsection{Batch Normalization}

In this section, we observe the mean loss training curves associated to Adam, AdamW, AdamG, Adam w/o (a), Adam w/o (ab), Adam w/o (abc) on datasets CIFAR10, CIFAR100 and SVHN with architecture ResNet20, ResNet18 or VGG16 with BN, corresponding to the accuracies given in  Table~\ref{tab:results}. The parameter setting are specified in Table~\ref{tab:results-transposed}.  These curves are illustrated in  Figures~\ref{fig:train_comparison_resnet18_cifar10}~\ref{fig:val_comparison_resnet18_cifar10}-\ref{fig:train_comparison_vgg16_cifar10}~\ref{fig:train_comparison_vgg16_cifar10}~\ref{fig:val_comparison_vgg16_cifar10}~\ref{fig:train_comparison_resnet18_cifar100}~\ref{fig:val_comparison_resnet18_cifar100}~\ref{fig:train_comparison_vgg16_cifar100}~\ref{fig:val_comparison_vgg16_cifar100}~\ref{fig:train_comparison_resnet18_svhn}~\ref{fig:val_comparison_resnet18_svhn}~\ref{fig:train_comparison_vgg16_svhn}~\ref{fig:val_comparison_vgg16_svhn}. The case of ResNet20 is illustrated in Figure~\ref{fig:loss_comparison} of the paper.

\begin{table}[ht!]
\centering
\renewcommand{\figurename}{Table}
\renewcommand{\captionlabelfont}{\bf}
\caption{\textbf{Best learning rate and momentum factor.} We systematically found the same learning rate for each dataset and architecture while the momentum factor was fixed to 0.9.}\label{lr_gsearch}
\vspace{1mm}
\scalebox{0.8}
{
\begin{tabular}{ l  l  c }
\toprule
    \multicolumn{1}{l}{Method} & \multicolumn{1}{l  }{$\eta_0$} &  \multicolumn{1}{c}{$\beta$, $\beta_1$} \\
\midrule 
Adam  & $0.001$ & 0.9 \\
AdamW & $0.001$ & 0.9 \\
AdamG & $0.01$ & 0.9 \\
\midrule
Adam w/o (a) & $0.001$ & 0.9  \\
Adam w/o (ab) & $0.001$ & 0.9 \\
Adam w/o (abc) & $0.001$ & 0.9 \\
\bottomrule
\end{tabular}
}
\end{table}

\begin{table}[ht!]
\centering
\renewcommand{\figurename}{Table}
\renewcommand{\captionlabelfont}{\bf}
\caption{\textbf{Best $L_2$ regularization ($\lambda$) and order-2 moment factors ($\beta_2$).} 
}
\vspace{1mm}
\scalebox{0.7}
{
\begin{tabular}{l c @{~~~}c | c   c c  c  c  c}
\toprule
\multicolumn{3}{c|}{Setup}  & Adam & AdamW & AdamG  & Adam & Adam & Adam \\
& & & & & & w/o (a) & w/o (ab) & w/o (abc) \\
\midrule
\multirow{7}{*}{CIFAR10}  &  \multirow{2}{*}{\texttt{ResNet20}} & $\lambda\times10^4$ &  2.5 & 5 & 1.25 & 0.31 & 1.25 & 5  \\
  &  & $\beta_2$ & 0.99 & 0.99 & 0.99 & 0.99 & 0.99 & 0.99 \\
\cmidrule{4-9}
  & \multirow{2}{*}{\texttt{ResNet18}} & $\lambda\times10^4$ & 2.5 & 0.08 & 0.63 & 2.5 & 1.25 &  0.16  \\
  &  & $\beta_2$ & 0.999 & 0.99 & 0.99 & 0.99 & 0.99 & 0.99 \\
\cmidrule{4-9}
  & \multirow{2}{*}{\texttt{VGG16}} & $\lambda\times10^4$ & 2.5 & 0.31 & 2.5 & 0.63 & 0.00 & 0.31  \\
  &  & $\beta_2$ & 0.999 & 0.999 & 0.999 & 0.999 & 0.99 & 0.999 \\
\midrule
\multirow{4}{*}{CIFAR100} & \multirow{2}{*}{\texttt{ResNet18}} & $\lambda\times10^4$ & 1.25 & 1.25 & 1.25 & 1.25 & 1.25 & 0.00   \\
  &  & $\beta_2$ & 0.999 & 0.99 & 0.99 & 0.99 & 0.99 & 0.999 \\
\cmidrule{4-9}
  & \multirow{2}{*}{\texttt{VGG16}} & $\lambda\times10^4$ & 0.63 & 0.16 & 0.63 & 0.63 & 1.25 & 0.08  \\
  &  & $\beta_2$ & 0.99 & 0.99 & 0.99 & 0.99 & 0.99 & 0.99 \\
\midrule
\multirow{4}{*}{SVHN} & \multirow{2}{*}{\texttt{ResNet18}} & $\lambda\times10^4$ & 0.00 & 0.08 & 5 & 0.31 & 5 & 0.08   \\
  &  & $\beta_2$ & 0.999 & 0.999 & 0.99 & 0.99 & 0.999 & 0.999 \\
\cmidrule{4-9}
  & \multirow{2}{*}{\texttt{VGG16}} & $\lambda\times10^4$ & 0.00 & 0.31 & 5 & 0.08 & 2.5 & 2.5   \\
  &  & $\beta_2$ &  0.99 & 0.99 & 0.99 & 0.99 & 0.99 & 0.999 \\
\bottomrule
\end{tabular}
}
\vspace{-2mm}
\label{tab:results-transposed}
\end{table}

\begin{figure}
    \renewcommand{\captionlabelfont}{\bf}
    \centering
    \vspace{-4mm}
    \subfigure{
    \includegraphics[width=0.48\columnwidth]{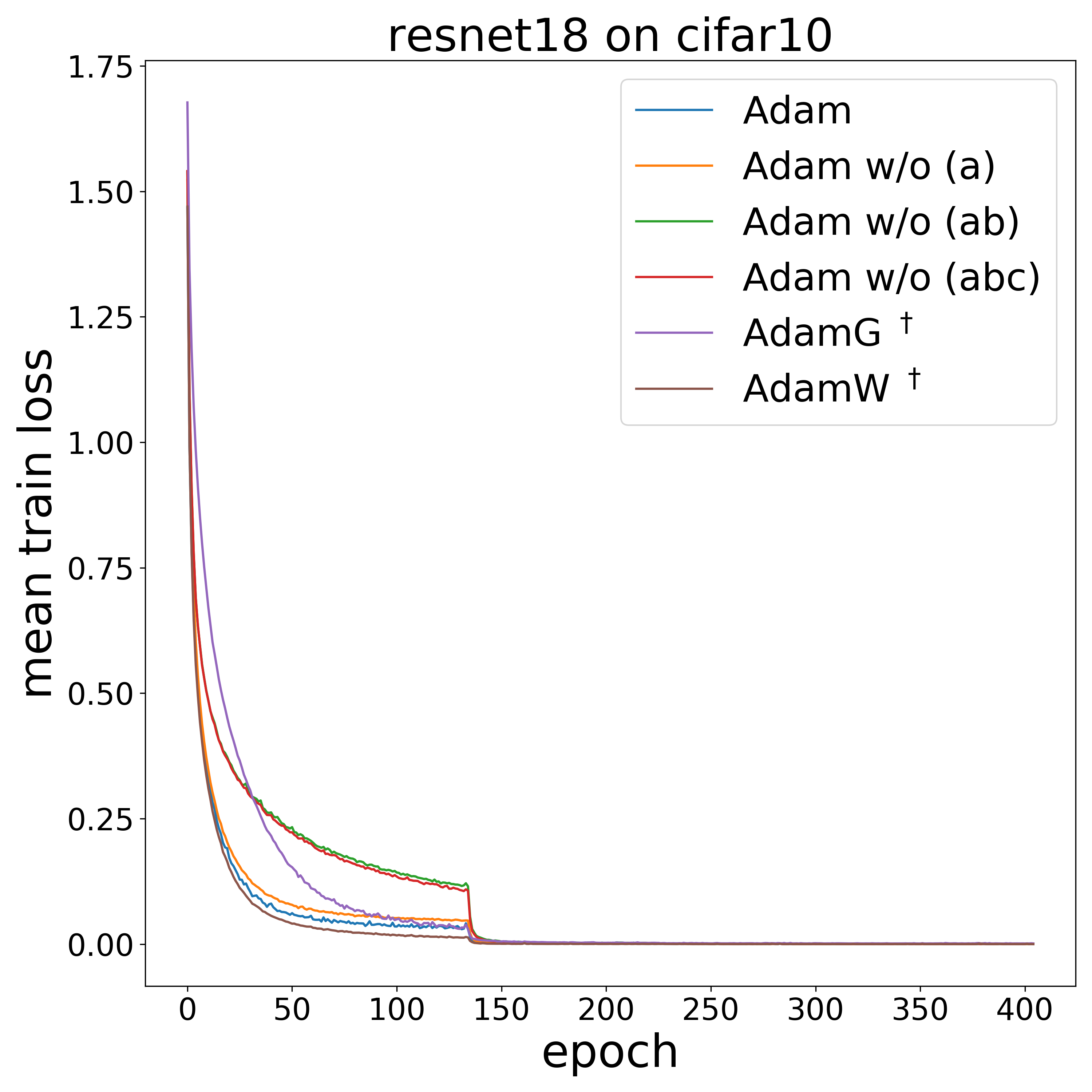}
    }
    \subfigure{
    \includegraphics[width=0.48\columnwidth]{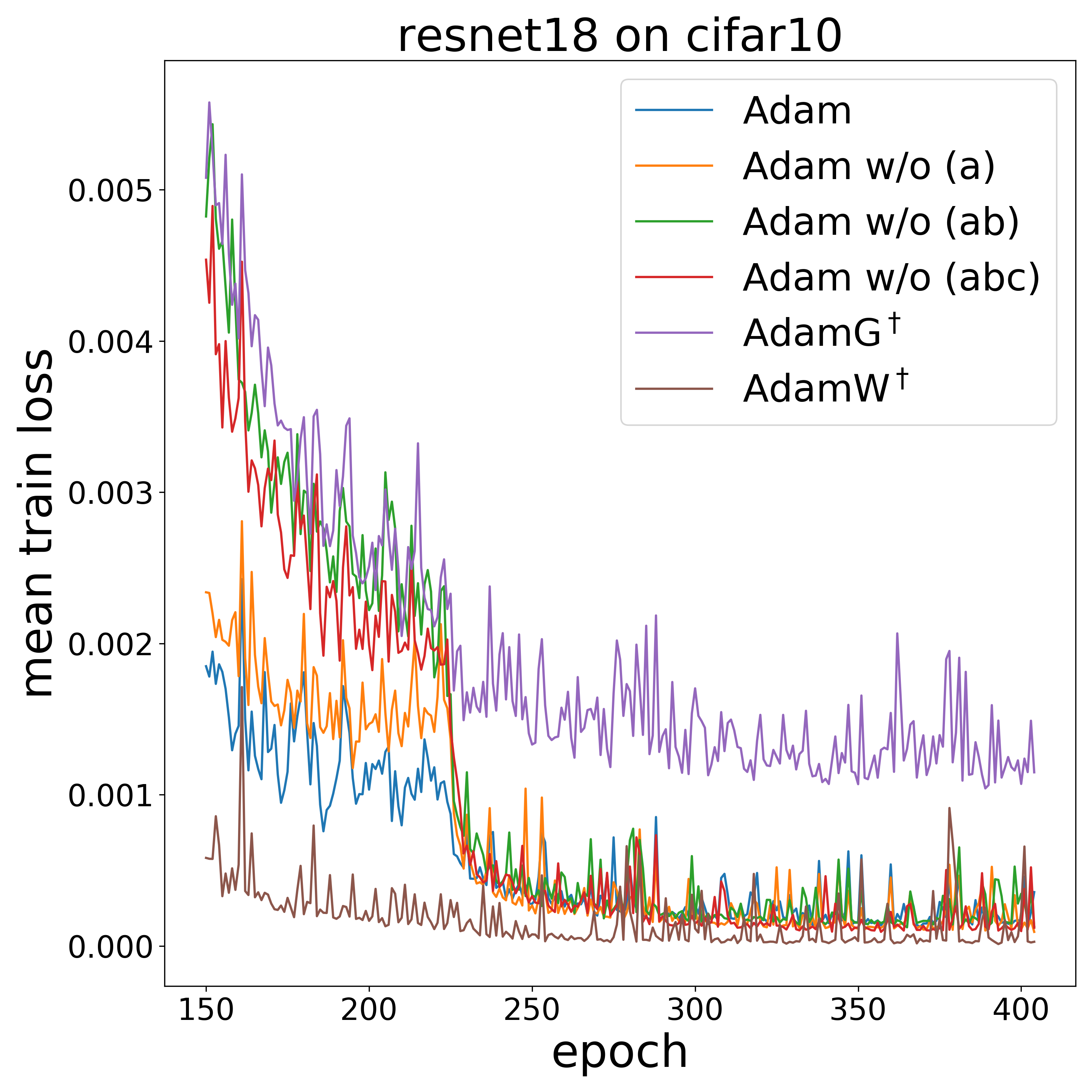}
    }
    \vspace{-2mm}
    \caption{\textbf{Training speed comparison with ResNet18 BN on CIFAR10.} \emph{Left:} Mean training loss over all training epochs (averaged across 5 seeds) for different Adam variants. \emph{Right:} Zoom-in on the last epochs. Please refer to Table~\ref{tab:results} for the corresponding accuracies.}
    \label{fig:train_comparison_resnet18_cifar10}
     \vspace{-2mm}
\end{figure}

\begin{figure}
    \renewcommand{\captionlabelfont}{\bf}
    \centering
    \vspace{-4mm}
    \subfigure{
    \includegraphics[width=0.48\columnwidth]{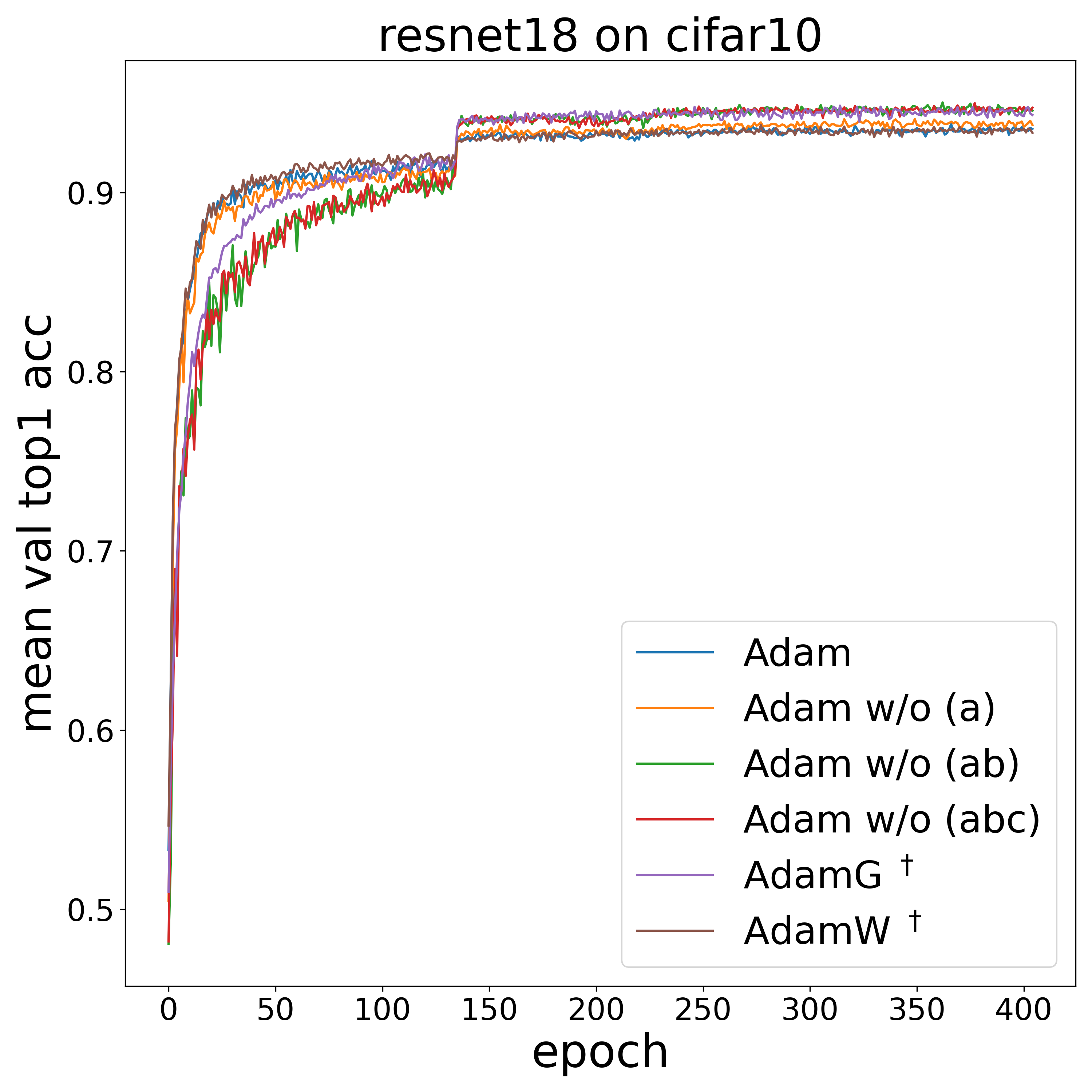}
    }
    \subfigure{
    \includegraphics[width=0.48\columnwidth]{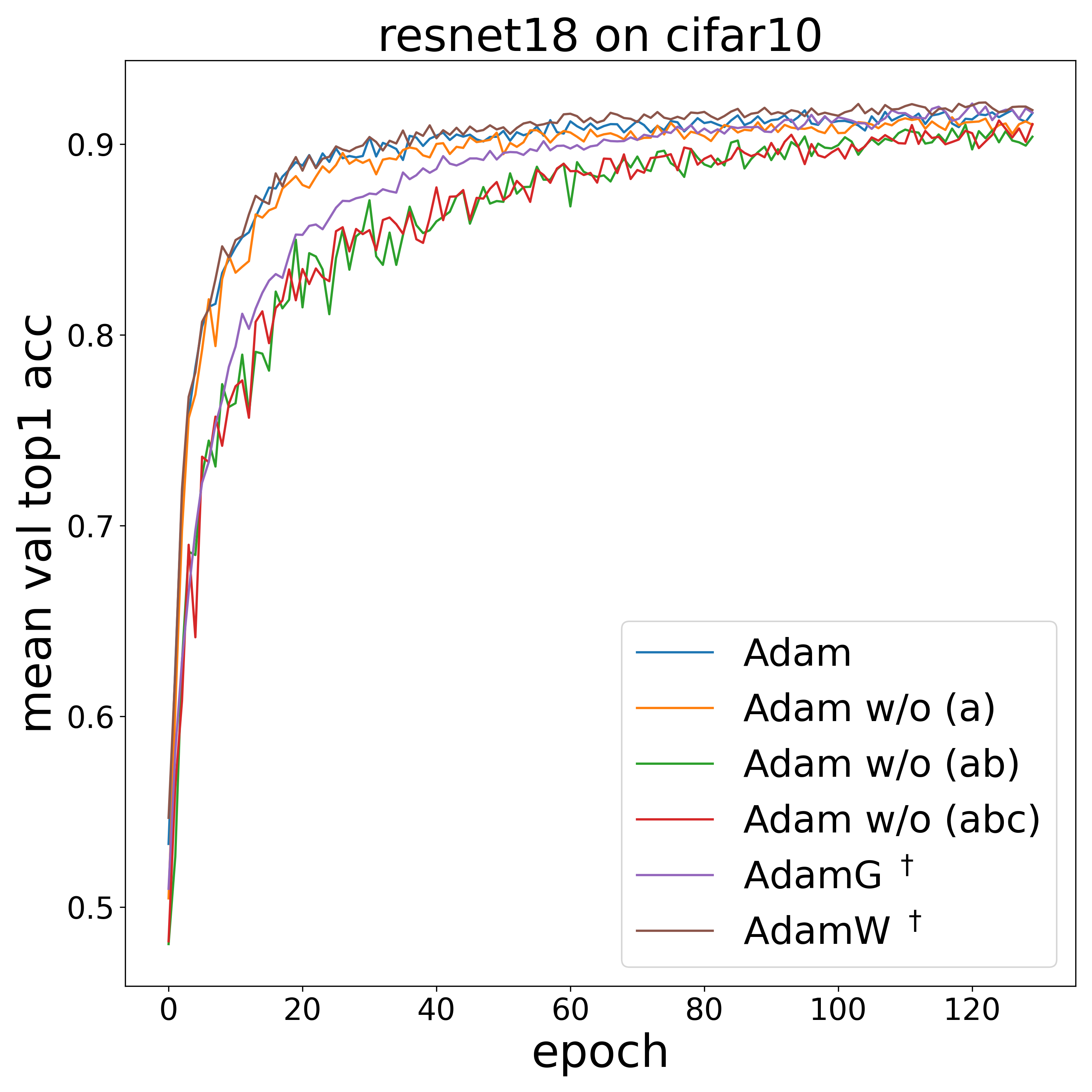}
    }
    \vspace{-2mm}
    \caption{\textbf{Accuracy comparison on the validation set with ResNet18 BN on CIFAR10.} \emph{Left:} Mean training loss over all training epochs (averaged across 5 seeds) for different Adam variants. \emph{Right:} Zoom-in on the first epochs. Please refer to Table~\ref{tab:results} for the corresponding accuracies.}
    \label{fig:val_comparison_resnet18_cifar10}
     \vspace{-2mm}
\end{figure}

\begin{figure}
    \renewcommand{\captionlabelfont}{\bf}
    \centering
    \vspace{-4mm}
    \subfigure{
    \includegraphics[width=0.48\columnwidth]{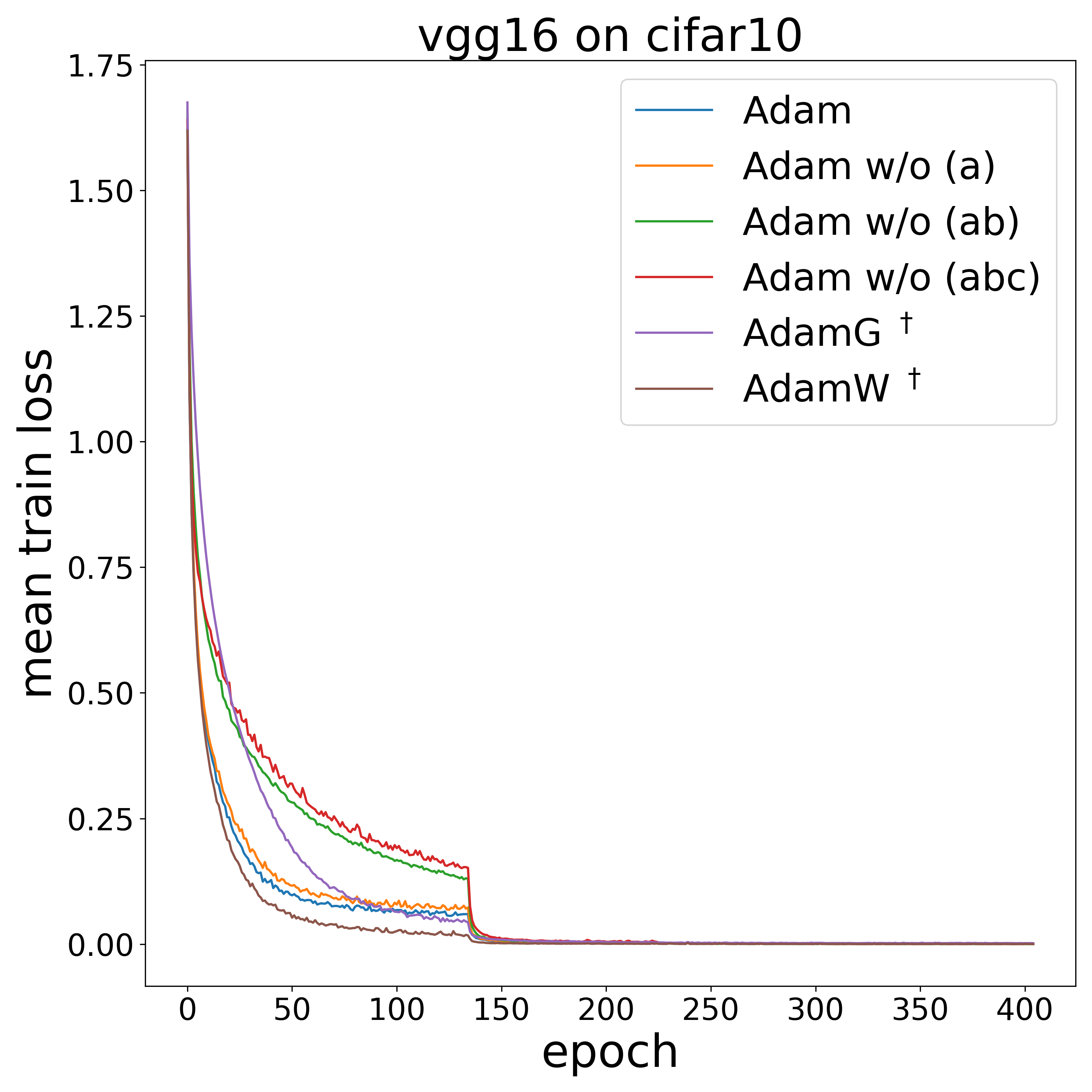}
    }
    \subfigure{
    \includegraphics[width=0.48\columnwidth]{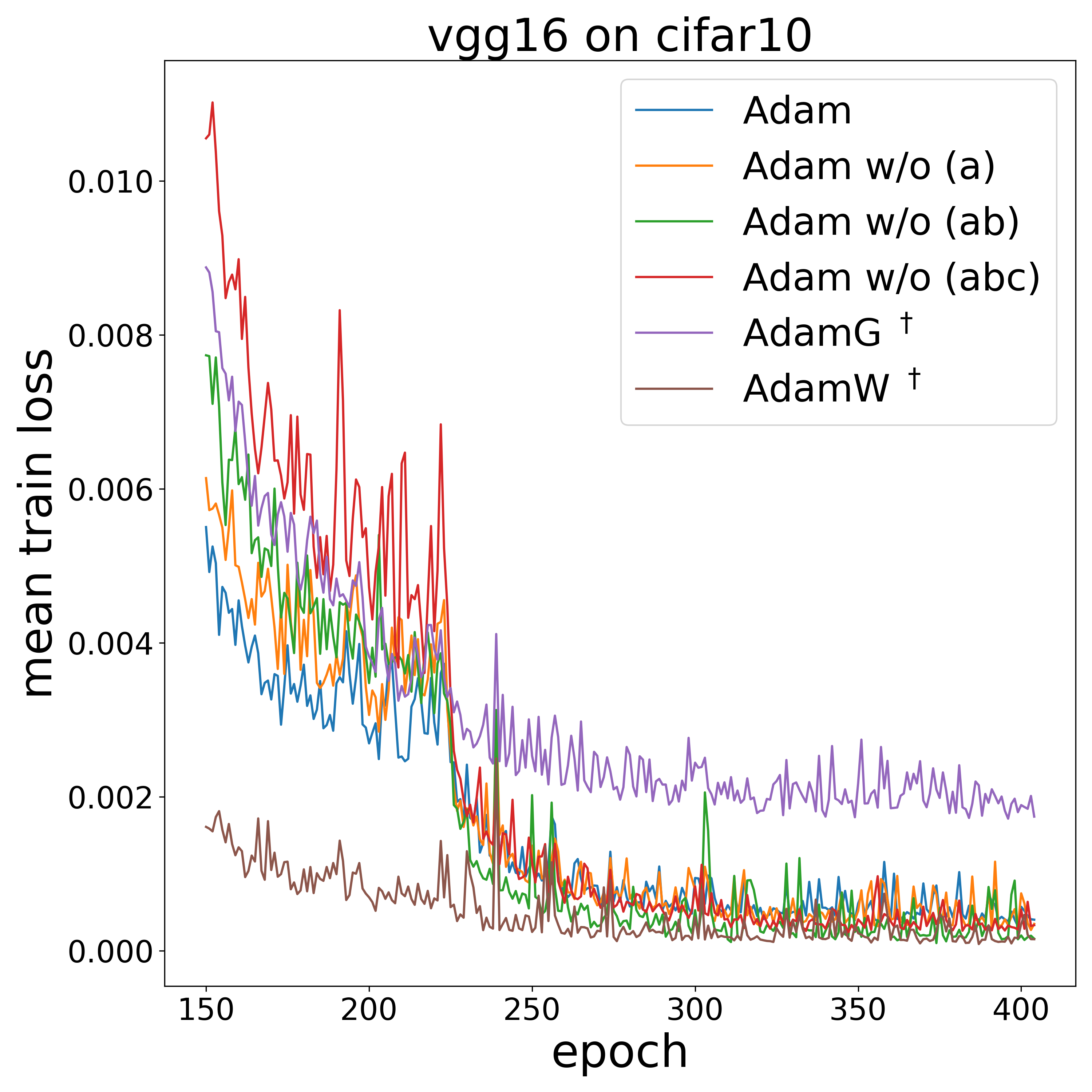}
    }
    \vspace{-2mm}
    \caption{\textbf{Training speed comparison with VGG16 on CIFAR10.} \emph{Left:} Mean accuracy on the validation set over all training epochs (averaged across 5 seeds) for different Adam variants. \emph{Right:} Zoom-in on the last epochs. Please refer to Table~\ref{tab:results} for the corresponding accuracies.}
    \label{fig:train_comparison_vgg16_cifar10}
     \vspace{-2mm}
\end{figure}

\begin{figure}
    \renewcommand{\captionlabelfont}{\bf}
    \centering
    \vspace{-4mm}
    \subfigure{
    \includegraphics[width=0.48\columnwidth]{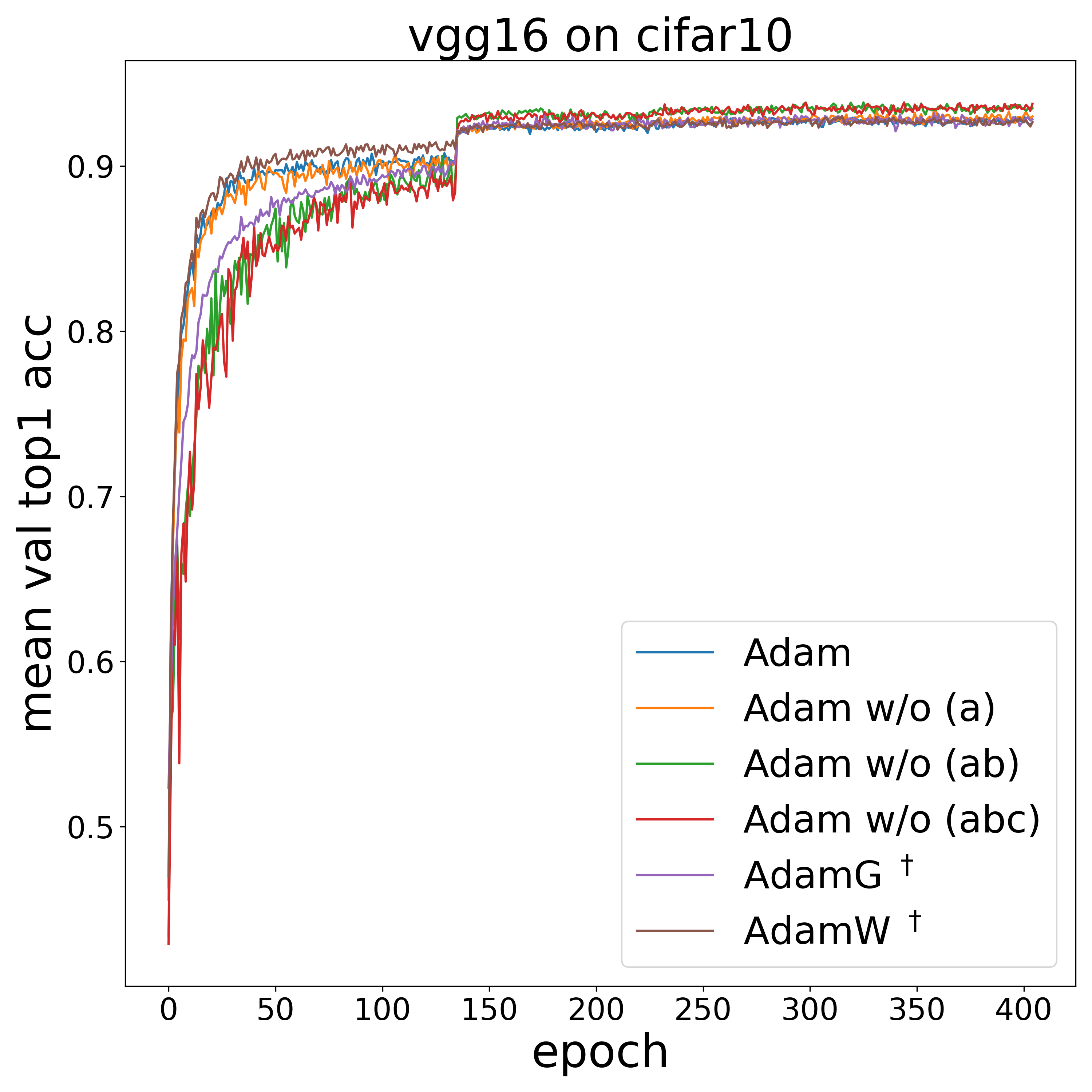}
    }
    \subfigure{
    \includegraphics[width=0.48\columnwidth]{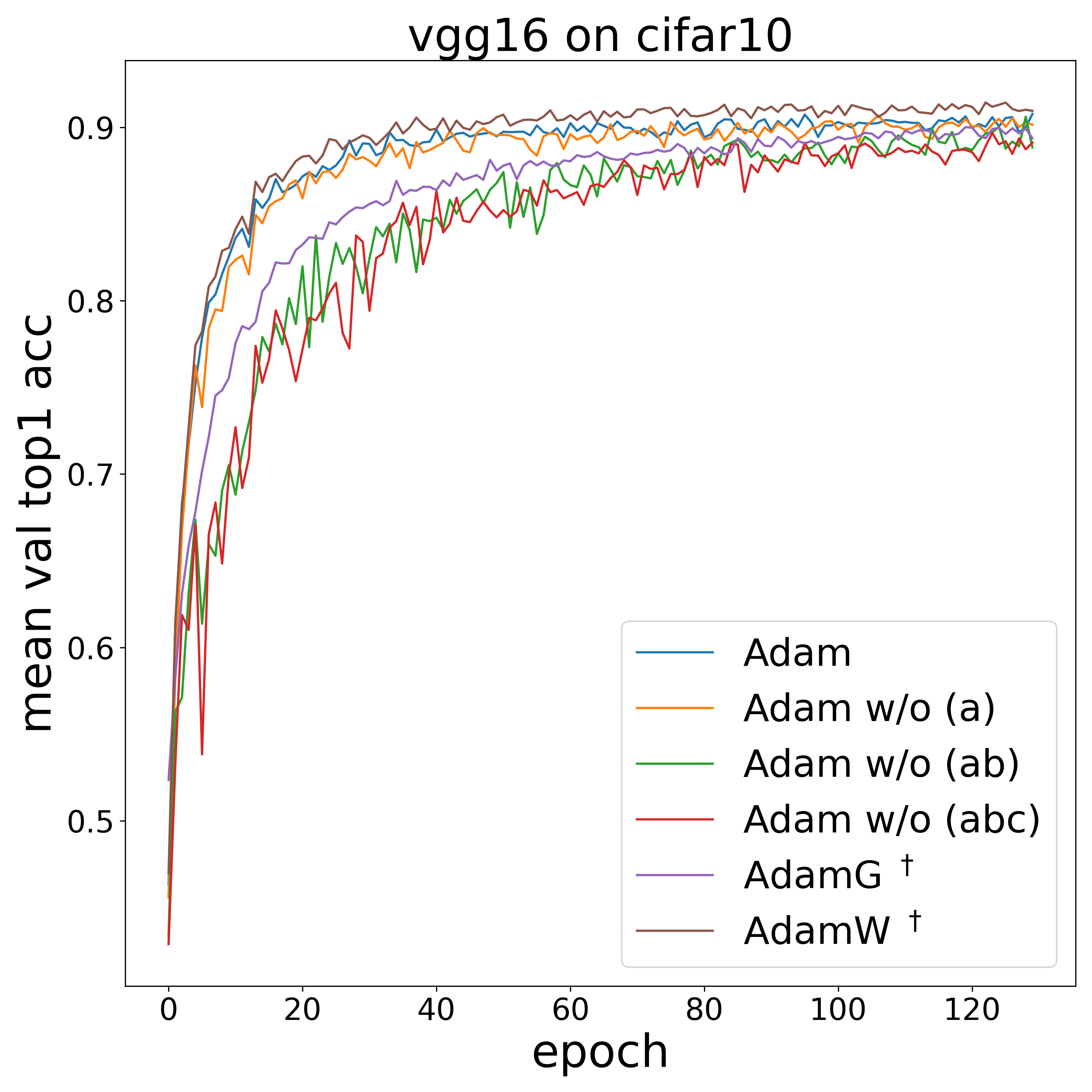}
    }
    \vspace{-2mm}
    \caption{\textbf{Accuracy comparison on the validation set with VGG16 BN on CIFAR10.} \emph{Left:} Mean training loss over all training epochs (averaged across 5 seeds) for different Adam variants. \emph{Right:} Zoom-in on the last epochs. Please refer to Table~\ref{tab:results} for the corresponding accuracies.}
    \label{fig:val_comparison_vgg16_cifar10}
     \vspace{-2mm}
\end{figure}

\begin{figure}
    \renewcommand{\captionlabelfont}{\bf}
    \centering
    \vspace{-4mm}
    \subfigure{
    \includegraphics[width=0.48\columnwidth]{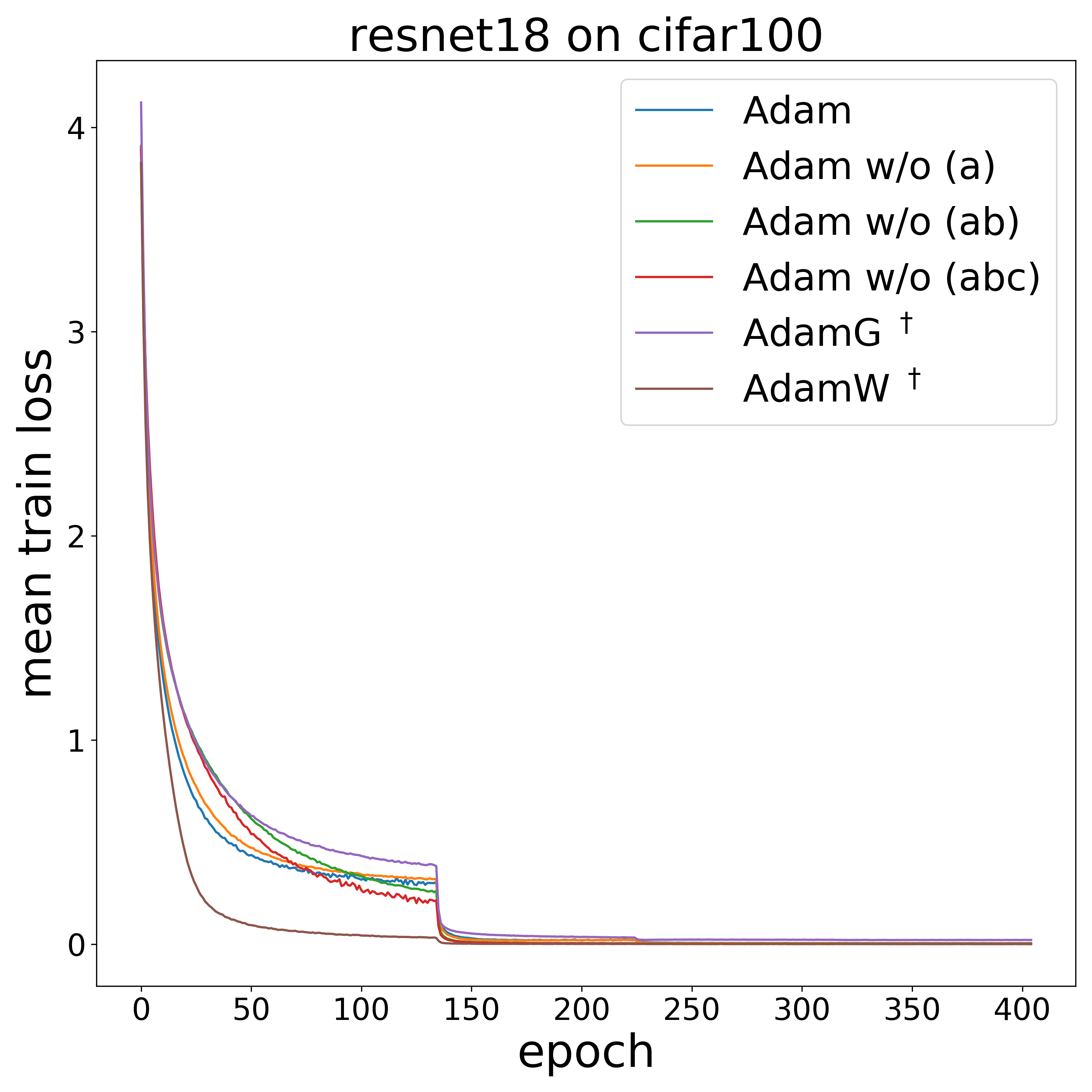}
    }
    \subfigure{
    \includegraphics[width=0.48\columnwidth]{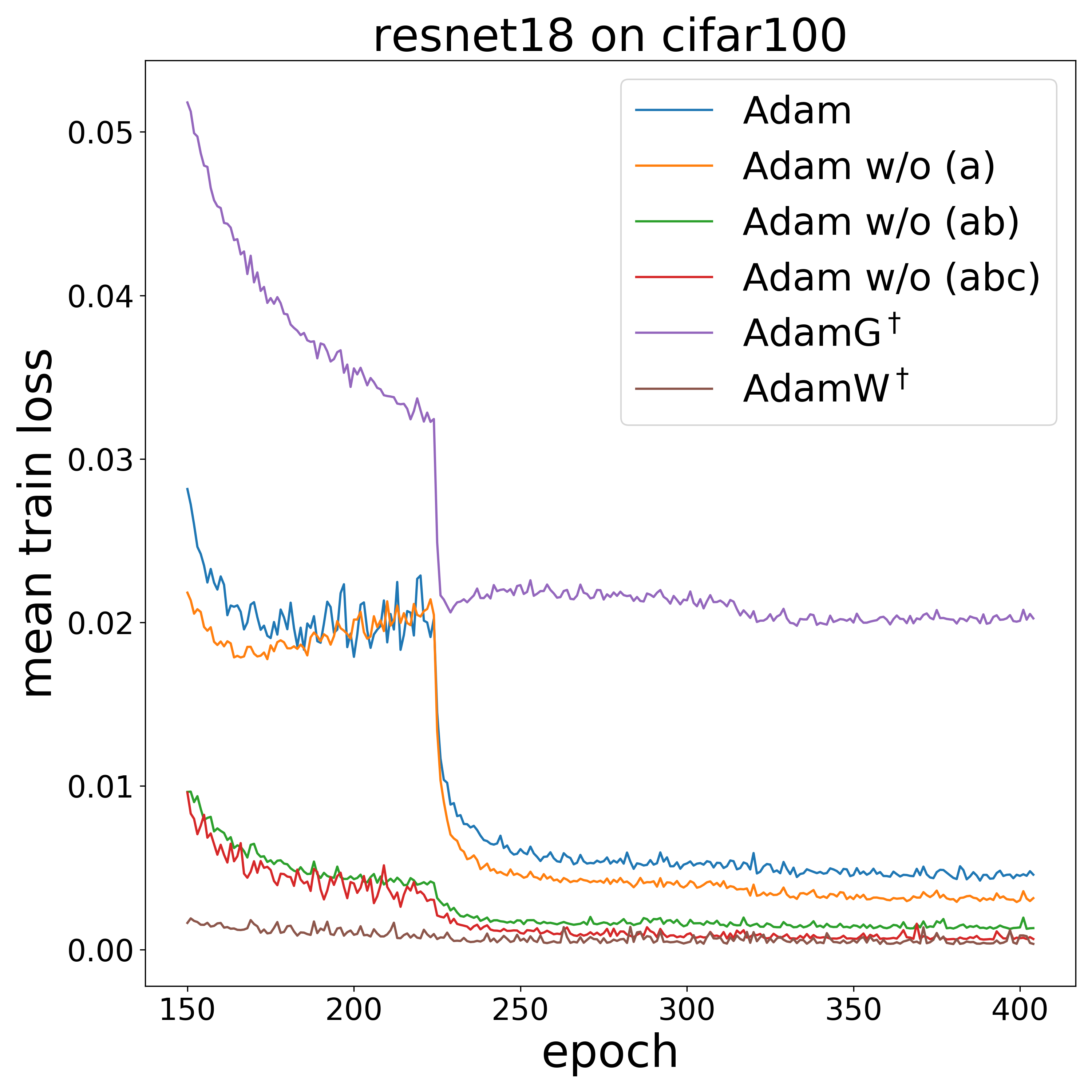}
    }
    \vspace{-2mm}
    \caption{\textbf{Training speed comparison with ResNet18 on CIFAR100.} \emph{Left:} Mean training loss over all training epochs (averaged across 5 seeds) for different Adam variants. \emph{Right:} Zoom-in on the last epochs. Please refer to Table~\ref{tab:results} for the corresponding accuracies.}
    \label{fig:train_comparison_resnet18_cifar100}
     \vspace{-2mm}
\end{figure}

\begin{figure}
    \renewcommand{\captionlabelfont}{\bf}
    \centering
    \vspace{-4mm}
    \subfigure{
    \includegraphics[width=0.48\columnwidth]{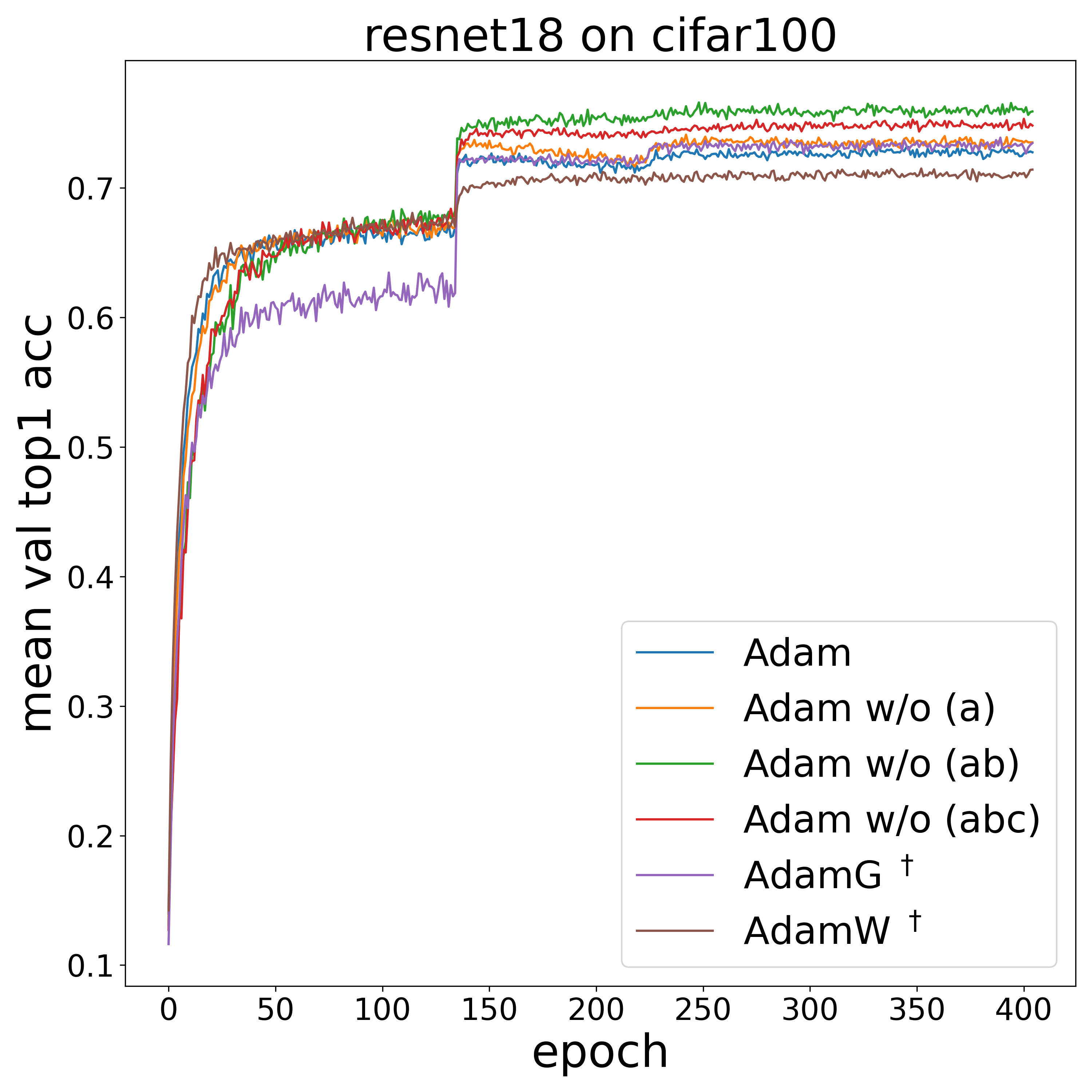}
    }
    \subfigure{
    \includegraphics[width=0.48\columnwidth]{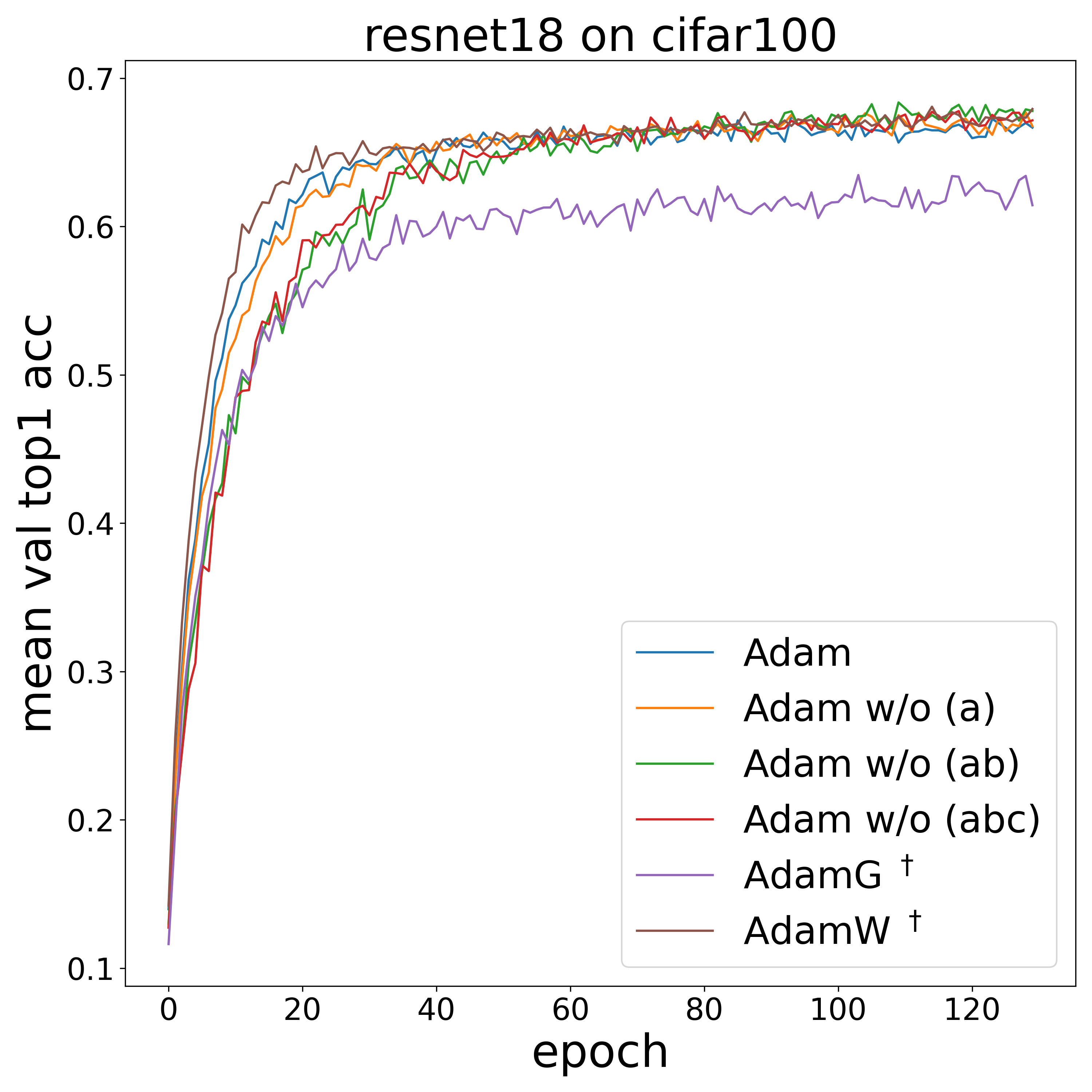}
    }
    \vspace{-2mm}
    \caption{\textbf{Accuracy comparison on the validation set with ResNet18 BN on CIFAR100.} \emph{Left:} Mean training loss over all training epochs (averaged across 5 seeds) for different Adam variants. \emph{Right:} Zoom-in on the last epochs. Please refer to Table~\ref{tab:results} for the corresponding accuracies.}
    \label{fig:val_comparison_resnet18_cifar100}
     \vspace{-2mm}
\end{figure}

\begin{figure}
    \renewcommand{\captionlabelfont}{\bf}
    \centering
    \vspace{-4mm}
    \subfigure{
    \includegraphics[width=0.48\columnwidth]{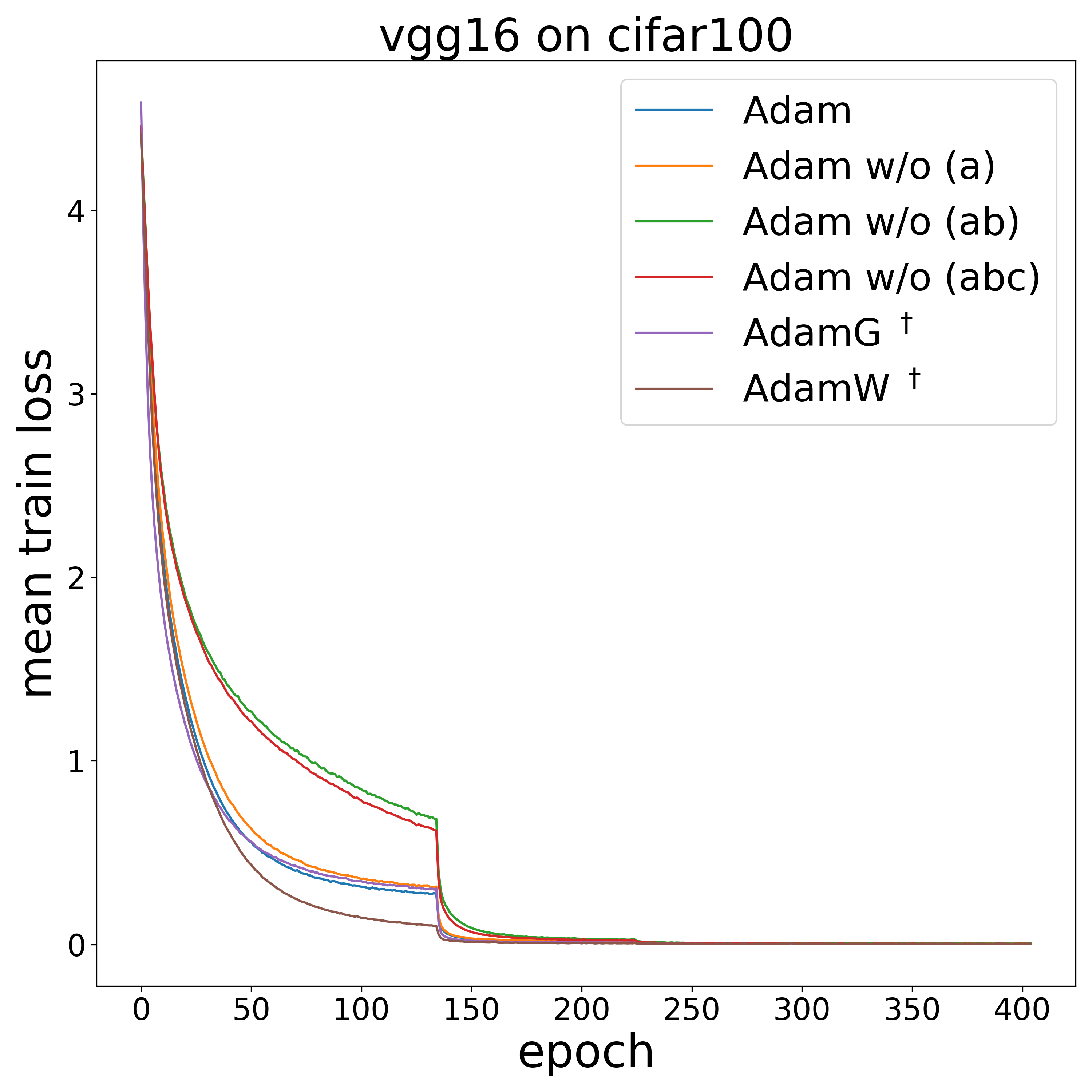}
    }
    \subfigure{
    \includegraphics[width=0.48\columnwidth]{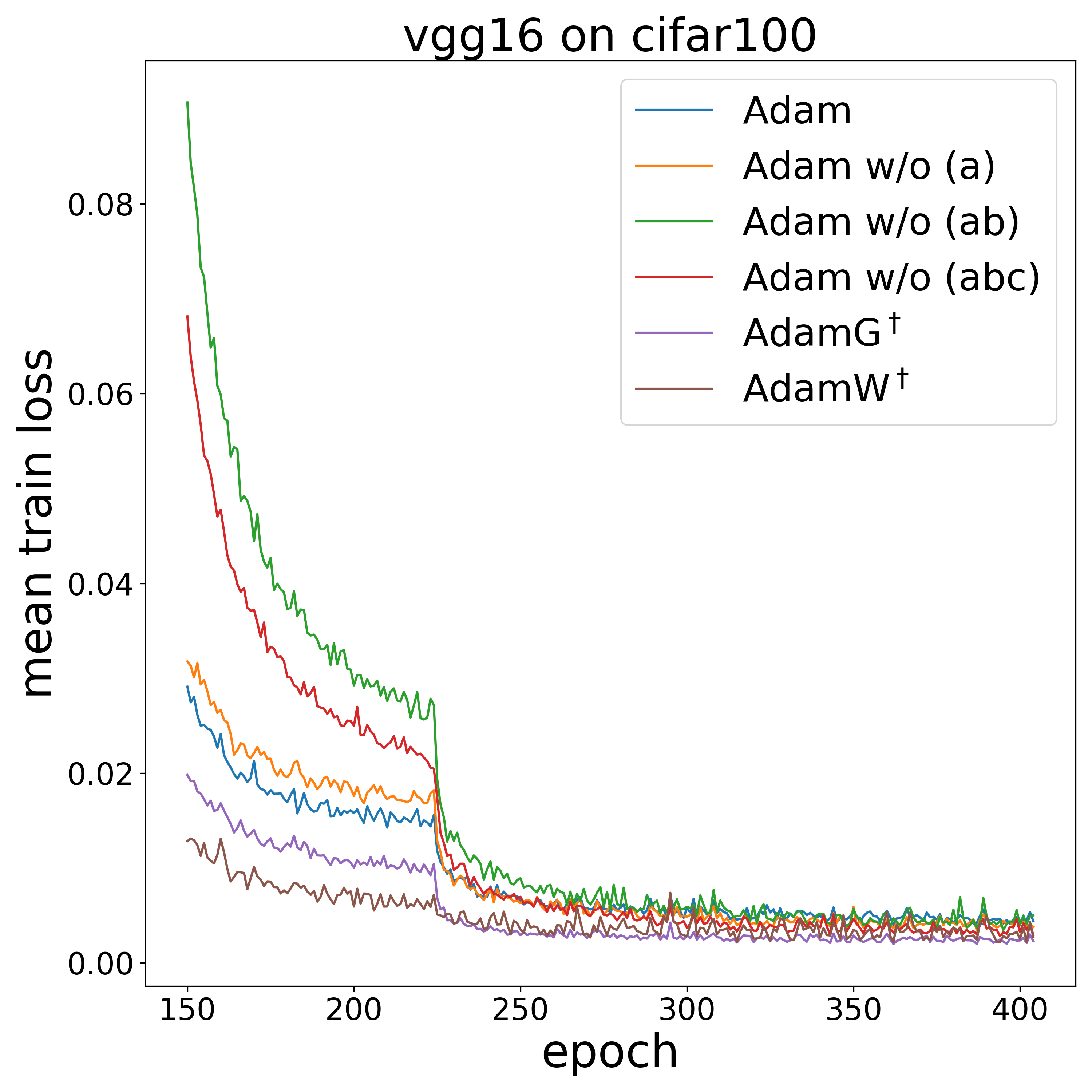}
    }
    \vspace{-2mm}
    \caption{\textbf{Training speed comparison with VGG16 on CIFAR100.} \emph{Left:} Mean training loss over all training epochs (averaged across 5 seeds) for different Adam variants. \emph{Right:} Zoom-in on the last epochs. Please refer to Table~\ref{tab:results} for the corresponding accuracies.}
    \label{fig:train_comparison_vgg16_cifar100}
     \vspace{-2mm}
\end{figure}

\begin{figure}
    \renewcommand{\captionlabelfont}{\bf}
    \centering
    \vspace{-4mm}
    \subfigure{
    \includegraphics[width=0.48\columnwidth]{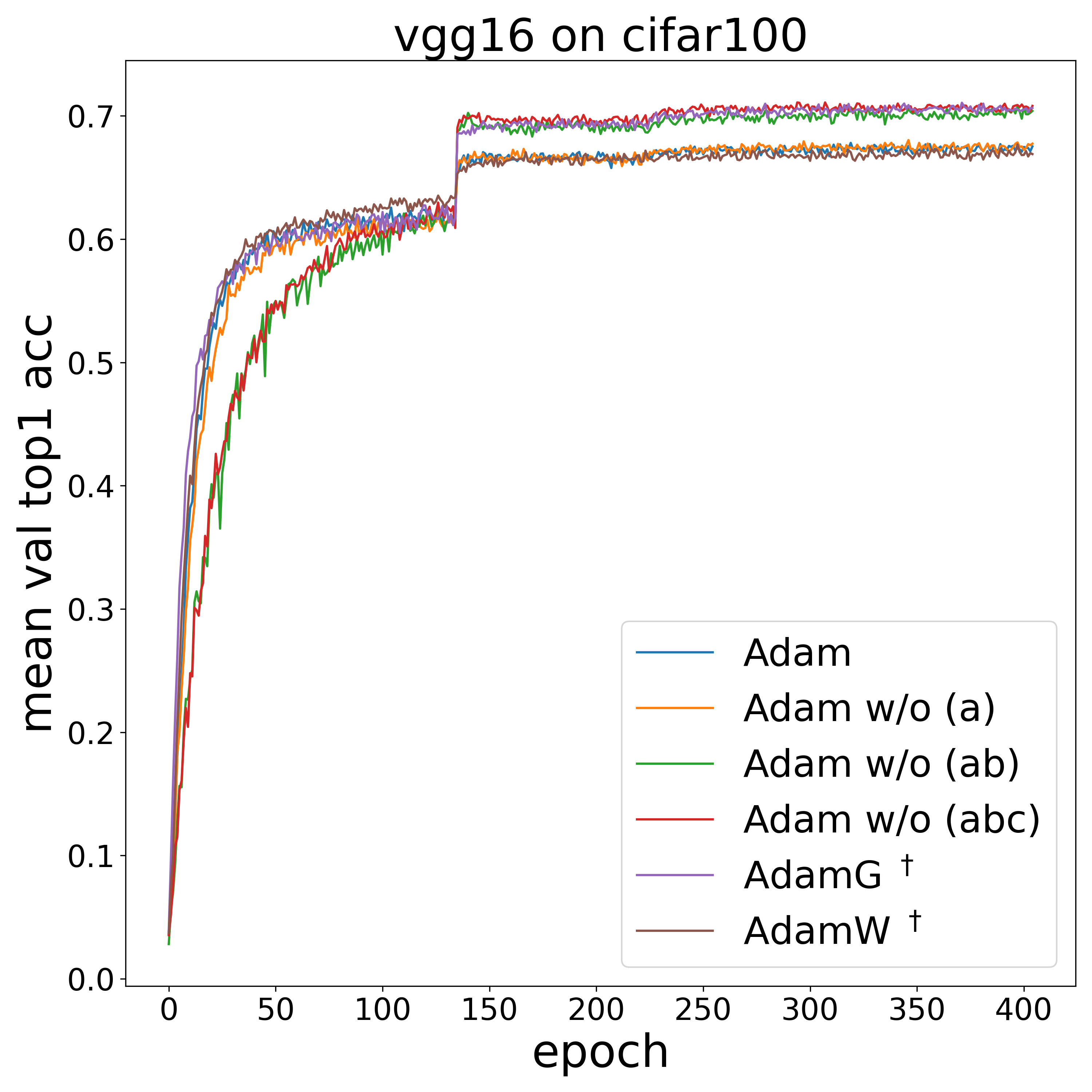}
    }
    \subfigure{
    \includegraphics[width=0.48\columnwidth]{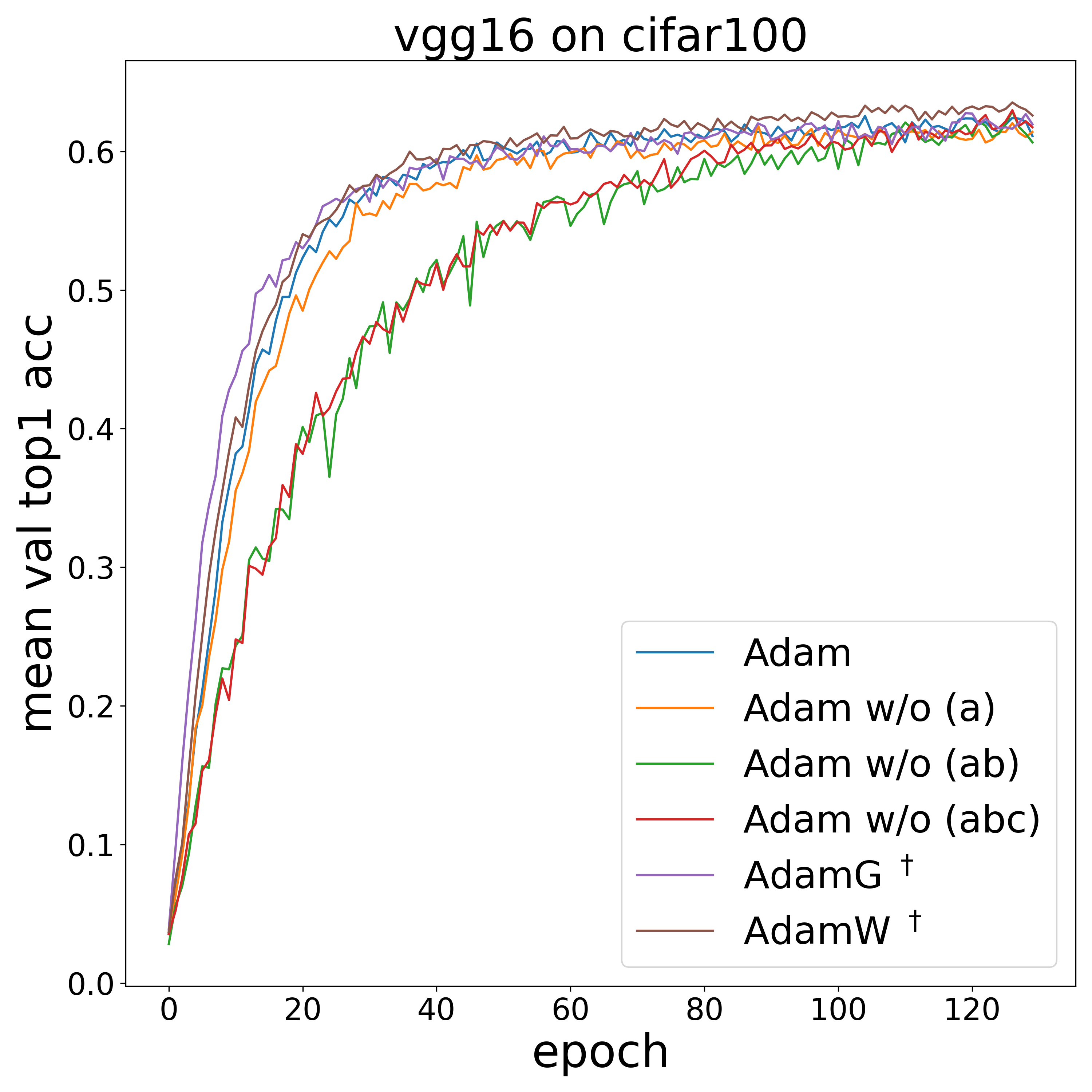}
    }
    \vspace{-2mm}
    \caption{\textbf{Accuracy comparison on the validation set with VGG16 BN on CIFAR100.} \emph{Left:} Mean training loss over all training epochs (averaged across 5 seeds) for different Adam variants. \emph{Right:} Zoom-in on the last epochs. Please refer to Table~\ref{tab:results} for the corresponding accuracies.}
    \label{fig:val_comparison_vgg16_cifar100}
     \vspace{-2mm}
\end{figure}

\begin{figure}
    \renewcommand{\captionlabelfont}{\bf}
    \centering
    \vspace{-4mm}
    \subfigure{
    \includegraphics[width=0.48\columnwidth]{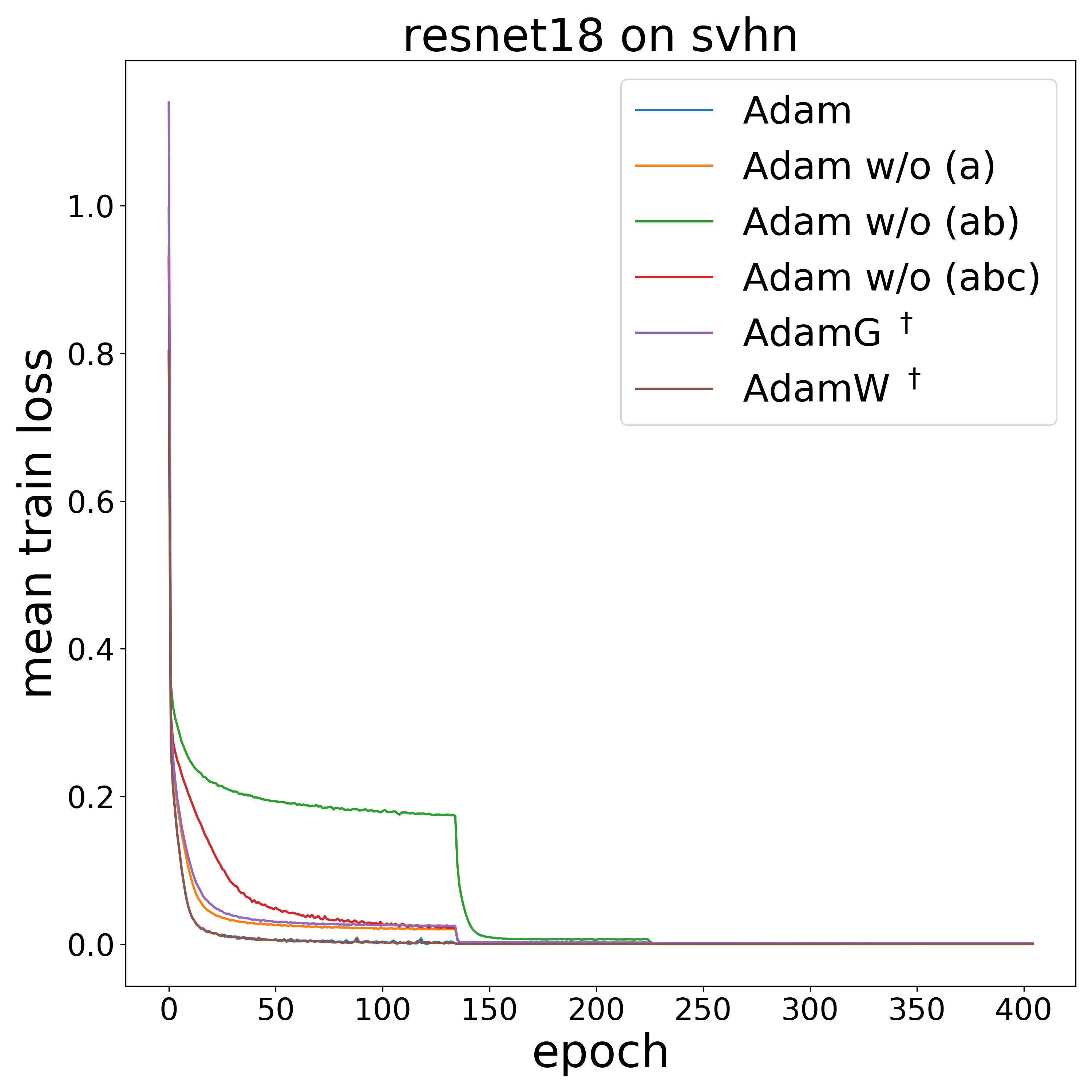}
    }
    \subfigure{
    \includegraphics[width=0.48\columnwidth]{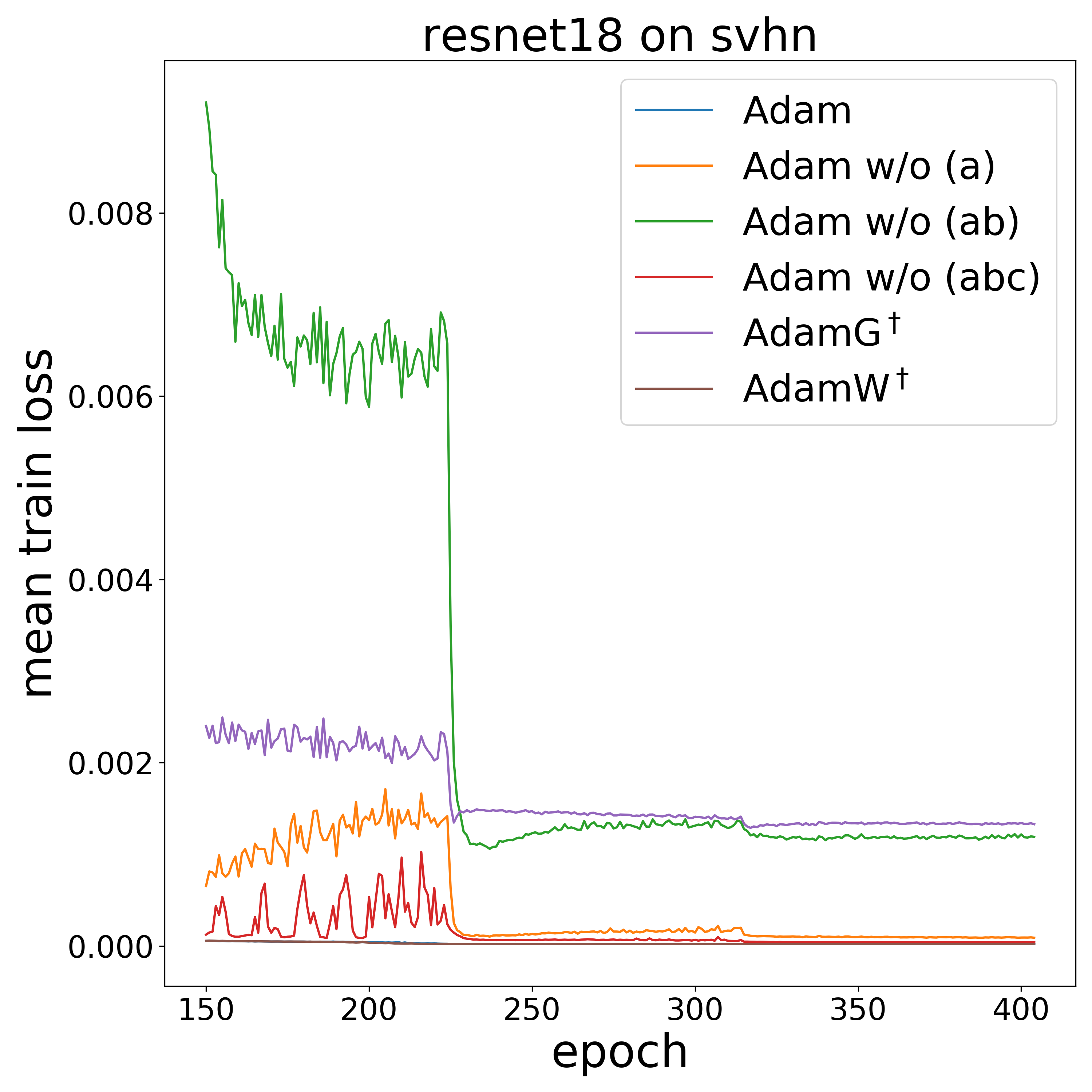}
    }
    \vspace{-2mm}
    \caption{\textbf{Training speed comparison with ResNet18 on SVHN.} \emph{Left:} Mean training loss over all training epochs (averaged across 5 seeds) for different Adam variants. \emph{Right:} Zoom-in on the last epochs. Please refer to Table~\ref{tab:results} for the corresponding accuracies.}
    \label{fig:train_comparison_resnet18_svhn}
     \vspace{-2mm}
\end{figure}

\begin{figure}
    \renewcommand{\captionlabelfont}{\bf}
    \centering
    \vspace{-4mm}
    \subfigure{
    \includegraphics[width=0.48\columnwidth]{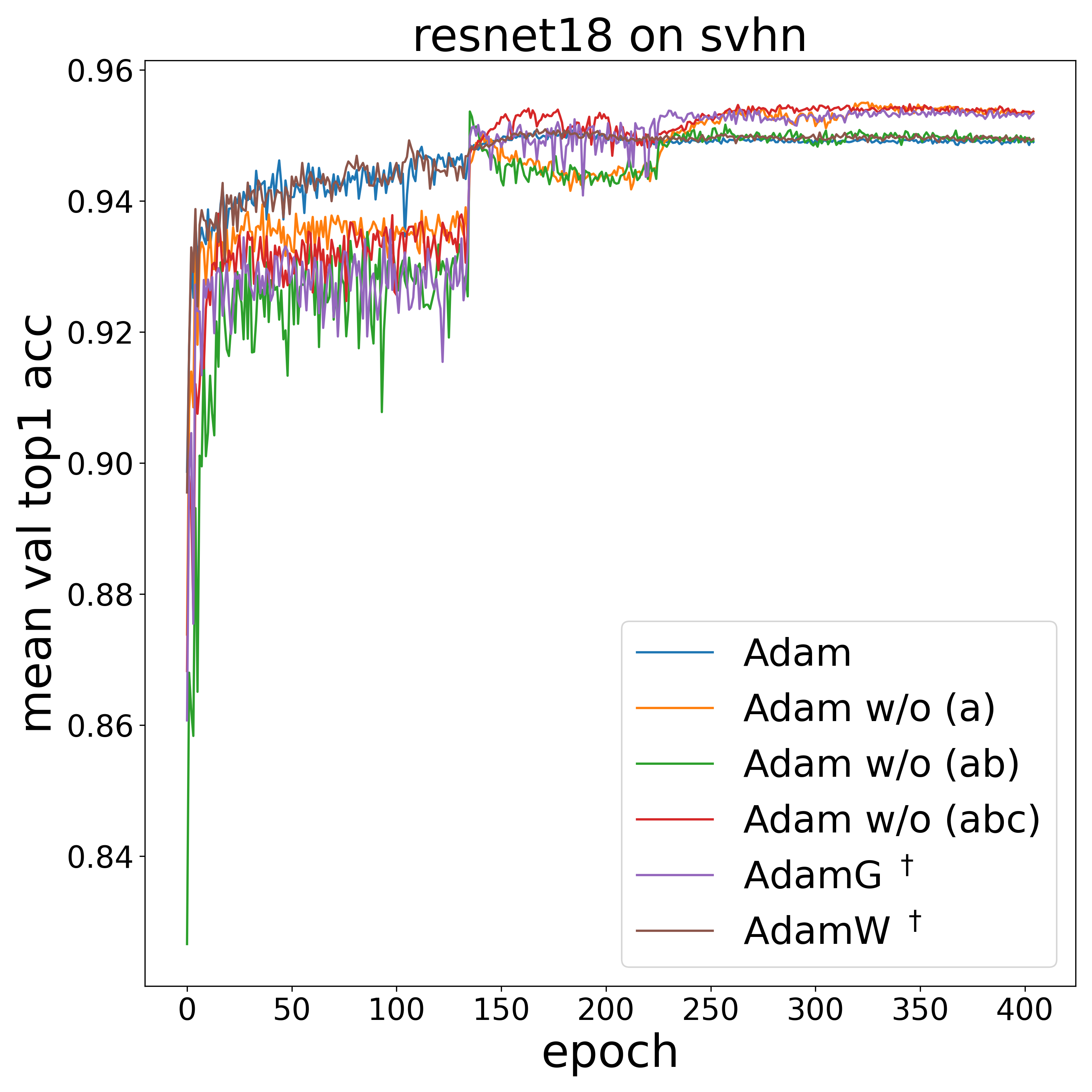}
    }
    \subfigure{
    \includegraphics[width=0.48\columnwidth]{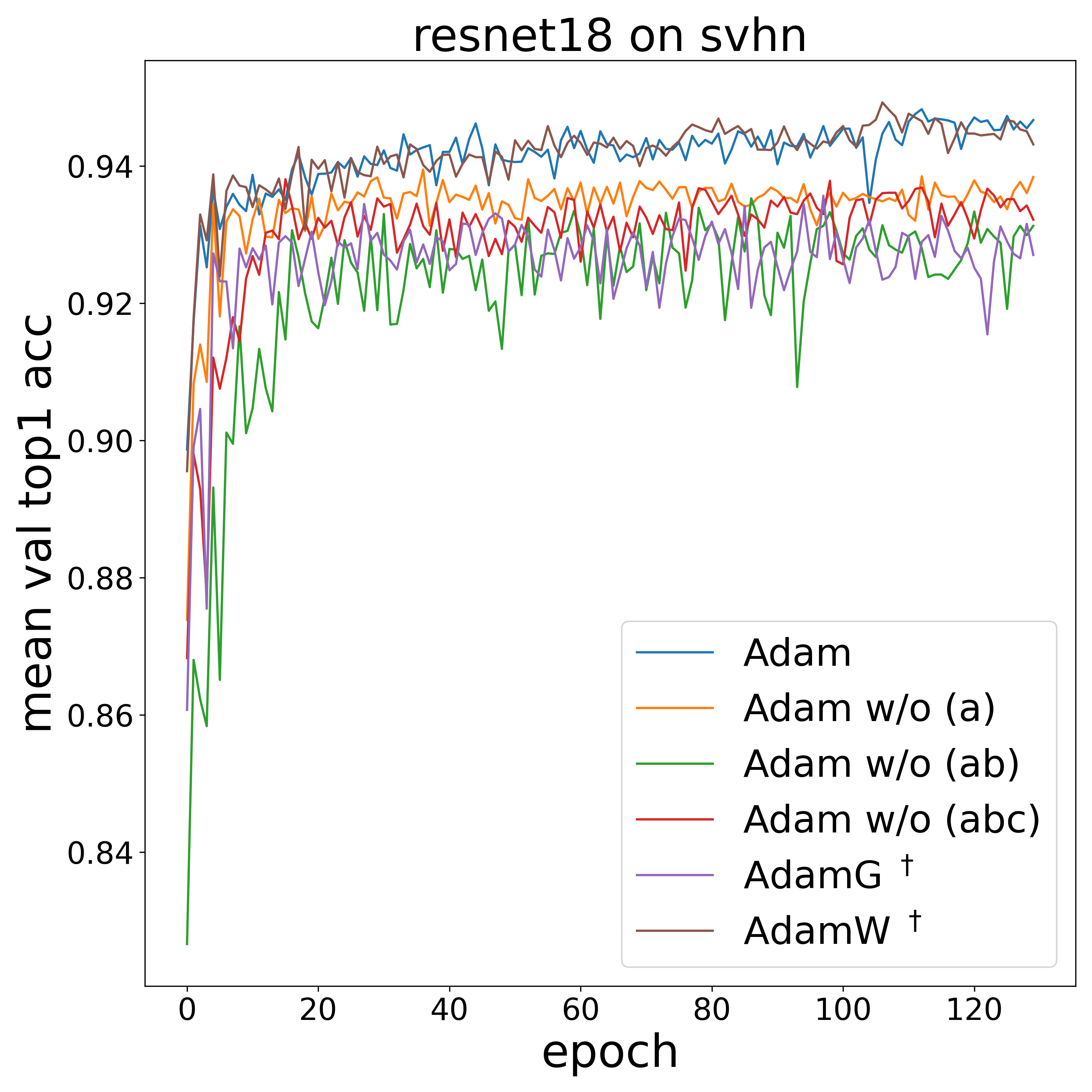}
    }
    \vspace{-2mm}
    \caption{\textbf{Accuracy comparison on the validation set with ResNet18 BN on SVHN.} \emph{Left:} Mean training loss over all training epochs (averaged across 5 seeds) for different Adam variants. \emph{Right:} Zoom-in on the last epochs. Please refer to Table~\ref{tab:results} for the corresponding accuracies.}
    \label{fig:val_comparison_resnet18_svhn}
     \vspace{-2mm}
\end{figure}

\begin{figure}
    \renewcommand{\captionlabelfont}{\bf}
    \centering
    \vspace{-4mm}
    \subfigure{
    \includegraphics[width=0.48\columnwidth]{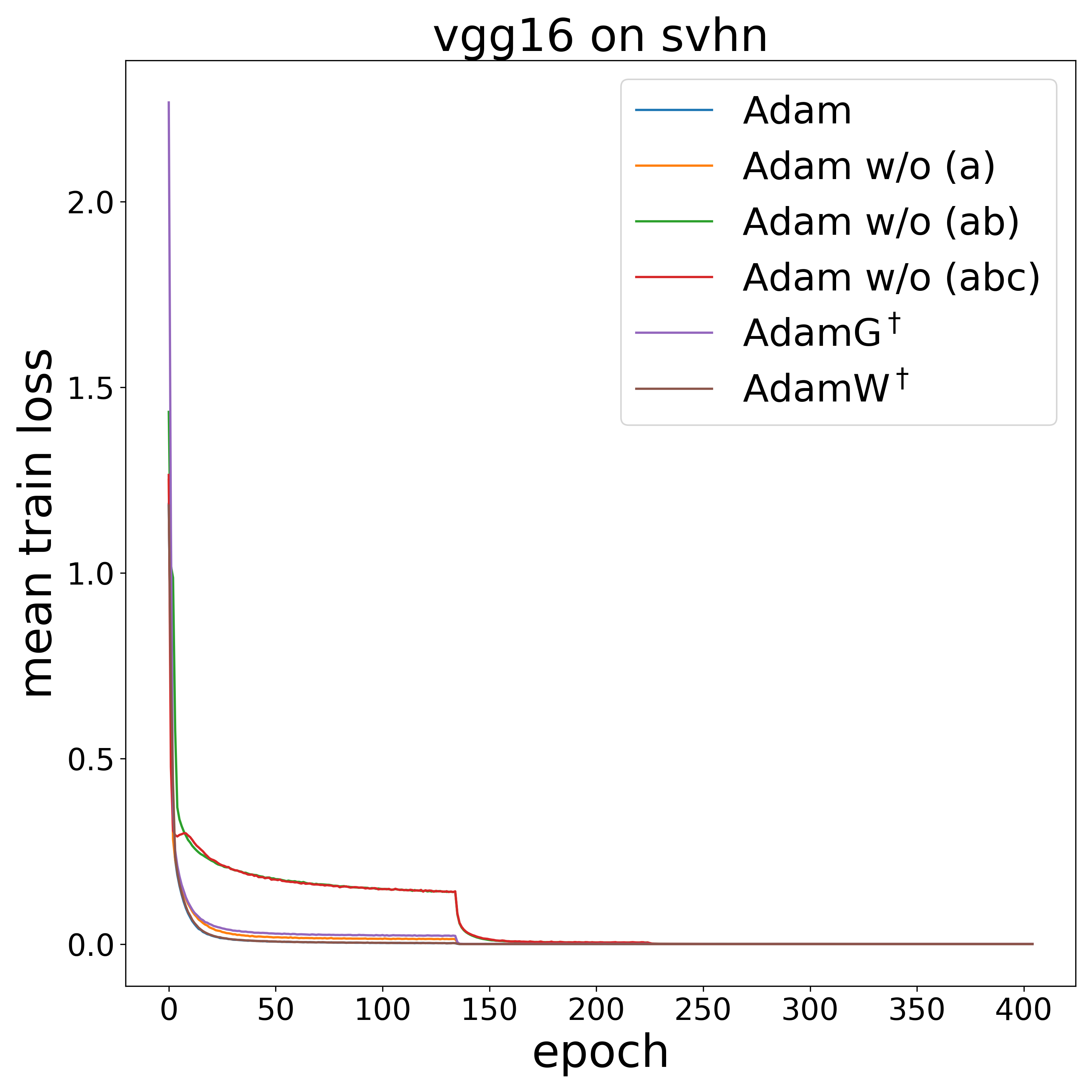}
    }
    \subfigure{
    \includegraphics[width=0.48\columnwidth]{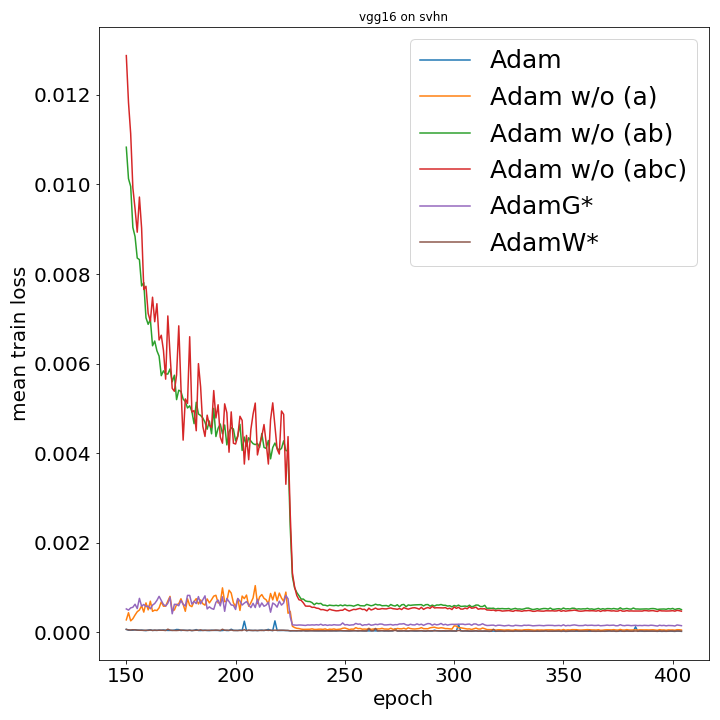}
    }
    \vspace{-2mm}
    \caption{\textbf{Training speed comparison with VGG16 on SVHN.} \emph{Left:} Mean training loss over all training epochs (averaged across 5 seeds) for different Adam variants. \emph{Right:} Zoom-in on the last epochs. Please refer to Table~\ref{tab:results} for the corresponding accuracies.}
    \label{fig:train_comparison_vgg16_svhn}
     \vspace{-2mm}
\end{figure}

\begin{figure}
    \renewcommand{\captionlabelfont}{\bf}
    \centering
    \vspace{-4mm}
    \subfigure{
    \includegraphics[width=0.48\columnwidth]{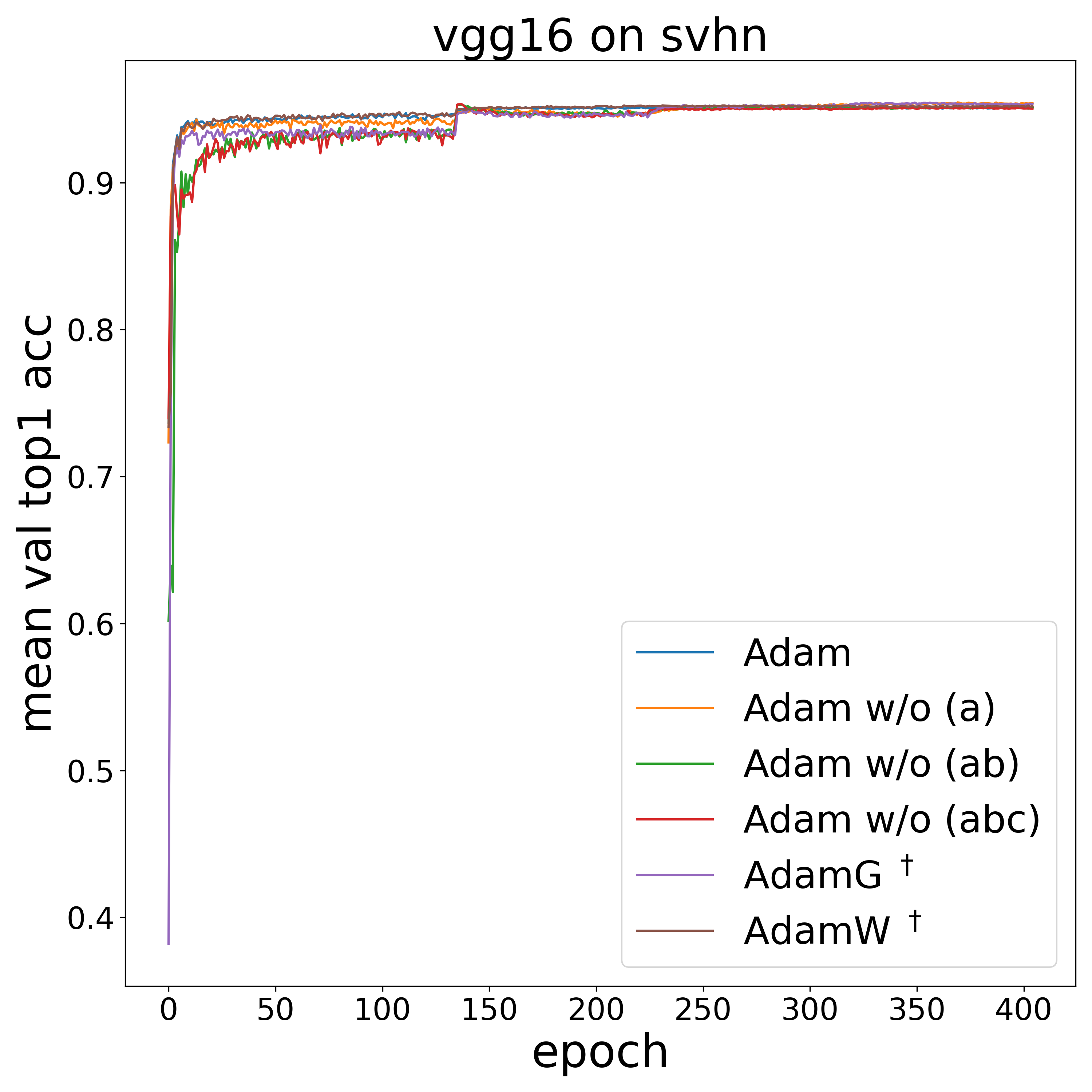}
    }
    \subfigure{
    \includegraphics[width=0.48\columnwidth]{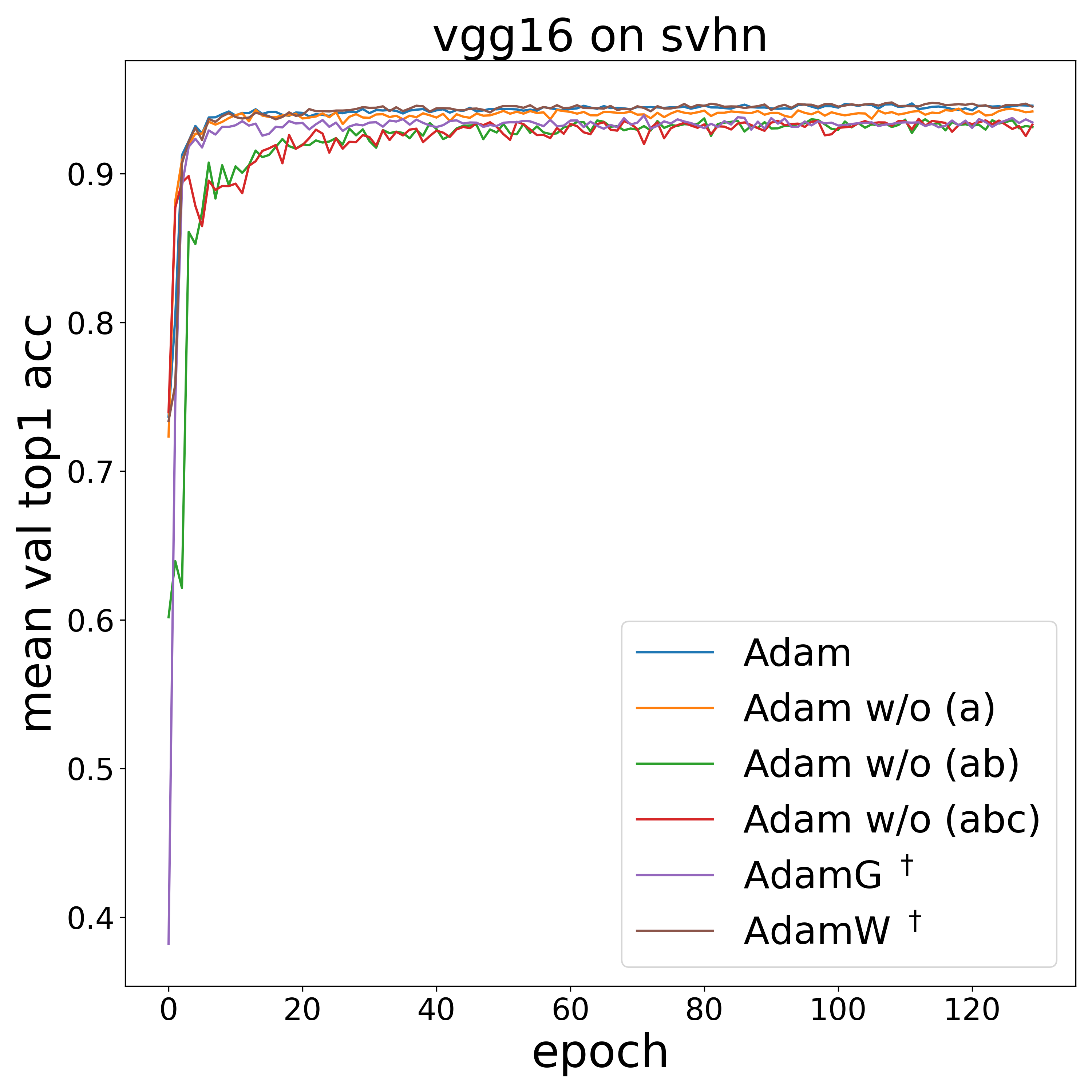}
    }
    \vspace{-2mm}
    \caption{\textbf{Accuracy comparison on the validation set with VGG16 BN on SVHN.} \emph{Left:} Mean training loss over all training epochs (averaged across 5 seeds) for different Adam variants. \emph{Right:} Zoom-in on the last epochs. Please refer to Table~\ref{tab:results} for the corresponding accuracies.}
    \label{fig:val_comparison_vgg16_svhn}
     \vspace{-2mm}
\end{figure}

\subsubsection{Batch Normalization without scaling and bias parameters}\label{bn_wo_affine}

In this last section, we observe the mean loss training curves associated to Adam, AdamW, AdamG, Adam w/o (a), Adam w/o (ab), Adam w/o (abc) on datasets CIFAR10, CIFAR100 and SVHN with architecture ResNet20, ResNet18 or VGG16 with BN w/o affine, corresponding to the accuracies given in  Table~\ref{tab:results2}. 

\begin{table}[t!]
\centering
\renewcommand{\figurename}{Table}
\caption{\textbf{Accuracy of Adam and its variants when training with BN w/o affine layers}. The figures in this table are the mean top1 accuracy $\pm$ the standard deviation over 5 seeds on the test set for CIFAR10, CIFAR100 and on the validation set for SVHN. $^\dagger$~indicates that the original method is only used on convolutional filters while Adam is used for other parameters.
}
\vspace{2mm}
\scalebox{0.73}
{
\begin{tabular}{@{} l | ccc | c c | c c @{}}
\toprule
       & \multicolumn{3}{c|}{CIFAR10}   & \multicolumn{2}{c|}{CIFAR100} & \multicolumn{2}{c}{SVHN}\\    
Method & \texttt{ResNet20} & \texttt{ResNet18} & \texttt{VGG16} & \texttt{ResNet18} & \texttt{VGG16} & \texttt{ResNet18} & \texttt{VGG16}\\
\midrule
Adam\hphantom{W*}  & 90.41 $\pm$ 0.06 & 93.67 $\pm$ 0.15 & 92.62 $\pm$ 0.15 & 71.60 $\pm$ 0.22 & 68.28 $\pm$ 0.19 & 95.29 $\pm$ 0.11 & 95.56 $\pm$ 0.18\\
AdamW$^\dagger$  & 90.36 $\pm$ 0.11 & 93.7 $\pm$ 0.16 & 93.03 $\pm$ 0.12 & 70.11 $\pm$ 0.31 & 69.68 $\pm$ 0.12 & 89.83 $\pm$ 0.28 & 95.63 $\pm$ 0.11 \\
AdamG$^\dagger$  & 91.12 $\pm$ 0.09 & 93.62 $\pm$ 0.14 &  93.20 $\pm$ 0.20 & 69.96 $\pm$ 0.34 & 70.07 $\pm$ 0.23  & 95.12 $\pm$ 0.09  & 95.62 $\pm$ 0.21\\
\midrule
Adam w/o (a)  & 91.15 $\pm$ 0.11 & 93.98 $\pm$ 0.18  & 93.12 $\pm$ 0.14 & 75.43 $\pm$ 0.13 & 70.01 $\pm$ 0.24 & 95.75 $\pm$ 0.09 & 95.64 $\pm$ 0.10\\
Adam w/o (ab)  & \textbf{91.38} $\pm$ 0.08 & \textbf{94.63} $\pm$ 0.08  & 93.45 $\pm$ 0.06 &  \textbf{75.68} $\pm$ 0.22 & 71.74 $\pm$ 0.15 & \textbf{95.77 $\pm$ 0.08} & \textbf{95.78} $\pm$ 0.07 \\
Adam w/o (abc) & 91.11 $\pm$ 0.11 &  94.02 $\pm$ 0.10 & \textbf{93.56} $\pm$ 0.09 & 75.38 $\pm$ 0.21 & \textbf{72.08} $\pm$ 0.22 &  95.51 $\pm$ 0.08 & 95.69 $\pm$ 0.09 \\
\bottomrule
\end{tabular}
}
\label{tab:results2}
\end{table}

\end{document}